\documentclass{article}

\usepackage[main, final]{neurips_2026}

\usepackage[pagebackref,breaklinks,colorlinks]{hyperref}

\usepackage[utf8]{inputenc} 
\usepackage[T1]{fontenc}    
\usepackage{url}            
\usepackage{booktabs}       
\usepackage{amsfonts}       
\usepackage{nicefrac}       
\usepackage{xcolor}         
\usepackage{graphicx}
\usepackage{subcaption}
\usepackage{amsmath}
\usepackage{wrapfig}
\usepackage{pifont}
\usepackage{tikz}
\usepackage{multirow}
\usepackage{makecell}
\usepackage{soul}

\usepackage[accsupp]{axessibility}  

\usepackage{enumitem}
\setlist[itemize]{leftmargin=5.5mm} 

\usepackage{algpseudocode}
\usepackage{placeins}
\usepackage[linesnumbered,ruled,vlined]{algorithm2e}
\SetKwInput{KwInput}{Input}  
\SetKwInput{KwOutput}{Output}
\SetKwInput{KwRequire}{Require}
\SetKwInput{KwIP}{\em Importance Probing}
\SetKwInput{KwMA}{\em Main Adaptation}

\SetCommentSty{mycommfont}

\usepackage{orcidlink}

\title{AIR: Rethinking Image-Text Offset Alignment in \\Multimodal Contrastive  Representation Space}

\author{Guimeng Liu \quad
Milad Abdollahzadeh \quad
Ngai-Man Cheung\\
\\
Temasek Laboratories, Singapore University of Technology and Design \\
}

\begin{document}

\maketitle


\begin{abstract}



Image-text offset alignment, which refers to the consistency between visual transformation directions and their corresponding textual transformation directions, has become  an important relational property in multimodal contrastive representation spaces such as CLIP and SigLIP.
This property plays a central role in 
text-driven visual transformation tasks,
including generative model domain adaptation and text-guided object-centric image editing, where textual offsets are used to guide visual transformations.
These methods fundamentally rely on the assumption that image and text offsets are well aligned, such that textual transformation directions can provide reliable guidance for corresponding visual transformations.
However, the validity of this alignment assumption has not been systematically examined.
In this work, we question this foundational assumption by conducting a comprehensive empirical analysis of image-text offset alignment in multimodal contrastive representation space. Our findings reveal not only noticeable offset misalignment but also a meaningful positive correlation between image-text offset misalignment and semantic concept distance across six large datasets and eight contrastive vision-language models. Based on this discovery, we propose Adaptation with Iterative Refinement (AIR), a method that iteratively refines text offsets through anchor sampling and our proposed concept description learning to mitigate the downstream effect of image-text offset misalignment and improve guidance accuracy.
Comprehensive experiments on zero-shot generative model domain adaptation and text-guided object-centric image editing across 40 setups,
including qualitative, quantitative, and user studies, consistently show that AIR enables state-of-the-art performance in these tasks.
\emph{Project Page:} \url{https://guimeng-leo-liu.github.io/AIR-Rethinking-Image-Text-Offset-Alignment/}

\end{abstract}

\addtocontents{toc}{\protect\setcounter{tocdepth}{-1}}
\section{Introduction}
\label{sec:intro}

\begin{figure*}[!ht]
    \centering
    \includegraphics[width=\textwidth]{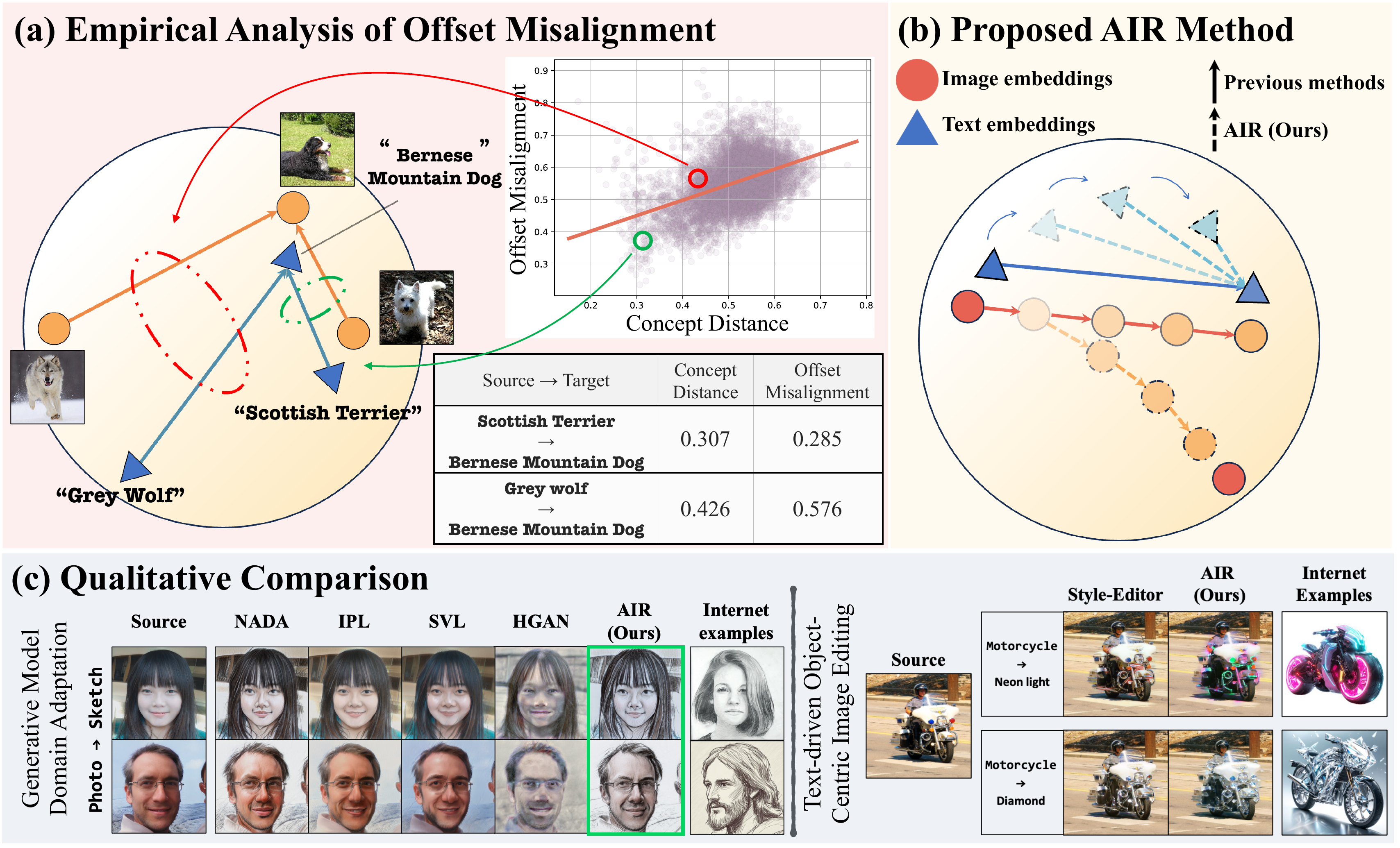}
    \caption{
    {\bf Our contributions:}
    {\bf (a)} We question the assumption of image-text offset alignment and perform a comprehensive analysis of offset alignment in multimodal contrastive representation space. 
    {\bf Our analysis reveals not only noticeable misalignment between image offsets (\textcolor{orange}{orange arrow}) and text offsets (\textcolor{blue}{blue arrow}), but also a meaningful positive correlation between offset misalignment and concept distance}.
    For example, in the ImageNet-1K dataset, the \texttt{``Grey Wolf''} is a more distant concept to the \texttt{``Bernese Mountain Dog''} (concept distance=0.426) than the \texttt{``Scottish Terrier''} (concept distance=0.307). 
    Accordingly,
    \texttt{``Grey Wolf''} $\rightarrow$ \texttt{``Bernese Mountain Dog''}
    has higher offset misalignment than
    \texttt{``Scottish Terrier''} $\rightarrow$ \texttt{``Bernese Mountain Dog''}
    (0.576 vs 0.285).
    This misalignment is overlooked in existing uses of image-text offsets (Sec.~\ref{sec:Empirical study}).
    {\bf (b)} Leveraging our discovery, we propose Adaptation with Iterative Refinement (AIR) to iteratively refine text offsets and mitigate the downstream effect of offset misalignment for text-driven visual transformations, which we refer as guidance misalignment (Sec.~\ref{sec:Methodology}).
    {\bf (c)} Our proposed AIR consistently reduces the guidance misalignment and enables SOTA performance in
    important text-guided generative modeling tasks, e.g., 
    capturing rich target concepts' style and details across diverse setups.
    (see Sec.~\ref{sec:experiments} and 
    Sec.~\ref{sec:more_experimental_results} 
    for detailed results).
    }
    \label{fig:CLIPspace}
\end{figure*}


Multimodal contrastive models, such as CLIP and SigLIP, learn a shared representation space by aligning paired image and text embeddings \cite{radford2021CLIP, zhai2023sigmoid}. Beyond this pairwise alignment, many downstream methods rely on a stronger relational property: for two semantic concepts in the shared representation space, the offset between their text embeddings is expected to align with the corresponding offset between their image embeddings. This consistency between textual and visual transformation directions is referred as {\em image-text offset alignment}.
This property plays a central role in vision-language tasks that reason about semantic transformations, where text offsets serve as compact representations of semantic changes for guiding, retrieving, or evaluating visual transformations. Consequently, image-text offset alignment has been widely exploited in tasks including zero-shot generative model domain adaptation \cite{gal2022stylegannada, guo2023ipl, jeon2023svl, Anees2024HyperGANCLIP}, text-guided image editing \cite{patashnik2021styleclip, diffusionclip, oh2025mitigating, kwon2022clipstyler, park2025styleeditor}, composed image retrieval \cite{couairon2022embedding, yang2024semantic, tursun2026pdv}, and direction-aware evaluation metrics \cite{diffusionclip, wang2025editclip, kim2025augclip}.
Specifically, in zero-shot generative model domain adaptation (ZSGM) and text-guided object-centric image editing (TOIE), text offsets are used to guide corresponding image offsets toward desired visual transformations; in composed image retrieval, image and text offsets are composed to retrieve target images matching the intended semantic change; and in direction-aware evaluation metrics, the alignment between image and text offsets measures whether an observed visual change follows the textual instruction. 

Despite their different formulations and objectives, these tasks fundamentally rely on the same underlying assumption: image and text offsets are sufficiently aligned in multimodal contrastive representation spaces, such that textual transformation directions can provide reliable guidance for corresponding visual transformations. {\em Critically, the validity of image-text offset alignment has not been systematically examined.}



In this work, we question this foundational assumption by conducting, to the best of our knowledge, the first systematic empirical analysis of image-text offset alignment within multimodal contrastive representation space. 
\textbf{Our primary findings} reveal not only noticeable offset misalignment, but also a meaningful positive correlation between image-text offset misalignment and semantic concept distance: for closely related concept pairs, misalignment tends to be smaller, whereas it increases as concepts become more distant
(Fig. \ref{fig:CLIPspace}).
This correlation is consistently observed across six large datasets and eight contrastive vision-language models.

Leveraging this discovery, we propose {\em Adaptation with Iterative Refinement (AIR)}, a framework designed to mitigate the downstream effect of image-text offset misalignment and enhance text-driven visual transformations.
We refer to the downstream effect of image-text offset misalignment in text-driven visual transformation as \emph{guidance misalignment}.
AIR iteratively samples intermediate transformed concepts as {\em anchors}, which lie closer to the target than the source in representation space (Sec. \ref{sec:concept_shifts_during_adaptation}), and refines text offsets with these anchors to reduce guidance misalignment and improve guidance accuracy. 
Since the textual descriptions of these anchor concepts are unknown, we introduce {\em concept description learning} to infer them dynamically. 
Our extensive results consistently demonstrate that AIR enables improved transformation quality
and SOTA performance in two important 
text-guided
generative modeling tasks: generative model domain adaptation and text-guided image editing.
Our main contributions are summarized as follows:
\begin{itemize}
    \item We question image-text offset alignment by conducting the first comprehensive empirical analysis on representative multimodal contrastive representation spaces. Our study reveals noticeable offset misalignment and a meaningful positive correlation with concept distance (Sec. \ref{sec:Empirical study}).
    \item Based on our discovery, we introduce Adaptation with Iterative Refinement (AIR), a novel framework that leverages the discovered correlation to reduce guidance misalignment and improve transformation quality through iterative anchor sampling and a new concept description learning strategy to dynamically infer anchor's concept descriptions (Sec. \ref{sec:Methodology}).
    \item Our extensive experiments
    under two important tasks, ZSGM and TOIE, demonstrate consistent improvements in transformation quality, validated by qualitative, quantitative, and user study results, achieving state-of-the-art performance
    (Sec. \ref{sec:experiments} and Supp.).
\end{itemize}

{\bf Remark:} 
Our discovery of image-text offset misalignment in multimodal contrastive space can be viewed as an analogy to text offset misalignment studies in unimodal text embedding space for natural language processing (NLP). In NLP, analogical reasoning
\cite{mikolov2013linguistic, mikolov2013efficient, mikolov2013distributed, levygoldberg2014linguistic}
relies on aligning offsets between word vectors, such as 
alignment between 
$E_v$(\texttt{``Man''}) - $E_v$(\texttt{``Woman''}), and 
$E_v$(\texttt{``King''}) - $E_v$(\texttt{``Queen''}), 
where $E_v$ denotes a text vector. Research
indicates that the accuracy of analogical reasoning improves with similar, nearby concepts but decreases with growing distance \cite{levy2015improving, koper2015multilingual, rogers2017too, fournier2020analogies}. Similarly, our finding of a positive correlation between image-text offset misalignment and concept distance in multimodal contrastive model reveals a similar distance-dependent relationship in multimodal contrastive representation spaces.

\section{Preliminaries: Image-Text Offset Alignment} \label{sec:Preliminaries}

\textbf{Image-Text Offset Alignment.}
Let $E_I$ and $E_T$ denote the image and text encoders of a multimodal contrastive model, respectively. 
For two concepts $\alpha$ and $\beta$, with corresponding visuals $I_\alpha, I_\beta$ and textual descriptions $T_\alpha, T_\beta$, we define the image offset and text offset as:
$\Delta I_{\alpha\rightarrow \beta} = E_I(I_\beta) - E_I(I_\alpha)$, and
$\Delta T_{\alpha\rightarrow \beta} = E_T(T_\beta) - E_T(T_\alpha)$.
Image-text offset alignment refers to the consistency between $\Delta I_{\alpha\rightarrow \beta}$ and $\Delta T_{\alpha\rightarrow \beta}$ in multimodal contrastive representation space. 
A common assumption is that the text offset between two concepts 
aligns with the corresponding image offset.

Image-text offset alignment
 is used in different vision-language tasks. In {\em text-driven visual transformation tasks} including zero-shot generative model adaptation and text-guided image editing, text offsets are used as optimization guidance for producing visual transformations. 
In composed image retrieval, image and text offsets are composed to retrieve target images matching the intended semantic change. 
In evaluation metrics, the alignment between image and text offsets is used to measure whether an observed visual change follows a textual instruction. 
Despite these different uses, these tasks share the same underlying assumption that image and text offsets are well aligned in multimodal contrastive representation space.

\textbf{Directional Loss.}
Specifically, in text-driven visual transformation tasks, image-text offset alignment is typically operationalized through directional loss formulations.
Given a source visual $I_{\mathcal{S}}$, a transformed visual $I_t$, and textual descriptions of the source and target concepts, denoted by $T_{\mathcal S}$ and $T_{\mathcal T}$, respectively, 
text-driven visual transformation methods 
compute a text offset from the source concept to the target concept and optimize the transformation so that the image offset aligns with this text offset:
\begin{equation}
\begin{split}
    \mathcal{L}_{direction} &= 1 - \text{cos}(\Delta I_{\mathcal{S}\rightarrow t}, \Delta T_{\mathcal{S}\rightarrow \mathcal{T}}), \\
    \text{where } \Delta I_{\mathcal{S}\rightarrow t} &= E_I(I_t) - E_I(I_{\mathcal{S}}), \\
    \text{and } \Delta T_{\mathcal{S}\rightarrow \mathcal{T}} &= E_T(T_{\mathcal{T}}) - E_T(T_{\mathcal{S}}).
\end{split}
\label{eq:Loss_directional}
\end{equation}
Here, \(\text{cos}(x,y)=x \cdot y/(|x||y|)\) denotes cosine similarity.
The directional loss is minimized when the offset between the source and transformed images is aligned with the offset between the source and target text embeddings. As such, directional loss fundamentally assumes that textual offsets faithfully capture the desired semantic concept transformation and can therefore provide reliable guidance for corresponding visual transformations.

\section{A Closer Look at Image-Text Offset Misalignment}  \label{sec:Empirical study}

\begin{figure*}[htbp]
    \centering
    \includegraphics[width=\textwidth]{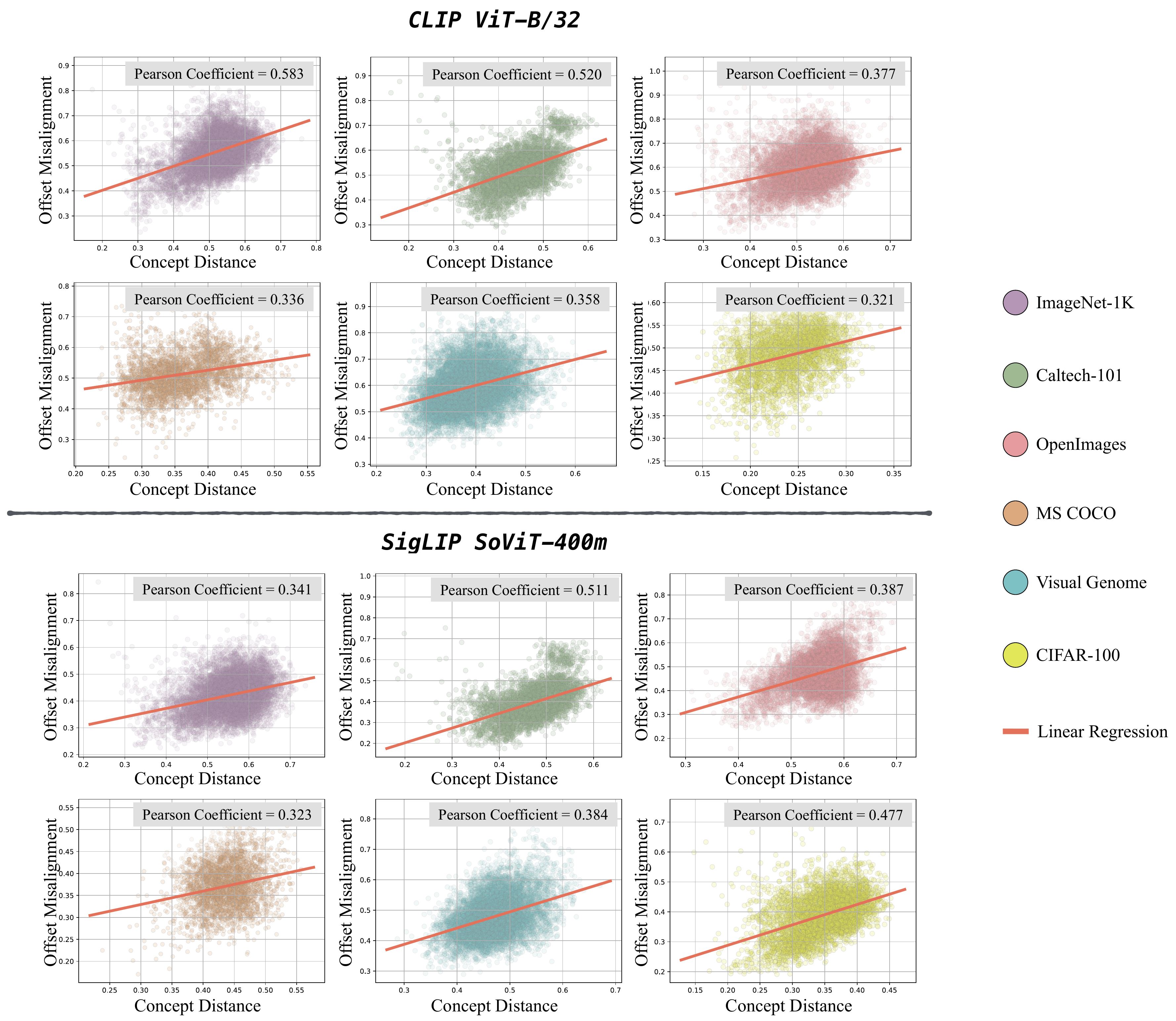}
    \caption{{\bf Empirical analysis of offset misalignment.}
    We plot the offset misalignment (Eq. \ref{eq:Offset Misalignment}) vs concept distance in CLIP ViT-B/32 and SigLIP SoViT-400m for $N = 10,000$ text-image pairs sampled from 6 large publicly available datasets (details in 
    Sec. \ref{ssec:Detail_of_empirical_study}
    ; total 120, 000 text-image pairs
    ).
    Our results show that there is a meaningful correlation (measured by Pearson coefficient \cite{pearsoncorr}) between offset misalignment and concept distance for datasets with different distributions, {i.e.,} close concepts has less offset misalignment.
    \textbf{
    Furthermore, we have consistent observations for other multimodal contrastive representation  spaces, see
    Sec. \ref{sec:offset_misalignment_in_more_clip_space}.
    }}
    \label{fig:Empirical study}
\end{figure*}

Previous works  assume that, for two semantic concepts $\alpha$ and $\beta$, the image offset $\Delta I_{\alpha \rightarrow \beta}$ and the corresponding text offset $\Delta T_{\alpha \rightarrow \beta}$ are well aligned in multimodal contrastive representation spaces (e.g., CLIP and SigLIP). This assumption serves as a foundational principle underlying several vision-language tasks discussed in Sec.~\ref{sec:Preliminaries}, and is directly operationalized through directional loss formulations in text-driven visual transformation tasks.
Despite its widespread adoption, we identify two fundamental limitations of this image-text offset alignment assumption:
\begin{itemize}
    \item Multimodal contrastive models are trained with contrastive loss to maximize cosine similarity between corresponding image-text pairs, {\em i.e.,} maximize $\cos(E_{I}(I_{\alpha}), E_{T}(T_{\alpha}))$ for concept $\alpha$ ({\em e.g.}, cat), or maximize $\cos(E_{I}(I_{\beta}), E_{T}(T_{\beta}))$ for concept $\beta$ ({\em e.g.}, dog). Note that the degree of alignment of image offset $\Delta I_{\alpha \rightarrow \beta}$ and text offset $\Delta T_{\alpha \rightarrow \beta}$ in multimodal contrastive space is not studied in the literature.
    \item In addition, this degree of alignment between $\Delta I_{\alpha \rightarrow \beta}$ and $\Delta T_{\alpha \rightarrow \beta}$ may vary based on the distance between two concepts pair $\alpha$ and $\beta$.
\end{itemize}

In this section, 
we take a closer look at this degree of offset alignment between two different modalities.
Inspired by offset misalignment in NLP, we conduct an empirical study on large public datasets to analyze the offset misalignment between image and text modalities in multimodal contrastive space. Our analysis suggests that {\bf there is  misalignment between $\Delta I_{\alpha \rightarrow \beta}$ and $\Delta T_{\alpha \rightarrow \beta}$, and this misalignment tends to increase as concepts $\alpha$ and $\beta$ become more distant}.

\subsection{Empirical Analysis of Image-Text Offset Misalignment} \label{ssec:OM}
We conduct an empirical experiment on public datasets to evaluate the degree of alignment between image and text offsets.
We randomly sample two classes for each dataset as a pair of concept $(\alpha, \beta)$. Then, the images within each class are used alongside the related concept textual description ({\em e.g.,} label) of each class to measure offset misalignment $\mathcal{M}(\alpha, \beta)$:
\begin{equation}
\begin{split}
    \mathcal{M} (\alpha, \beta) &= 1 - \text{cos}(\Delta I_{\alpha\rightarrow \beta}, \Delta T_{\alpha\rightarrow \beta}), \\
    \text{where } \Delta I_{\alpha\rightarrow \beta} &= E_I\overline{(I_{\beta})} - E_I\overline{(I_{\alpha})}, \\
    \text{and } \Delta T_{\alpha\rightarrow \beta} &= E_T(T_{\beta}) - E_T(T_{\alpha})
\end{split}
\label{eq:Offset Misalignment}
\end{equation}
where $E_I\overline{(I_{\alpha})}$ is the average embedding of all images of the class (concept) $\alpha$ in multimodal contrastive space. 
Note we compute offset misalignment at the concept-distribution level rather than the instance level, since instance-specific visual variations may introduce semantic differences that are not described by the class-level text offset.
To measure the distance between two concepts (concept distance) denoted by $\mathcal{D} (\alpha, \beta)$, we use cosine distance between images of two classes, {\em i.e.}, $\mathcal{D} (\alpha, \beta) = 1 - \text{cos}(E_I\overline{(I_{\beta})}, E_I\overline{(I_{\alpha})})$.
We repeat this for all pairs of concepts in each dataset. 
Then, we plot $\mathcal{M}(\alpha, \beta)$ against $\mathcal{D}(\alpha, \beta)$ for each pair of concepts.

{\bf Experimental Setup.}
We conduct experiment across 8 multimodal contrastive models, including CLIP ViT-B/32 \cite{radford2021CLIP}, CLIP RN50x64, CLIP ConvNext-L, SigLIP ViT-L/16-256 \cite{zhai2023sigmoid}, SigLIP SoViT-400m, EVA-CLIP-8B \cite{evaclip}, DFN5B-CLIP \cite{DFNCLIP}, and MobileCLIP-S2 \cite{faghrimobileclip2}. We use 6 large and multi-class datasets that are publicly available, including ImageNet \cite{deng2009imagenet}, Caltech-101 \cite{fei2007caltech101}, OpenImages \cite{OpenImages}, COCO \cite{lin2014mscoco}, Visual Genome \cite{krishna2017visualgenome}, and CIFAR-100 \cite{krizhevsky2009cifar100} (details in 
Sec. \ref{ssec:Detail_of_empirical_study}).

{\bf Results.}
Fig. \ref{fig:Empirical study} shows the offset misalignment against the concept distance in CLIP ViT-B/32 and SigLIP SoViT-400m for 6 public datasets.
As shown in the plots, for all datasets, apart from their different distributions and characteristics, there is a positive correlation between offset misalignment and concept distance.
Particularly, if two concepts $\alpha$ and $\beta$ are distant, there is a higher misalignment between image offset $\Delta I_{\alpha\rightarrow \beta}$ and corresponding text offset $\Delta T_{\alpha\rightarrow \beta}$.
This means that given $I_{\alpha}$, $T_{\alpha}$ and $T_{\beta}$, it is sub-optimal to align $\Delta I_{\alpha\rightarrow \beta}$ and $\Delta T_{\alpha\rightarrow \beta}$ to find $I_{\beta}$.
On the other hand, if two concepts $\alpha$ and $\beta$ are closer, potentially, it is more accurate to align $\Delta I_{\alpha\rightarrow \beta}$ and $\Delta T_{\alpha\rightarrow \beta}$ to find $I_{\beta}$. We have consistent observations for other multimodal contrastive representation spaces, see Sec. \ref{sec:offset_misalignment_in_more_clip_space}.

\noindent
\begin{tikzpicture}
\node[
    draw=teal!70!black,
    rounded corners=3pt,
    line width=0.8pt,
    fill=teal!3,
    inner xsep=8pt,
    inner ysep=6pt,
    text width=\dimexpr\linewidth-18pt\relax,
    align=left
] {
    \textbf{\emph{Finding:}} \hspace{0.1em}
    Offset misalignment between image and text modalities in multimodal contrastive representation space correlates positively with concept distance.
};
\end{tikzpicture}
\vspace{-1em}


\section{Methodology: Adaptation with Iterative Refinement} \label{sec:Methodology}

\begin{figure*}[ht]
    \centering
    \includegraphics[width=\textwidth]{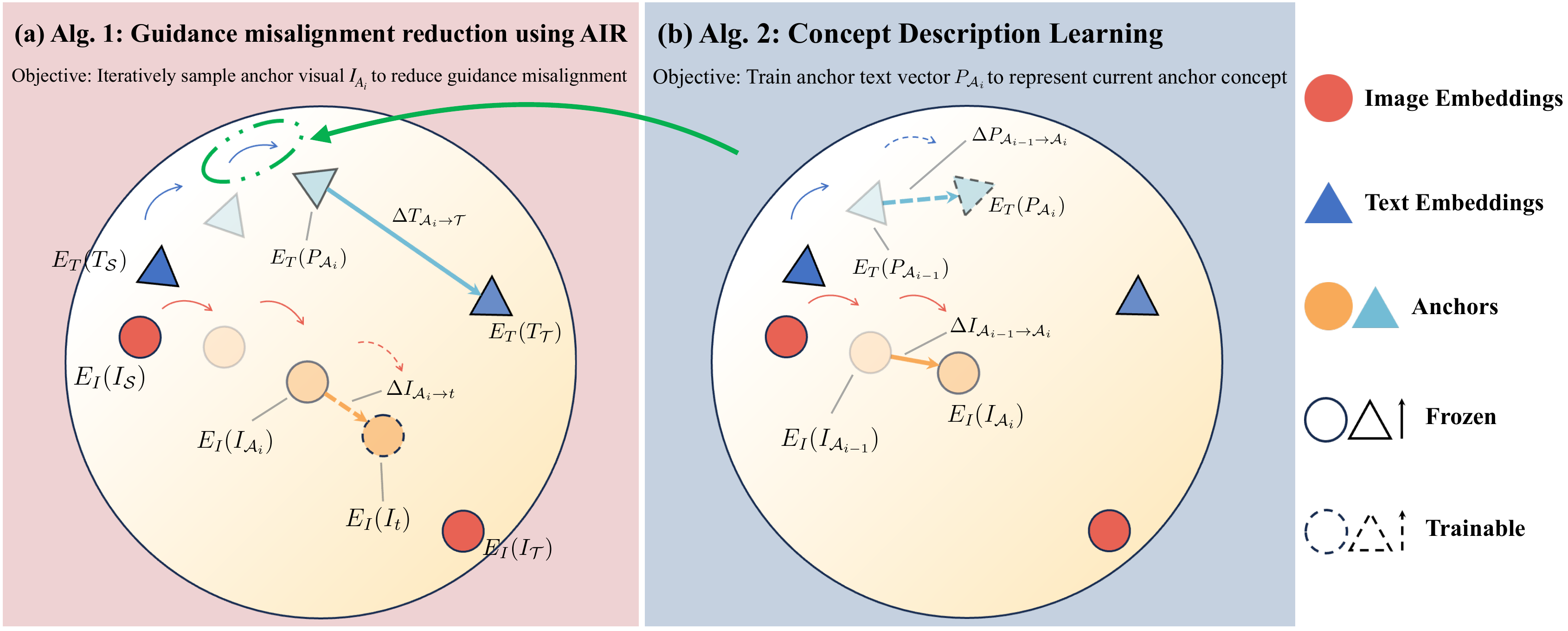}
    \caption{{\bf Illustration of the proposed AIR method:} (a) Guidance misalignment reduction using Adaptation with Iterative Refinement (AIR) in text-driven visual transformation tasks (Sec.~\ref{ssec:Step1}). (b) The proposed concept description learning method to learn anchor concept textual description $P_{\mathcal{A}_i}$ for the anchor $\mathcal{A}_i$ (Sec.~\ref{ssec:Step2}).
    \textit{Symbol descriptions}:
    $E_T$/$E_I$: multimodal contrastive text/image encoder;
    $I_{\mathcal{S}}$/$I_t$/$I_{\mathcal{A}}$: source/transformed/anchor visual;
    $T_{\mathcal{S}}$/$T_{\mathcal{T}}$/$P_{\mathcal{A}}$: source/target/anchor textual descriptions;
    $\Delta I$/$\Delta T$/$\Delta P$: image/text/anchor text offset.
    }
    \label{fig:alg}
\end{figure*}

Our analysis in Sec.~\ref{sec:Empirical study} suggests that closer concepts tend to have less offset misalignment in multimodal contrastive representation space. 
When applied to text-driven visual transformation, this offset misalignment manifests as a mismatch between the text-offset guidance and the desired visual transformation; we refer to this task-specific mismatch as guidance misalignment.
Leveraging this finding, we propose {\em Adaptation with Iterative Refinement (AIR)}, aiming to reduce guidance misalignment by refining text-offset guidance over multiple shorter steps. 
AIR is particularly suitable for text-driven visual transformation tasks, such as zero-shot generative model adaptation and text-guided object-centric image editing, since these tasks are optimized iteratively under directional-loss guidance; reducing guidance misalignment at each step can therefore improve the overall transformation quality (see Sec. \ref{ssec:Impact_of_Offset_Misalignment_on_Generative_Model_Adaptation}).


Specifically, even though the concept distance between source $\mathcal{S}$ and target $\mathcal{T}$ is fixed, 
text-driven visual transformation are inherently iterative: the visual state progressively moves closer to the target concept over the course of optimization (see Sec.~\ref{sec:concept_shifts_during_adaptation}).
Therefore, AIR is based on the idea that intermediate transformed concept can be used as an {\bf anchor} (denoted by $I_{\mathcal{A}}$). AIR computes directional guidance from this anchor to the target. 
This anchor is updated iteratively during the transformation as the transformed visual state moves closer to the target concept. 
Because the anchor-to-target concept distance is smaller than the original source-to-target distance, our analysis suggests that the corresponding guidance misalignment is smaller, leading to more accurate directional guidance and improved transformation quality (see Sec.~\ref{ssec:Impact_of_Offset_Misalignment_on_Generative_Model_Adaptation}). 
One challenge of using $I_{\mathcal{A}}$ for directional loss is that the corresponding textual description $P_\mathcal{A}$ that describes this anchor is unknown. 
In what follows, we first introduce AIR in Sec.~\ref{ssec:Step1}, and then the concept description learning to infer $P_\mathcal{A}$ in Sec.~\ref{ssec:Step2}.

\subsection{Adaptation with Iterative Refinement (AIR)} 
\label{ssec:Step1}

In text-driven visual transformation tasks, existing methods rely on a fixed text offset in Eq.~\ref{eq:Loss_directional} to guide the transformed visual state toward the target concept throughout optimization. In contrast, AIR adopts an iterative refinement strategy that progressively updates the textual guidance during optimization. Specifically, we first optimize the transformation for $t_{\text{thresh}}$ iterations using the current directional loss, allowing the transformed visual state to move closer to target concept. AIR then adaptively refines the textual offset to reduce guidance misalignment for subsequent transformation.


Then, in each $t_{int}$ interval of transformation, we sample the current transformed visual state as a new anchor.\footnote{\scriptsize We use the same settings of $t_{thresh}$ and $t_{int}$ across all 32 experiment setups.} 
For zero-shot GAN and diffusion adaptation, images are sampled from the current transformed generator and encoded by CLIP to obtain the anchor visual representations. For object-centric editing, the current transformed image is directly encoded by CLIP as the visual anchor.
We denote the $i^{th}$ sampled anchor by $I_{\mathcal{A}_i}$.
{\bf To reduce guidance misalignment and provide more accurate guidance, AIR uses the anchor $\mathcal{A}_i$ instead of the original source $\mathcal{S}$ for computing directional loss.}
The proposed AIR scheme is illustrated in Fig.~\ref{fig:alg} (a). 
The image offset with anchor $\mathcal{A}_i$ is computed between the current transformed visual $I_t$ and the anchor visual $I_{\mathcal{A}_i}$:
$\Delta I_{\mathcal{A}_i \rightarrow t}= E_I(I_t) - E_I(I_{\mathcal{A}_i})$.
Assuming that the anchor is described by the textual description $P_{\mathcal{A}_i}$ (details of inferring $P_{\mathcal{A}_i}$ are discussed in Sec.~\ref{ssec:Step2}), the anchor-to-target text offset is calculated as:
$\Delta T_{\mathcal{A}_i \rightarrow \mathcal{T}} = E_T(T_\mathcal{T}) - E_T(P_{\mathcal{A}_i}).$
Finally, the adaptive loss $\mathcal{L}_{adaptive}$ is computed by aligning the image and text offsets from anchor $\mathcal{A}_i$ to target $\mathcal{T}$:
\begin{equation}
    \mathcal{L}_{adaptive} = 1 - \text{cos}(\Delta I_{\mathcal{A}_i \rightarrow t}, \Delta T_{\mathcal{A}_i \rightarrow \mathcal{T}}).
\label{eq:Loss_adaptive}
\end{equation}
By replacing the original source-to-target offset with a closer anchor-to-target offset, $\mathcal{L}_{adaptive}$ reduces the expected guidance misalignment and provides more accurate guidance for the remaining transformation.
We empirically find that adding this adaptive loss to $\mathcal{L}_{direction}$ results in more stable optimization. 
The pseudo-code can be found in 
Sec. \ref{sec:alg}.

\subsection{Learning Anchor Concept Descriptions} \label{ssec:Step2}

Here, we explain the details of the proposed method for learning textual description $P_{\mathcal{A}_i}$ that describes $i^{th}$ anchor concept $\mathcal{A}_i$.
Inspired by \cite{zhou2022coop, teo2024fairqueue},
we define anchor textual description  $P_{\mathcal{A}_i} \in \mathbb{R}^{(M+1) \times d}$ as combination of $M$ learnable tokens $[V]_{j}^{i} \in \mathbb{R}^d$ and a label token $Y_{\mathcal{A}_i} \in \mathbb{R}^d$:
\begin{equation}
    P_{\mathcal{A}_i} = [V]_{1}^{i} [V]_{2}^{i} \dots [V]_{M}^{i} [Y_{\mathcal{A}_i}] 
    \label{eq:prompt}
\end{equation}
Following \cite{zhang2023inclusive}, 
we learn the anchor textual description  $P_{\mathcal{A}_i}$ by aligning text offset to the corresponding image offset.
Here, the image offset is calculated between the current and previous anchors: $\Delta I_{\mathcal{A}_{i-1} \rightarrow \mathcal{A}_{i}} = E_I(I_{\mathcal{A}_i}) - E_I(I_{\mathcal{A}_{i-1}})$.
Similarly, the anchor text offset is calculated as follows:
$\Delta P_{\mathcal{A}_{i-1}
 \rightarrow \mathcal{A}_{i}} = E_T(P_{\mathcal{A}_i}) - E_T(P_{\mathcal{A}_{i-1}})$.
Note that the only trainable parameter is the unknown anchor textual description  $P_{\mathcal{A}_i}$ which is learned by aligning image and anchor text offsets:
\begin{equation}
    \mathcal{L}_{align} = 1 - \text{cos}(\Delta I_{\mathcal{A}_{i-1}
 \rightarrow \mathcal{A}_{i}}, \Delta P_{\mathcal{A}_{i-1}
 \rightarrow \mathcal{A}_{i}})
\label{eq:Loss_align}
\end{equation}
The proposed concept description learning approach is shown in Fig. \ref{fig:alg} (b).
We remark that $P_{\mathcal{A}_{i}}$ is the tokenized anchor textual description before the multimodal contrastive text encoder, and for simplicity, we slightly abuse the notation and use $E_T(P_{\mathcal{A}_{i}})$ to show multimodal contrastive text embedding for anchor $\mathcal{A}_{i}$.

{\bf Remark:} 
Given that offset misalignment is less for closer concepts, we propose to use the previous anchor $\mathcal{A}_{i-1}$ as the source for learning the textual description of the current anchor $\mathcal{A}_{i}$.  Since consecutive anchors are closer, the offset alignment used for learning $P_{\mathcal{A}_i}$ is expected to be more accurate.

{\bf Regularizer:} 
We further propose to use the interpolation between tokenized source and target concept descriptions as anchor label, {\em i.e.}, $Y_{\mathcal{A}_i} = (1 - p_i) Y_\mathcal{S} + p_i Y_\mathcal{T}$, with $p_i$ denoting the proportion of the transformation progress until anchor $\mathcal{A}_i$. The label token acts as a regularizer during concept  description learning (see motivation in 
Sec.~\ref{ssec:Ablation_on_Anchor_Label_Initialization}).

Our experimental results in Sec.~\ref{ssec:Ablation_Study} shows that these design choices lead to better transformation quality with AIR compared to directly learning anchor textual descriptions from the anchor visuals. 
More details of the method are summarized in the pseudo-code in Sec.~\ref{sec:alg}.    
\section{Experiments}
\label{sec:experiments}

In this section, we evaluate AIR on two text-driven visual transformation tasks: zero-shot generative model domain adaptation and text-guided image editing.
We first test whether AIR reduces guidance misalignment where ground-truth visual offsets are available, and then evaluate whether refined guidance improves transformation quality.
We conclude with ablations on AIR's design choices.

\subsection{AIR on Zero-Shot Generative Model Adaptation}
\label{ssec:AIR_on_Zero-Shot_Generative_Model_Domain_Adaptation}

\textbf{Experimental setup.}
We evaluate AIR on ZSGM in both GAN and diffusion settings.
Following prior ZSGM methods \citep{gal2022stylegannada, guo2023ipl, jeon2023svl, Anees2024HyperGANCLIP}, we adapt StyleGAN2-ADA models pretrained on FFHQ, AFHQ-Dog, and Church, as well as Guided Diffusion models pretrained on FFHQ and AFHQ-Dog.
We compare AIR with the SoTA ZSGM methods: NADA \citep{gal2022stylegannada}, IPL \citep{guo2023ipl}, SVL \citep{jeon2023svl}, and HGAN \citep{Anees2024HyperGANCLIP}.
All methods use the same CLIP for guidance.
Dataset, metric, hyperparameter, and implementation details are provided in Sec.~\ref{sec:detailed_experimental_setting}.

\begin{figure*}[t]
    \centering
    \includegraphics[width=\textwidth]{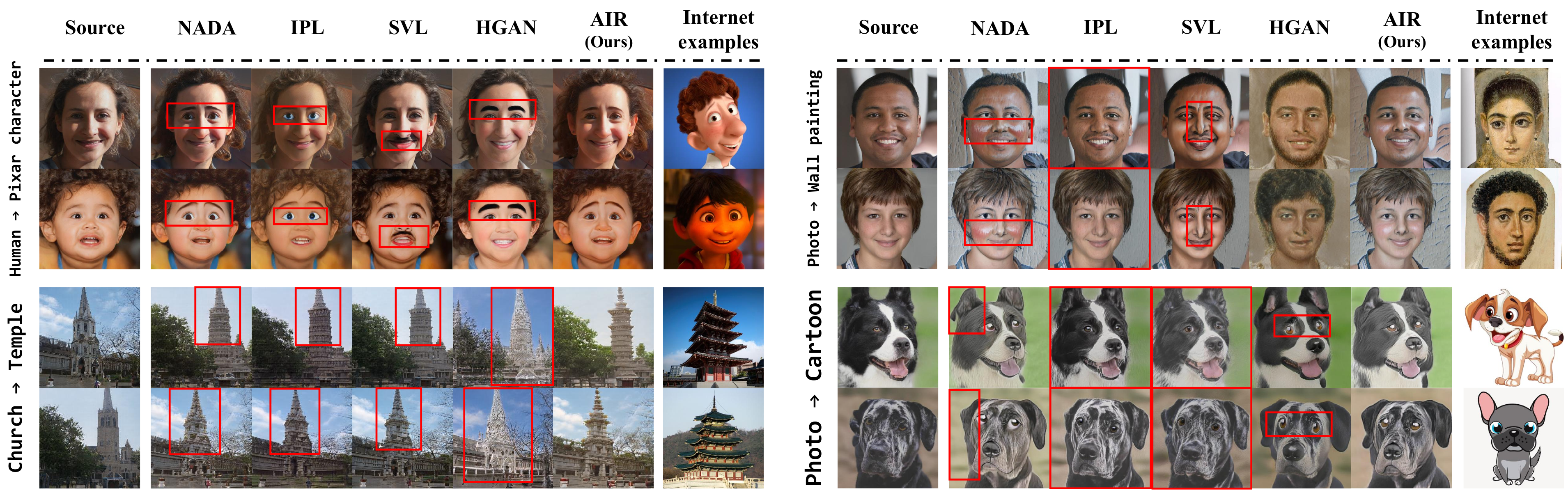}
    \caption{{\bf Qualitative comparison for zero-shot generative model adaptation.} Degraded images or regions are highlighted with \textcolor{red}{red} boxes.
    NADA and HGAN often introduce undesirable changes, such as thick eyebrows in \texttt{Human} $\rightarrow$ \texttt{Pixar character} and red cheeks in \texttt{Photo} $\rightarrow$ \texttt{Wall painting}. 
    IPL and SVL often under-adapt, yielding images that lack target-domain features or styles, particularly for transformations requiring substantial feature changes, such as \texttt{Church} $\rightarrow$ \texttt{Temple} and \texttt{Photo} $\rightarrow$ \texttt{Cartoon}. 
    AIR avoids these artifacts and more faithfully captures target-domain styles, as illustrated by \texttt{Pixar character}, \texttt{Wall painting}, \texttt{Temple}, and \texttt{Cartoon}. 
    Additional qualitative results are provided in Sec.~\ref{ssec:Zero-shot_GAN_Adaptation}. (Best viewed in color and zoomed in.)
    }
    \label{fig:Qualitative}
    
\end{figure*}

\subsubsection{AIR Reduces Guidance Misalignment} \label{ssec:AIR_Reduces_Offset_Misalignment}

We first verify whether AIR reduces guidance misalignment during zero-shot adaptation.
For each adaptation, we compare the directional guidance used by each method with a ground-truth visual offset computed from real target-domain images.
For AIR, we measure the last anchor-to-target direction, matching its refined guidance formulation.
Tab.~\ref{Tab:OM_Alleviation} shows that AIR consistently produces lower offset misalignment, confirming that iterative anchor refinement provides more accurate guidance.
Detailed computation is provided in Sec.~\ref{sec:detailed_experimental_setting}.

\begin{table*}[t]
    \centering
    \caption{AIR reduces guidance misalignment in downstream text-driven visual transformation task. We report guidance misalignment computed between adaptation directions and ground-truth visual offsets. IPL and SVL report multiple directions.}
    \resizebox{0.6\linewidth}{!}{
    \begin{tabular}{@{}l|cccc@{}}
    \toprule
    \multirow{2}{*}{Transformation Setting} & \multicolumn{4}{c}{Guidance Misalignment $\downarrow$} \\
    \cmidrule(l){2-5}
    & NADA & IPL & SVL & AIR \\
    \midrule
    \texttt{Human} $\rightarrow$ \texttt{Baby} & 0.68{\scriptsize$\pm$ 0.01} & 0.70{\scriptsize$\pm$ 0.07} & 0.92{\scriptsize$\pm$ 0.04} & \textbf{0.49}{\scriptsize$\pm$ 0.06} \\
    \texttt{Dog} $\rightarrow$ \texttt{Cat} & 0.54{\scriptsize$\pm$ 0.01} & 0.66{\scriptsize$\pm$ 0.07} & 0.59{\scriptsize$\pm$ 0.03} & \textbf{0.26}{\scriptsize$\pm$ 0.11} \\
    \bottomrule
    \end{tabular}
    \label{Tab:OM_Alleviation}
    }
\end{table*}

\subsubsection{Improved Directional Guidance Leads to Better Adaptation}

We next assess how reduced offset misalignment improves downstream task's performance, i.e., adaptation quality.
Qualitative results in Figs.~\ref{fig:CLIPspace} and~\ref{fig:Qualitative} show that NADA and HGAN often introduce undesirable feature changes, whereas IPL and SVL may under-adapt, yielding samples that lack target-domain features or styles. AIR avoids these common failures and more faithfully captures the target domain; see the figure captions for details.

We present quantitative results in Tab.~\ref{Tab:Quant_GAN} and Tab.~\ref{Tab:Quant_Diff}.
AIR consistently improves target-domain quality measured by CLIP Distance and FID, while maintaining competitive diversity measured by Intra-LPIPS.
We additionally report comparisons of SigLIP and DINOv2 Distances in Sec.~\ref{ssec:Zero-shot_GAN_Adaptation}, and human evaluation in Sec.~\ref{sec:User_study}.
The consistent improvements across CLIP, SigLIP, and DINOv2 Distances, FID, and human evaluation provide complementary evidence that AIR improves adaptation quality.
See detailed metric definitions in Sec.~\ref{ssec:evaluation_details}.

\begin{table*}[ht]
\centering
\caption{Quantitative results of zero-shot GAN adaptation.
For fair comparison, all methods use the same CLIP for guidance.
FID is reported only for \texttt{Baby} and \texttt{Cat}, which have sufficient target-domain samples.
AIR substantially improves adaptation quality (e.g., average CLIP Distance improvements ranging from 9\% to 25\%, and average FID improvements ranging from a minimum of 16\%), while also maintaining competitive diversity measured by Intra-LPIPS.
More qualitative results are shown in Fig.~\ref{fig:CLIPspace}, Fig.~\ref{fig:Qualitative}, and Sec.~\ref{ssec:Zero-shot_GAN_Adaptation}.
}
\resizebox{\linewidth}{!}{
\label{Tab:Quant_GAN}
\begin{tabular}{@{}cc|ccccc|ccccc|ccccc@{}}
\toprule
\multirow{2}{*}{\makecell[c]{Pre-trained \\ Dataset}} & \multirow{2}{*}{Adaptation} & \multicolumn{5}{c|}{CLIP Distance ($\downarrow$)} & \multicolumn{5}{c|}{Intra-LPIPS ($\uparrow$)} & \multicolumn{5}{c}{FID ($\downarrow$)}  \\
\cline{3-7} \cline{8-12} \cline{13-17}
\noalign{\vskip 2pt} & & NADA & IPL & SVL & HGAN & AIR & NADA & IPL & SVL & HGAN & AIR & NADA & IPL & SVL & HGAN & AIR \\
\midrule
\multirow{8}{*}{FFHQ} & 
\texttt{Human} $\rightarrow$ \texttt{Baby} 
& 0.3327 & 0.3562 & 0.3838 & 0.3596 & \textbf{0.3325} 
& 0.4474 & 0.4518 & 0.4506 & 0.4110 & \textbf{0.4520} 
& 68.35 & 68.48 &  158.76 & 123.55 & \textbf{62.13} \\ 
& \texttt{Human} $\rightarrow$ \texttt{Pixar} 
& 0.2335 & 0.2343 & 0.4224 & 0.2418 & \textbf{0.2213} 
& \textbf{0.4759} & 0.4488 & 0.4618 & 0.4013 & 0.4717 
& - & - & - & - & - \\ 
& \texttt{Human} $\rightarrow$ \texttt{Werewolf} 
& 0.3467 & 0.3200 & 0.3998 & 0.3424 & \textbf{0.2431} 
& 0.4301 & 0.4387 & 0.4316 & 0.4395 & \textbf{0.4410} 
& - & - &  - & - & - \\ 
& \texttt{Photo} $\rightarrow$ \texttt{Wall painting} 
& 0.4382 & 0.4898 & 0.4952 & \textbf{0.3747} & 0.4306 
& 0.4217 & 0.4320 & 0.4332 & 0.4208 & \textbf{0.4381} 
& - & - &  - & - & - \\ 
& \texttt{Photo} $\rightarrow$ \texttt{Sketch} 
& 0.3606 & 0.3955 & 0.4092 & 0.3327 & \textbf{0.3126} 
& 0.4190 & 0.4292 & \textbf{0.4476} & 0.4354 & 0.4257 
& - & - &  - & - & - \\ 
& \texttt{Photo} $\rightarrow$ \texttt{Watercolor} 
& 0.3548 & 0.3621 & 0.3639 & 0.3865 & \textbf{0.3376} 
& 0.4598 & \textbf{0.4671} & 0.4612 & 0.4544 & 0.4656
& - & - &  - & - & - \\ 
& \texttt{Photo} $\rightarrow$ \texttt{Ukiyo-e} 
& 0.2437 & 0.2467 & 0.3906 & 0.3286 & \textbf{0.2315} 
& 0.4583 & 0.4652 & 0.4406 & 0.4647 & \textbf{0.4670}
& - & - &  - & - & - \\ 
\midrule
\multirow{7}{*}{Dog} 
& \texttt{Dog} $\rightarrow$ \texttt{Cat} 
& 0.1493 & 0.1530 & 0.1644 & 0.1642 & \textbf{0.1320} 
& 0.4439 & 0.4522 & 0.4547 & 0.4436 & \textbf{0.4628} 
& 70.87 & 83.29 & 65.79 & 88.71 & \textbf{56.20} \\ 
& \texttt{Dog} $\rightarrow$ \texttt{Hamster} 
& 0.1616 & 0.1457 & 0.1826 & 0.1282 & \textbf{0.1306} 
& 0.4196 & \textbf{0.4340} & 0.3918 & 0.3822 & 0.4213
& - & - &  - & - & - \\ 
& \texttt{Dog} $\rightarrow$ \texttt{Capybara} 
& 0.1359 & 0.1446 & 0.1861 & 0.1543 & \textbf{0.1121} 
& 0.4312 & 0.4377 & 0.4264 & 0.4217 & \textbf{0.4401}
& - & - &  - & - & - \\ 
& \texttt{Dog} $\rightarrow$ \texttt{Wolf} 
& 0.1480 & 0.1519 & 0.2249 & \textbf{0.1199} & 0.1421 
& 0.4305 & 0.4261 & 0.4272 & 0.4056 & \textbf{0.4349}
& - & - &  - & - & - \\ 
& \texttt{Dog} $\rightarrow$ \texttt{the Joker} 
& 0.3708 & 0.3827 & 0.4574 & 0.3230 & \textbf{0.3131} 
& 0.4155 & 0.4206 & \textbf{0.4327} & 0.4058 & 0.4173
& - & - &  - & - & - \\ 
& \texttt{Photo} $\rightarrow$ \texttt{Cartoon} 
& 0.2433 & 0.2419 & 0.2543 & 0.2936 & \textbf{0.2258} 
& 0.4356 & 0.4413 & 0.4400 & 0.3741 & \textbf{0.4427}
& - & - &  - & - & - \\ 
& \texttt{Photo} $\rightarrow$ \texttt{Watercolor} 
& 0.1535 & 0.1711 & 0.1646 & 0.1611 & \textbf{0.1507} 
& 0.4639 & \textbf{0.4703} & 0.4622 & 0.4566 & 0.4667 
& - & - &  - & - & - \\ 
\midrule
\multirow{4}{*}{Church} 
& \texttt{Church} $\rightarrow$ \texttt{Skyscraper} 
& 0.3860 & 0.3441 & 0.4209 & 0.3631 & \textbf{0.3270} 
& 0.4823 & 0.481 & 0.4777 & 0.4433 & \textbf{0.4839}
& - & - &  - & - & - \\ 
& \texttt{Church} $\rightarrow$ \texttt{Temple} 
& 0.3197 & 0.3632 & 0.3177 & 0.3229 & \textbf{0.3146} 
& 0.4689 & 0.4623 & 0.4839 & 0.5103 & \textbf{0.4862}
& - & - &  - & - & - \\ 
& \texttt{Photo} $\rightarrow$ \texttt{Sketch} 
& 0.2948 & 0.3070 & 0.3401 & 0.2957 & \textbf{0.2805} 
& 0.5084 & \textbf{0.5373} & 0.5356 & 0.5085 & 0.5176
& - & - &  - & - & - \\ 
& \texttt{Photo} $\rightarrow$ \texttt{Anime} 
& 0.2067 & 0.2180 & 0.3503 & 0.3504 & \textbf{0.1990} 
& 0.5223 & \textbf{0.5392} & 0.5356 & 0.5071 & 0.5245
& - & - &  - & - & - \\ 
\midrule
\multicolumn{2}{c|}{Avg.}
& 0.2711 & 0.2793 & 0.3293 & 0.2802 & \textbf{0.2465} 
& 0.4519 & 0.4575 & 0.4552 & 0.4359 & \textbf{0.4588}
& 69.91 & 75.86 & 112.28 & 106.13 & \textbf{59.16} \\ 

\bottomrule
\end{tabular}
}
\end{table*}


\subsection{AIR on Text-Guided Object-Centric Image Editing}

\textbf{Experimental setup.}
We evaluate AIR on TOIE following the SoTA Style-Editor \cite{park2025styleeditor}.
AIR refines Style-Editor's directional guidance using anchor-based text offsets while leaving the rest of the editing framework unchanged. We evaluate diverse source images and target text descriptions spanning a range of source objects and object-level transformations, including changes in color, material, and artistic style. See more details in Sec.~\ref{sec:detailed_experimental_setting}.

\begin{figure*}[t]
    \centering
    \includegraphics[width=\textwidth]{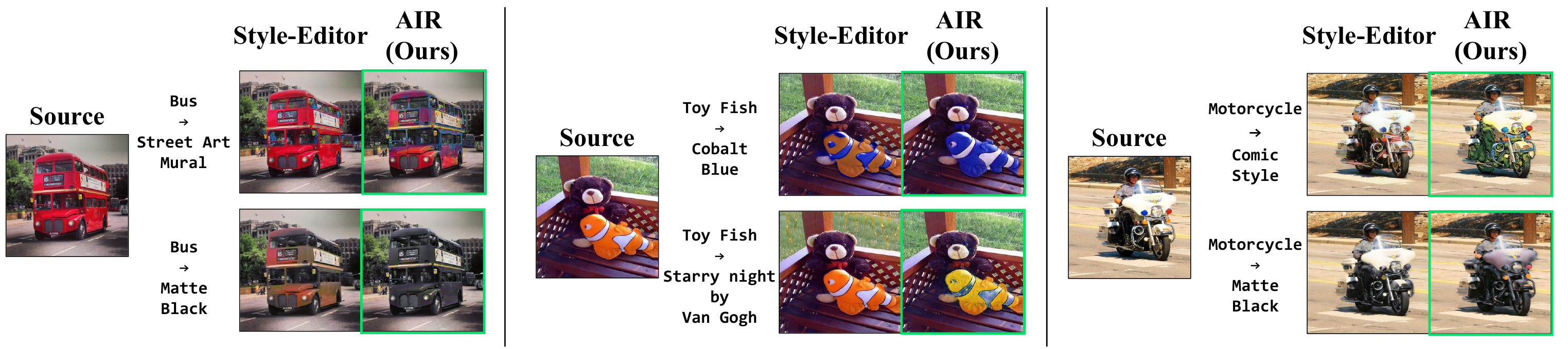}
    \caption{{\bf Qualitative comparison for text-guided object-centric image editing.}
    Compared with Style-Editor, AIR better follows the target text while preserving image regions unrelated to the edit.
    These result indicates that AIR's refined directional guidance improves object-centric editing.
    {\bf (Best viewed with color and zoom in.)}
    }
    \label{fig:Qualitative_imageediting}
    
\end{figure*}

\subsubsection{Improved Directional Guidance Leads to Better Editing}

We evaluate whether AIR improves object-centric editing quality by providing more accurate directional guidance.
As shown in Fig.~\ref{fig:Qualitative_imageediting}, AIR follows the target text more faithfully than Style-Editor while preserving regions unrelated to the edited object.

Following Style-Editor, we report foreground metrics $\mathrm{Sim}_F$ and $\mathrm{Con}_F$, and background metrics $\mathrm{Con}_B$ and $\mathrm{Sty}_B$ (details in Sec.~\ref{ssec:evaluation_details}).
Tab.~\ref{Tab:TGIE_Quant} shows that AIR improves foreground editing quality while maintaining comparable background preservation, indicating that refined guidance improves text-guided object-centric image editing.

\subsection{Validate Learned Anchor Concept Descriptions}
\label{sec:Visualize_Learned_Anchors}

\begin{figure*}[htbp]
    \centering
    \includegraphics[width=0.55\linewidth]{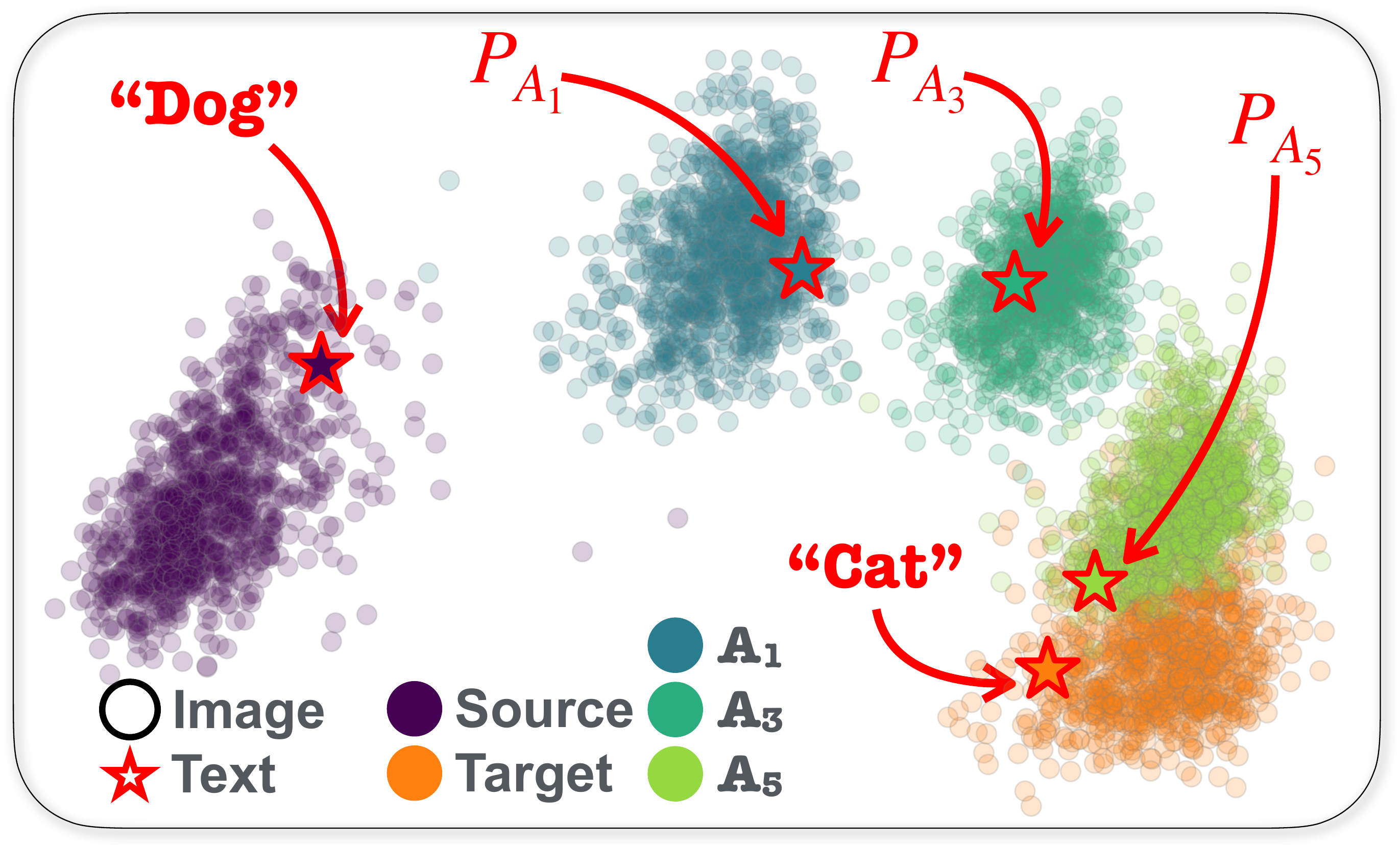}
    \caption{
    {\small PCA visualization of anchor description learning. 
    We use ZSGM setting of {\small \texttt{Dog}} $\rightarrow$ {\small \texttt{Cat}} as an example.
    For each anchor point $\mathcal{A}_i$, our learned concept description $P_{\mathcal{A}_i}$ lies within the distribution of 1000 sampled anchor images $I_{\mathcal{A}_i}$ for that anchor point. This shows that the learned descriptions produced by our concept description learning can accurately represent the
    anchor.}
    }
    \label{fig:PromptLearningVisualization}
\end{figure*}

Additionally, to validate our concept description learning, we visualize the sampled anchor images and the corresponding learned anchor concept descriptions in CLIP space. As shown in Fig.~\ref{fig:PromptLearningVisualization}, the anchor text descriptions learned by our proposed concept description learning accurately represent the anchor images (We show 3 of 5 anchors for presentation clarity).

\subsection{Ablation Study}
\label{ssec:Ablation_Study}

We conduct an ablation study on zero-shot generative model domain adaptation to evaluate the proposed concept description learning strategy. We compare: 
i) $\mathcal{I} \rightarrow \mathcal{T}$: We follow IPL to learn a mapper that produces a concept description for each image. 
ii) $\mathcal{S} \rightarrow \mathcal{A}_{i}$: We learn the concept description by capturing the semantic difference between $\mathcal{S}$ and $\mathcal{A}_{i}$ with the directional loss $\mathcal{L}_{align}^\mathcal{S} = 1 - \text{cos}(\Delta I_{\mathcal{S} \rightarrow \mathcal{A}_{i}}, \Delta P_{\mathcal{S} \rightarrow \mathcal{A}_{i}})$.
iii) $\mathcal{A}_{i-1} \rightarrow \mathcal{A}_{i}$: We use our proposed strategy, which captures the semantic difference between consecutive anchors $\mathcal{A}_{i-1}$ and $\mathcal{A}_{i}$ using the loss in Eq.~\ref{eq:Loss_align}.
The results in Tab.~\ref{Tab:Ablation_PromptLearningScheme} show that our concept description learning strategy produces more accurate textual description and improves zero-shot adaptation,
supporting the use of shorter, less misaligned offsets for learning anchor descriptions.
Our visual ablation results in 
Sec.~\ref{ssec:Visual_Ablation_Studies}
further confirm this observation.
See more of our comprehensive ablation studies in
Sec.~\ref{sec:Add_Ablation}.


\begin{table}[ht]
\centering
\begin{minipage}[t]{0.55\textwidth}
  \centering
  \captionof{table}{Quantitative results of zero-shot diffusion domain adaptation.
  AIR improves adaptation quality while maintaining competitive diversity.
  FID is reported only for \texttt{Baby} and \texttt{Cat}, which have sufficient target-domain samples. See more results in Sec.~\ref{sec:more_experimental_results}.}
  \label{Tab:Quant_Diff}
  \resizebox{\linewidth}{!}{
  \begin{tabular}{@{}cc|cc|cc|cc@{}}
  \toprule
  \multirow{2}{*}{\makecell[c]{Pre-trained \\ Dataset}} & \multirow{2}{*}{Adaptation} & \multicolumn{2}{c|}{CLIP Distance ($\downarrow$)} & \multicolumn{2}{c|}{Intra-LPIPS ($\uparrow$)} & \multicolumn{2}{c}{FID ($\downarrow$)} \\
  \cline{3-8}
  \noalign{\vskip 2pt} & & NADA & AIR & NADA & AIR & NADA & AIR \\
  \midrule
  \multirow{7}{*}{FFHQ}
  & \texttt{Human} $\rightarrow$ \texttt{Baby} & 0.2612 & \textbf{0.2172} & 0.5690 & \textbf{0.5763} & 63.68 & \textbf{58.74} \\
  & \texttt{Human} $\rightarrow$ \texttt{Werewolf} & 0.2777 & \textbf{0.2323} & \textbf{0.5194} & 0.5185 & - & - \\
  & \texttt{Human} $\rightarrow$ \texttt{Pixar character} & 0.4296 & \textbf{0.3865} & \textbf{0.4565} & 0.4558 & - & - \\
  & \texttt{Photo} $\rightarrow$ \texttt{Sketch} & 0.4390 & \textbf{0.3601} & \textbf{0.4873} & 0.4815 & - & - \\
  & \texttt{Photo} $\rightarrow$ \texttt{Wall painting} & 0.4781 & \textbf{0.4744} & 0.5252 & \textbf{0.5270} & - & - \\
  & \texttt{Photo} $\rightarrow$ \texttt{Watercolor} & 0.3264 & \textbf{0.3229} & 0.6224 & \textbf{0.6399} & - & - \\
  & \texttt{Photo} $\rightarrow$ \texttt{Ukiyo-e} & 0.2307 & \textbf{0.2152} & 0.5431 & \textbf{0.5524} & - & - \\
  \midrule
  \multirow{7}{*}{Dog}
  & \texttt{Dog} $\rightarrow$ \texttt{Cat} & 0.1434 & \textbf{0.1428} & 0.5411 & \textbf{0.5494} & 84.75 & \textbf{77.15} \\
  & \texttt{Dog} $\rightarrow$ \texttt{Wolf} & 0.1460 & \textbf{0.1337} & 0.4649 & \textbf{0.4735} & - & - \\
  & \texttt{Dog} $\rightarrow$ \texttt{Hamster} & 0.1851 & \textbf{0.1588} & 0.4854 & \textbf{0.4952} & - & - \\
  & \texttt{Dog} $\rightarrow$ \texttt{Capybara} & 0.1576 & \textbf{0.1204} & \textbf{0.4832} & 0.4604 & - & - \\
  & \texttt{Dog} $\rightarrow$ \texttt{the Joker} & 0.3955 & \textbf{0.3208} & 0.4571 & \textbf{0.4663} & - & - \\
  & \texttt{Photo} $\rightarrow$ \texttt{Cartoon} & 0.2516 & \textbf{0.2481} & 0.5572 & \textbf{0.5611} & - & - \\
  & \texttt{Photo} $\rightarrow$ \texttt{Watercolor} & 0.1932 & \textbf{0.1853} & 0.5173 & \textbf{0.5322} & - & - \\
  \midrule
  \multicolumn{2}{c|}{Avg.} & 0.2797 & \textbf{0.2513} & 0.5164 & \textbf{0.5207} & 74.22 & \textbf{67.95} \\
  \bottomrule
  \end{tabular}}
\end{minipage}%
\hspace{0.2cm}
\begin{minipage}[t]{0.40\textwidth}
    \centering
    \captionof{table}{Quantitative results of text-guided object-centric image editing.}
    \label{Tab:TGIE_Quant}
    \vspace{0.3cm}
    \resizebox{\linewidth}{!}{
    \begin{tabular}{@{}c|cc|cc@{}}
    \toprule
    \multirow{2}{*}{Methods} & \multicolumn{2}{c|}{Foreground quality metrics} & \multicolumn{2}{c}{Background quality metrics} \\
    \cline{2-5}
    \noalign{\vskip 2pt}
    & $\mathrm{Sim}_F \uparrow$ & $\mathrm{Con}_F \downarrow$ & $\mathrm{Con}_B \downarrow$ & $\mathrm{Sty}_B \downarrow$ \\
    \midrule
    Style-Editor & 0.27 & 4.58 & \textbf{1.10} & \textbf{0.10} \\
    AIR (ours) & \textbf{0.39} & \textbf{3.22} & 1.11 & \textbf{0.10} \\
    \bottomrule
    \end{tabular}}

    \vspace{1cm}
    \captionof{table}{Ablation study on concept description learning strategy.}
    \label{Tab:Ablation_PromptLearningScheme}
    \resizebox{\linewidth}{!}{
    \begin{tabular}{@{}c|cc|cc@{}}
    \toprule
    \multirow{2}{*}{Methods} & \multicolumn{2}{c|}{\texttt{Human} $\rightarrow$ \texttt{Baby}} & \multicolumn{2}{c}{\texttt{Dog} $\rightarrow$ \texttt{Cat}} \\
    \cline{2-5}
    \noalign{\vskip 2pt} & FID ($\downarrow$) & Intra-LPIPS ($\uparrow$) & FID ($\downarrow$) & Intra-LPIPS ($\uparrow$) \\
    \midrule
    NADA & 68.35 & 0.4474 & 70.87 & 0.4439 \\
    $\mathcal{I} \rightarrow \mathcal{T}$ & 98.35 & 0.4308 & 104.59 & 0.4452 \\
    $\mathcal{S} \rightarrow \mathcal{A}_{i}$ & 64.39 & 0.4503 & 61.75 & \textbf{0.4630} \\
    $\mathcal{A}_{i-1} \rightarrow \mathcal{A}_{i}$ & \textbf{62.13} & \textbf{0.4520} & \textbf{56.20} & 0.4628 \\
    \bottomrule
    \end{tabular}}
\end{minipage}
\end{table}

\section{Conclusion}
In this work, we revisit the widely adopted assumption of image-text offset alignment in multimodal contrastive representation spaces. 
Through comprehensive empirical analysis, we reveal noticeable image-text offset misalignment and show that it has a meaningful positive correlation with semantic concept distance across multiple datasets and contrastive vision-language models. 
Motivated by this finding, we propose AIR, an iterative refinement framework that reduces the downstream effect of offset misalignment by replacing a single source-to-target text offset with refined anchor-to-target guidance during text-driven visual transformation. 
Extensive experiments on zero-shot generative model domain adaptation and text-guided image editing demonstrate that AIR improves directional guidance and consistently enhances transformation quality, achieving state-of-the-art performance in both tasks. 





%
%
\clearpage
\section*{Acknowledgment}

This research is supported by Temasek Laboratories, Singapore University of Technology and Design; the National Research Foundation, Singapore under its
AI Singapore Programmes (AISG Award No.: AISG2-TC-2022-007); the Agency for Science, Technology and Research (A*STAR) under its MTC Programmatic
Funds (Grant No. M23L7b0021); the National Research Foundation, Singapore and Infocomm Media Development Authority under its Trust Tech Funding Initiative. Any opinions, findings and conclusions or recommendations expressed in this material are those of the author(s) and do not reflect the views of National Research Foundation, Singapore and Infocomm Media Development Authority.
\bibliographystyle{unsrt}
\bibliography{main}

\newpage
\appendix
\addtocontents{toc}{\protect\setcounter{tocdepth}{3}}
\setcounter{secnumdepth}{4}
\setcounter{figure}{6} 
\setcounter{table}{5} 


\section*{Supplementary} \label{sec:supp}
\noindent
In this supplementary material, we provide additional experiments, ablation studies, and reproducibility details to support our findings.
These sections are not included in
the main paper due to space constraints.
In particular:

\begin{itemize}
\item
Sec.~\ref{sec:more_details_and_results_of_empirical_analysis} expands our offset misalignment study with additional multimodal contrastive models, for a total of \textbf{8 different multimodal contrastive models} considered in the whole paper.
Across all multimodal contrastive models,
we consistently observe
a meaningful positive correlation between image-text offset misalignment and concept distance.

\item
Sec.~\ref{sec:detailed_experimental_setting} reports extensive experimental results
on zero-shot generative model adaptation and text-guided object-centric image editing,
bringing the total number of adaptation settings evaluated in this work to \textbf{40}.
Across all settings,
we
consistently observe that our proposed AIR achieves state-of-the-art performance.
These results support our claim that reducing image-text offset misalignment improves directional guidance and downstream transformation quality.
\end{itemize}
Codes are available in the supplementary files.



{
\hypersetup{linkcolor=black}
\tableofcontents
}

\newpage

\section{More Details and Results of Empirical Analysis}
\label{sec:more_details_and_results_of_empirical_analysis}

\subsection{Details of Empirical Analysis}
\label{ssec:Detail_of_empirical_study}
For datasets with a single class label for each image, such as ImageNet, Caltech-101, and CIFAR-100, we use the original images from the dataset. For datasets with multiple objects in an image, such as OpenImages, MS COCO, and Visual Genome, to better align with the setting in NADA, we extract the objects using bounding boxes and classify them into their labeled classes.

For a certain concept $\alpha$, we use the images of the class as $I_\alpha$. 
For text description $T_\alpha$, we use the corresponding class label with INt, e.g., "a photo of a [cat]" when $\alpha=$ \texttt{cat}.

\subsection{Full Empirical Analysis Results}
\label{sec:offset_misalignment_in_more_clip_space}

\begin{figure*}[p]
    \centering
    \includegraphics[width=\textwidth]{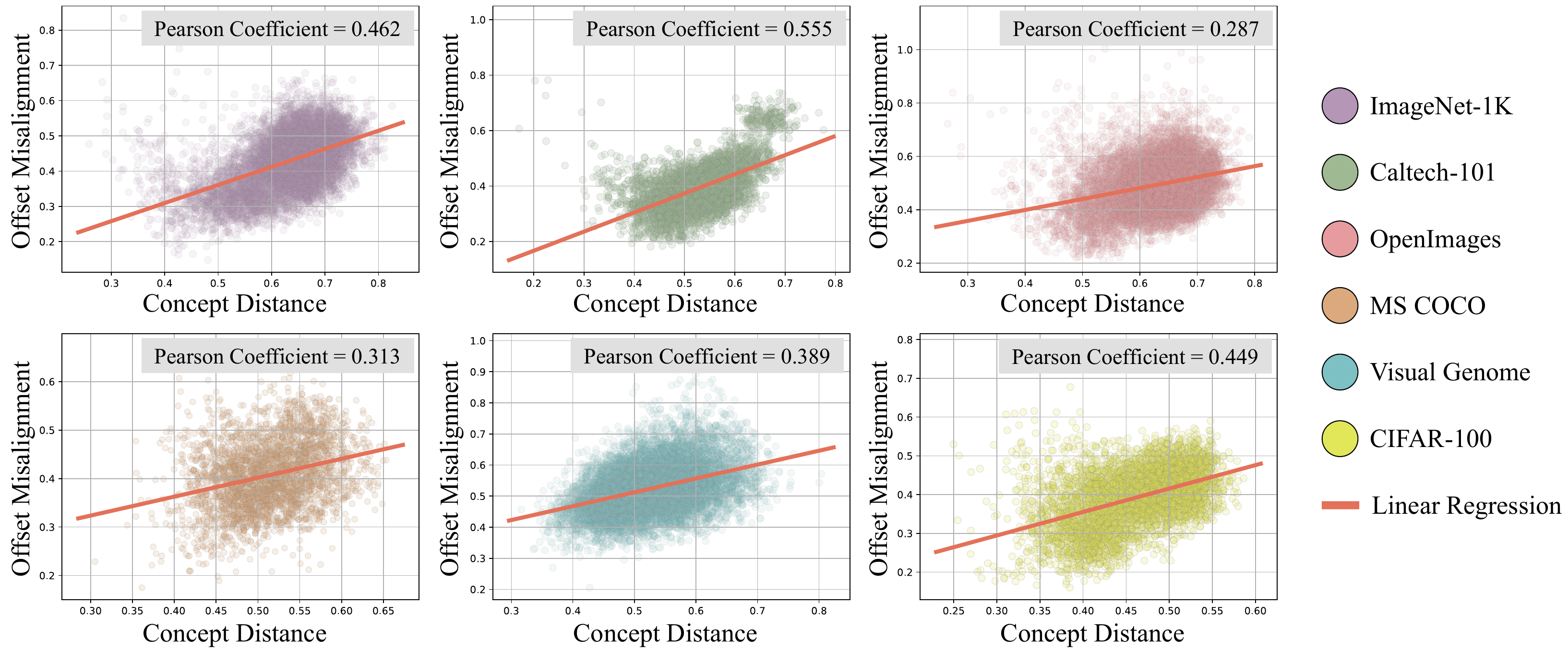}
    \caption{Empirical analysis of offset misalignment in CLIP ConvNext-L space. Experiment setup is the same as 
    Sec.~\ref{ssec:OM} 
    except that CLIP ConvNext-L is used as the multimodal contrastive representation space. Our results
    show that the {\bf meaningful correlation} (measured by Pearson coefficient \cite{pearsoncorr}) between offset misalignment and concept distance consistently exists in various multimodal contrastive representation spaces.}
    \label{fig:Emprical_ConvNext}
\end{figure*}

\begin{figure*}[p]
    \centering
    \includegraphics[width=\textwidth]{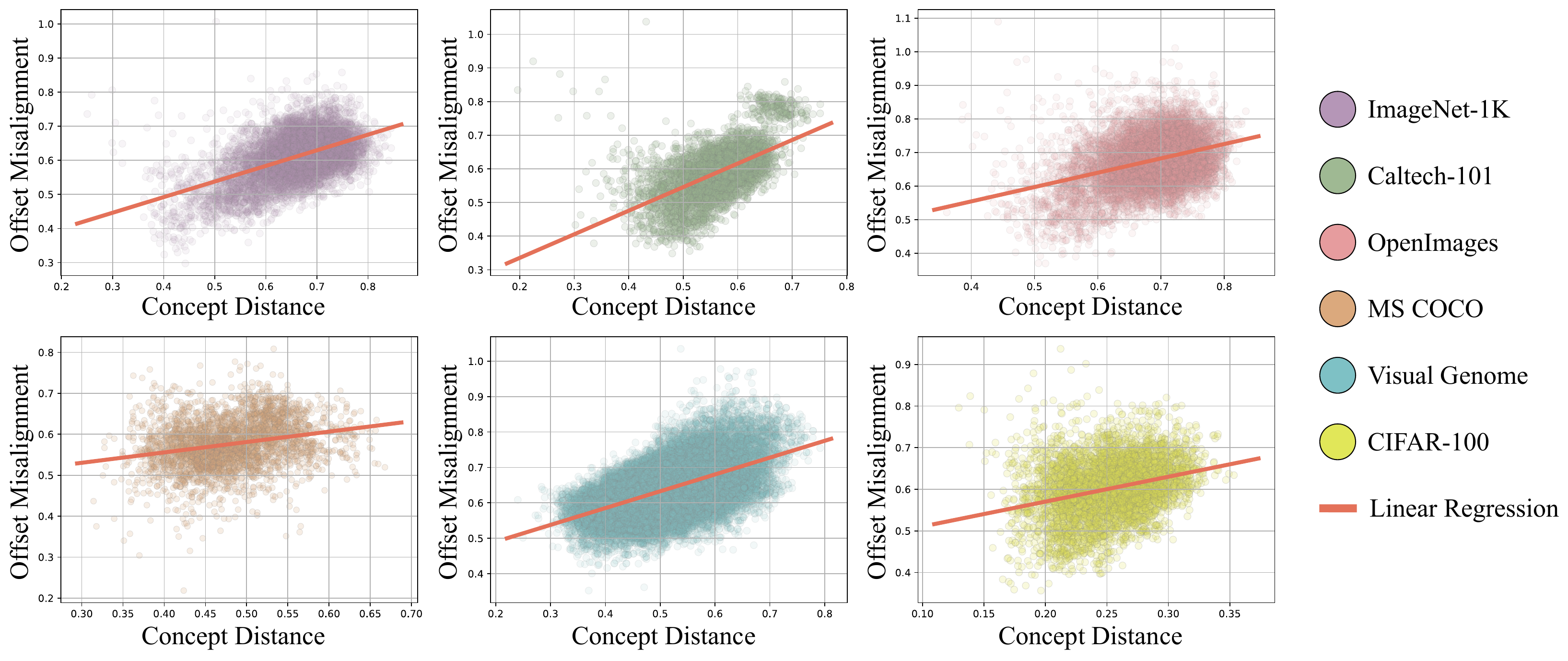}
    \caption{Empirical analysis of offset misalignment in CLIP RN50x64 space. Experiment setup is the same as 
    Sec.~\ref{ssec:OM}
    except that CLIP RN50x64 is used as the multimodal contrastive representation space. Our results
    show that the {\bf meaningful correlation} (measured by Pearson coefficient \cite{pearsoncorr}) between offset misalignment and concept distance consistently exists in various multimodal contrastive representation spaces.}
    \label{fig:Emprical_RN}
\end{figure*}

\begin{figure*}[p]
    \centering
    \includegraphics[width=\textwidth]{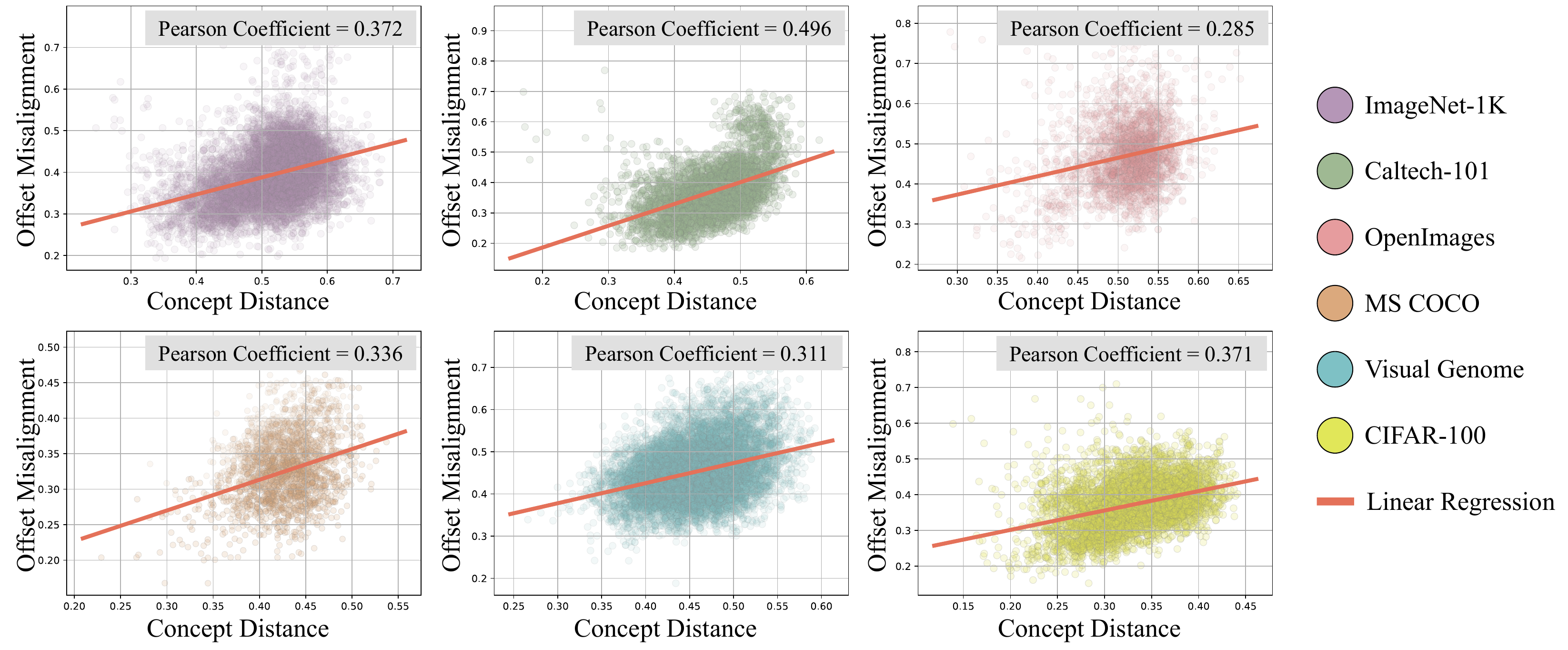}
    \caption{Empirical analysis of offset misalignment in SigLIP ViT-L/16-256 space. Experiment setup is the same as 
    Sec.~\ref{ssec:OM}
    except that SigLIP ViT-L/16-256 is used as the multimodal contrastive representation space. Our results
    show that the {\bf meaningful correlation} (measured by Pearson coefficient \cite{pearsoncorr}) between offset misalignment and concept distance consistently exists in various multimodal contrastive representation spaces.}
    \label{fig:Emprical_Siglip}
\end{figure*}

\begin{figure*}[p]
    \centering
    \includegraphics[width=\textwidth]{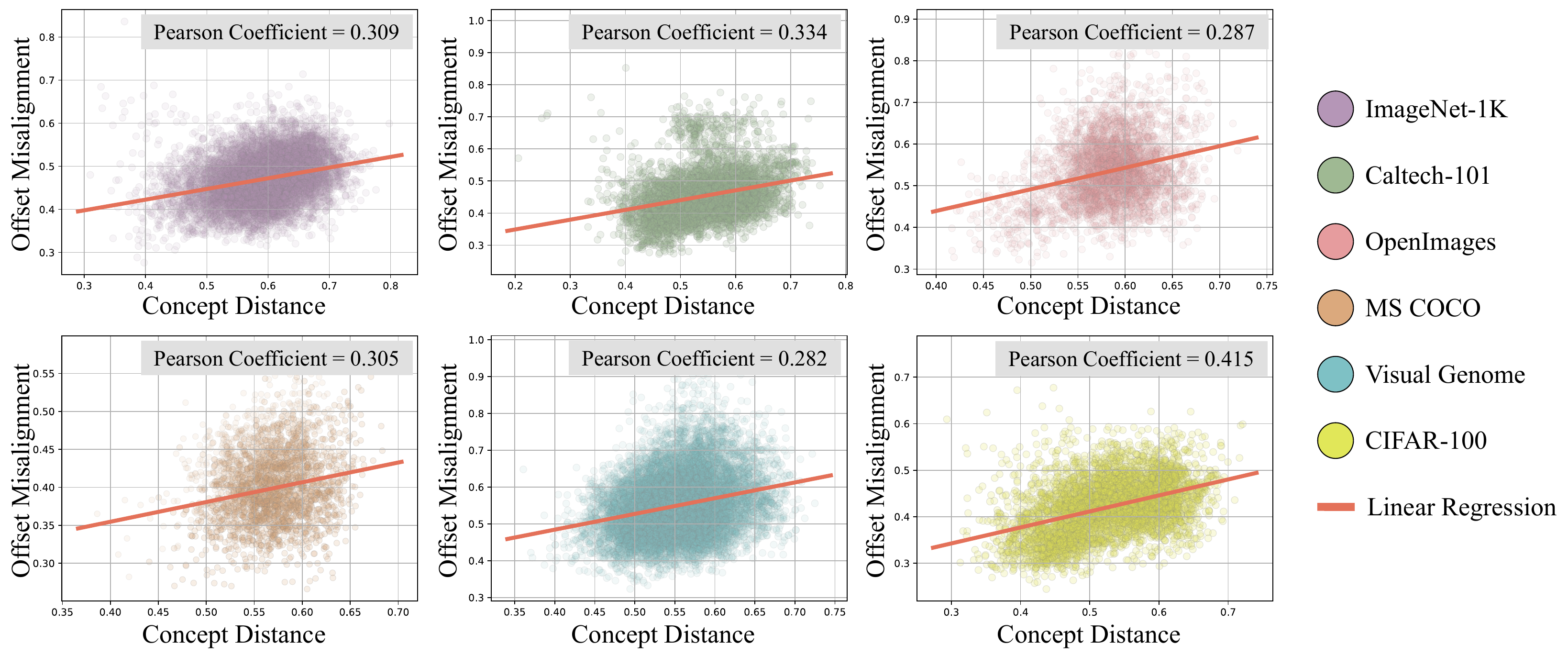}
    \caption{Empirical analysis of offset misalignment in EVA-CLIP-8B space. Experiment setup is the same as 
    Sec.~\ref{ssec:OM}
    except that EVA-CLIP-8B is used as the multimodal contrastive representation space. Our results
    show that the {\bf meaningful correlation} (measured by Pearson coefficient \cite{pearsoncorr}) between offset misalignment and concept distance consistently exists in various multimodal contrastive representation spaces.}
    \label{fig:Emprical_EvaCLIP}
\end{figure*}

\begin{figure*}[p]
    \centering
    \includegraphics[width=\textwidth]{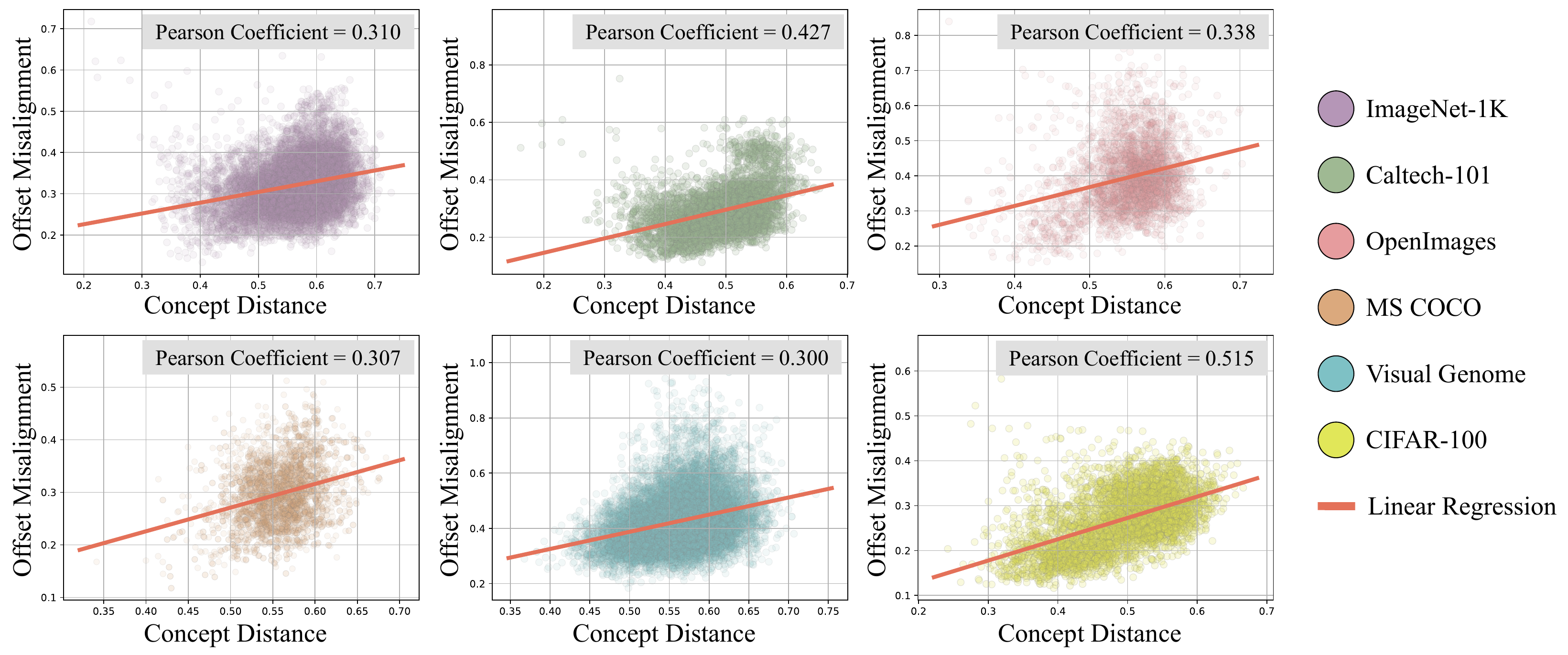}
    \caption{Empirical analysis of offset misalignment in DFN5B-CLIP space. Experiment setup is the same as 
    Sec.~\ref{ssec:OM}
    except that DFN5B-CLIP is used as the multimodal contrastive representation space. Our results
    show that the {\bf meaningful correlation} (measured by Pearson coefficient \cite{pearsoncorr}) between offset misalignment and concept distance consistently exists in various multimodal contrastive representation spaces.}
    \label{fig:Emprical_DFN}
\end{figure*}

\begin{figure*}[p]
    \centering
    \includegraphics[width=\textwidth]{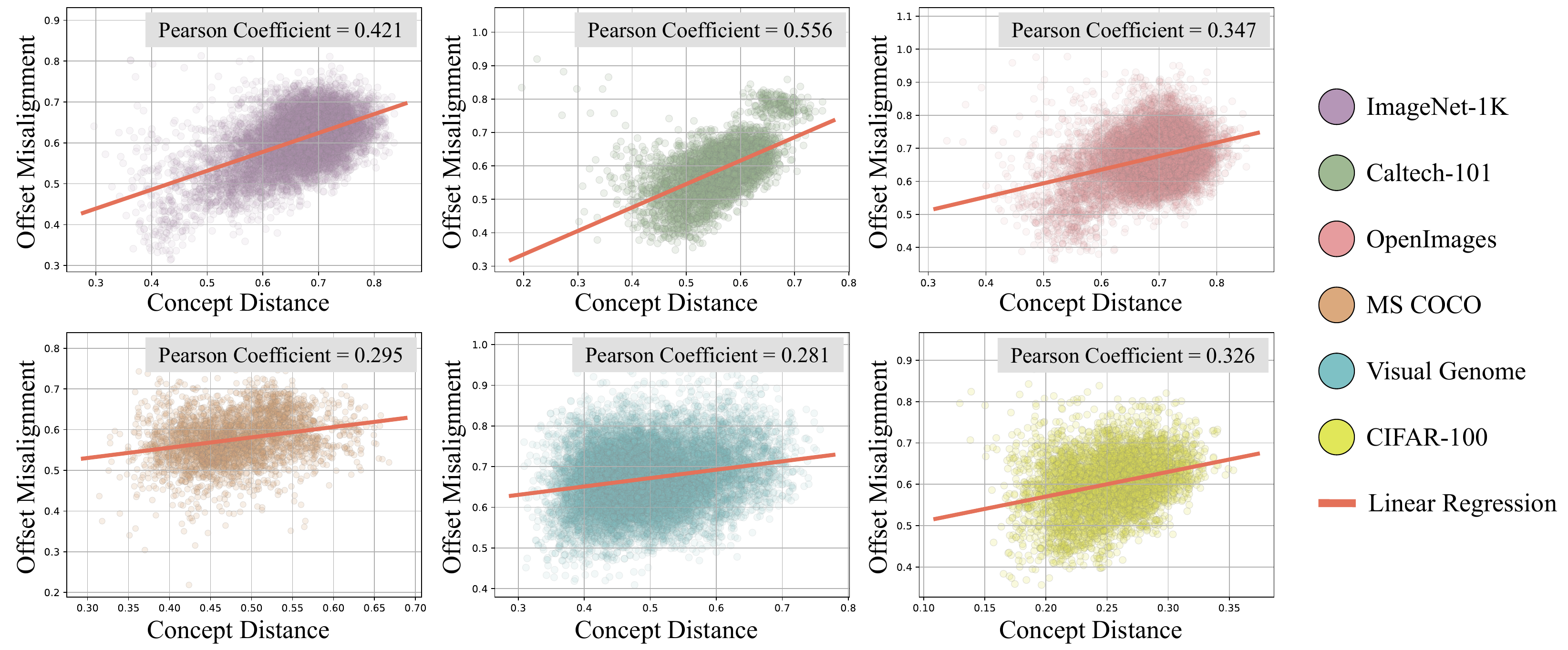}
    \caption{Empirical analysis of offset misalignment in MobileCLIP space. Experiment setup is the same as 
    Sec.~\ref{ssec:OM}
    except that MobileCLIP is used as the multimodal contrastive representation space. Our results
    show that the {\bf meaningful correlation} (measured by Pearson coefficient \cite{pearsoncorr}) between offset misalignment and concept distance consistently exists in various multimodal contrastive representation spaces.}
    \label{fig:Emprical_MetaCLIP}
\end{figure*}

In this section, we present an additional empirical analysis of offset misalignment for {\bf 6 additional contrastive learning-based multimodal representation spaces}. Following the experimental setup in 
Sec. \ref{sec:Empirical study} and in addition to 
CLIP ViT-B/32 \cite{radford2021CLIP} and SigLIP SoViT-400m \cite{zhai2023sigmoid},
we use CLIP ConvNext-L, CLIP RN50x64, SigLIP ViT-L/16-256  \cite{zhai2023sigmoid}, EVA-CLIP-8B \cite{evaclip}, DFN5B-CLIP \cite{DFNCLIP}, and MobileCLIP \cite{faghrimobileclip2} as multimodal contrastive representation spaces. We plot offset misalignment against concept distance for six public datasets in Fig.~\ref{fig:Emprical_ConvNext}--\ref{fig:Emprical_MetaCLIP}. Our results demonstrate consistent positive correlations between offset misalignment and concept distance across different multimodal contrastive representation spaces.


One possible explanation for the consistent and meaningful positive correlation  
lies in the geometry of multimodal contrastive representation spaces. Contrastive models align individual image-text pairs but do not explicitly enforce relational consistency between concepts. Consequently, semantically close concepts, which often lie within locally smooth regions of the representation manifold, tend to yield more consistent transformation directions across modalities, whereas larger concept shifts span broader regions of the manifold where curvature and modality-specific distortions may lead to greater misalignment. Understanding the factors underlying this phenomenon is an interesting direction for future work.



\newpage

\clearpage

\section{More Details of AIR}
\label{sec:more_details_of_air}

\subsection{Algorithm}
\label{sec:alg}

We provide a figure illustration corresponding pseudo-code for applying AIR to text-driven visual transformation in Fig.~\ref{fig:alg}(a) and Alg.~\ref{alg:AIR}, and for the proposed concept description learning module in Fig.~\ref{fig:alg}(b) and Alg.~\ref{alg:prompt_learning}. We use $I_{\mathcal{S}}(z)$ and $I_t(z;\theta_t)$ to denote the source visual state and the transformed visual state at iteration $t$, respectively, where $\theta_t$ denotes the task-specific trainable parameters. For zero-shot generative model domain adaptation, the visual states correspond to $G_{\mathcal{S}}(w)$ and $G_t(w)$ generated using the same latent code $w$. In the GAN setting, $\theta_0$ is initialized from the pretrained source generator. In the diffusion setting, the pretrained denoiser remains frozen and $\theta_t$ denotes the trainable LoRA adapter parameters. For text-guided object-centric image editing, the visual states correspond to the source and current edited images, $\theta_t$ denotes the StyleNet parameters optimized for the given source image following Style-Editor, and $z$ can be omitted. 

We initialize the anchor sequence with the source concept, {\em i.e.}, \(I_{\mathcal{A}_0}(z)=I_{\mathcal{S}}(z)\) and \(P_{\mathcal{A}_0}=T_{\mathcal{S}}\). Thus, the first intermediate anchor is learned relative to the original source, while each subsequent anchor is learned relative to its preceding anchor.


\begin{algorithm*}[!htbp]
	\DontPrintSemicolon
	\SetAlgoLined	

\KwRequire{Source visual state $I_{\mathcal{S}}(z)$, task-specific transformation producing $I_t(z;\theta_t)$, textual descriptions $T_{\mathcal{S}}$ and $T_{\mathcal{T}}$, $t_{adapt}$, $t_{thresh}$, $t_{int}$,
learning rate $\eta$, and multimodal contrastive image and text encoders $E_I$ and $E_T$
}

\KwOut{
Optimized transformed visual state $I_{t_{adapt}}(z;\theta_{t})$
}

Initialize $\theta_0$ following the task-specific baseline and keep $I_{\mathcal{S}}$ fixed, $i=0$, $\mathcal{L}_{adaptive} = 0$

$\Delta T_{\mathcal{S} \rightarrow \mathcal{T}} = E_T(T_{\mathcal{T}}) - E_T(T_{\mathcal{S}})$

\For{$t = 0$; $t{+}{+}$; $t < t_{adapt}$} 
{

$\Delta I_{\mathcal{S} \rightarrow t} = E_I(I_t(z;\theta_t)) - E_I(I_{\mathcal{S}}(z))$

$\mathcal{L}_{direction} = 1 - \text{cos}(\Delta I_{\mathcal{S} \rightarrow t}, \Delta T_{\mathcal{S} \rightarrow \mathcal{T}})$

\If{$t \% t_{int} = 0$}
{
$i++$

$I_{\mathcal{A}_i}(z)=I_t(z;\theta_t)$ 

$P_{\mathcal{A}_i}=$ Concept-Description-Learning $(I_{\mathcal{A}_i}, I_{\mathcal{A}_{i-1}}, P_{\mathcal{A}_{i-1}})$  \tcc{\small refer to Alg.~\ref{alg:prompt_learning} for details} 

}

\If{$t > t_{thresh}$}
{

$\Delta I_{\mathcal{A}_i \rightarrow t}= E_I(I_t(z;\theta_t)) - E_I(I_{\mathcal{A}_i}(z))$ \tcc{\small if $I_t=I_{\mathcal{A}_i}$, apply the task-specific perturbation to $I_t$} 

$\Delta T_{\mathcal{A}_i \rightarrow \mathcal{T}} = E_T(T_\mathcal{T}) - E_T(P_{\mathcal{A}_i})$

$\mathcal{L}_{adaptive} = 1 - \text{cos}(\Delta I_{\mathcal{A}_i \rightarrow t}, \Delta T_{\mathcal{A}_i \rightarrow \mathcal{T}})$
}

$\mathcal{L} = \mathcal{L}_{baseline} + \mathcal{L}_{adaptive}$


Update $\theta_{t+1} \leftarrow \theta_t - \eta \nabla_{\theta_t} \mathcal{L}$
}

\caption{AIR for Text-Driven Visual Transformation}
\label{alg:AIR}
\end{algorithm*}


\begin{algorithm*}[!htbp]
	\DontPrintSemicolon
	\SetAlgoLined	

\KwRequire{Current and previous anchor visual states $I_{\mathcal{A}_i}$ and $I_{\mathcal{A}_{i-1}}$, learned concept description for the previous anchor $P_{\mathcal{A}_{i-1}}$,
learning rate $\mu$, and multimodal contrastive image and text encoders $E_I$ and $E_T$
}

\KwOut{
Textual description $P_{\mathcal{A}_i}$ to represent the current anchor concept.
}


$\Delta I_{\mathcal{A}_{i-1} \rightarrow \mathcal{A}_{i}} = E_I(I_{\mathcal{A}_i}(z)) - E_I(I_{\mathcal{A}_{i-1}}(z))$

\For{$k = 0$; $k{+}{+}$; $k < k_{iter}$} 
{
$\Delta P_{\mathcal{A}_{i-1}
 \rightarrow \mathcal{A}_{i}} = E_T(P_{\mathcal{A}_i}) - E_T(P_{\mathcal{A}_{i-1}})$.
 
$\mathcal{L}_{align} = 1 - \text{cos}(\Delta I_{\mathcal{A}_{i-1} \rightarrow \mathcal{A}_{i}}, \Delta P_{\mathcal{A}_{i-1}  \rightarrow \mathcal{A}_{i}})$

Update $P_{\mathcal{A}_i} \leftarrow P_{\mathcal{A}_i} - \mu \nabla_{P_{\mathcal{A}_i}} \mathcal{L}_{align}$
}

\caption{Concept Description Learning for Visual Anchors}
\label{alg:prompt_learning}
\end{algorithm*}

\section{More Details and Results of Experiments}
\label{sec:detailed_experimental_setting}

\subsection{Details of AIR Implementation}
\label{ssec:Details_of_AIR_Implementation}

In this work, we implement zero-shot generative model adaptation for both GANs and diffusion models. 
The implementation details for each type of model is as follows:
\begin{itemize}
    \item {\bf Zero-Shot Adaptation of GANs.} We follow previous ZSGM works \citep{gal2022stylegannada, guo2023ipl, jeon2023svl} setups to adapt StyleGAN2-ADA \citep{karras2020ada} pre-trained 
    on FFHQ \citep{karras2019style} and AFHQ-Dog \citep{choi2020starganv2} to various target domains. 
    \item {\bf Zero-Shot Adaptation of Diffusion Models.} 
     We use Guided Diffusion \citep{dhariwal2021diffusion} pre-trained on FFHQ and AFHQ-Dog from P2-Weighting \citep{choi2022perception} as source generator. To speedup training, we use DPM-Solver \citep{lu2022dpm} for 10-step image generation. To prevent overfitting, we keep all pretrained denoiser weights frozen and optimize only the LoRA adapter parameters \citep{hu2022lora}.


\end{itemize}
During the adaptation of both generators, we utilize the pre-trained ViT-B/32 
as vision encoder for CLIP. 
Hyperparameter details can be found in 
Sec.~\ref{sec:Hyperparameters_of_AIR}.
{\em Notably, the only varying hyperparameter for all adaptation setups is the total number of adaptation iterations (strictly follow NADA).}

Note that none of the four baselines NADA/IPL/SVL/HyperGAN-CLIP of ZSGM provides an official implementation for diffusion models. We therefore implemented NADA for the diffusion setting on generative domain adaptation, while the other three methods cannot be directly transferred for the diffusion setting without substantial methodological redesign. This is because:

\begin{itemize}
    \item IPL is not generator-agnostic, as it learns a mapper from StyleGAN style codes and therefore relies on the generator architecture and latent-space structure of StyleGAN. 
    \item HyperGAN-CLIP trains a hypernetwork to modulate StyleGAN style layers, making its formulation specific to StyleGAN-based generators.
    \item SVL also relies on a parameter-importance estimation which is designed for a one-pass GAN generator. In diffusion models, however, the same denoising network is repeatedly applied across different noise levels, and the importance of a parameter may vary substantially across timesteps. This behavior cannot be captured by directly applying SVL’s output-image-based importance estimation. 
\end{itemize}

Consequently, extending IPL, SVL, or HyperGAN-CLIP to diffusion models would require nontrivial new designs beyond a straightforward reimplementation.

For implementation of text-guided object-centric image editing, we follow the SoTA Style-Editor \citep{park2025styleeditor}.
AIR is applied by refining Style-Editor's directional guidance with anchor-based text offsets while keeping the remaining editing framework unchanged

\subsection{Details of AIR Evaluation}
\label{ssec:evaluation_details}
\textbf{Zero-Shot Generative Model Adaptation} A well-trained image generator is defined by its ability to produce high-quality and diverse images from target distribution. We follow existing zero-shot works in evaluation setup when applicable, and further improve on them. Specifically, following previous works \citep{gal2022stylegannada, guo2023ipl, jeon2023svl}, we have conducted comparisons on both public datasets and images collected from the internet.
Our evaluations include both visual inspections for qualitative evaluations and quantitative evaluations using the following metrics:

\begin{itemize}
    \item {\bf FID.} For target domains with large and publicly available datasets, we follow previous work \citep{jeon2023svl} to use FFHQ-Baby \citep{ojha2021cdc} (for target domain \texttt{Baby}), and AFHQ-Cat \citep{choi2020starganv2} (for target domain \texttt{Cat}) as target distribution. Then, we generate 5000 samples for each target domain \citep{zhao2022adam, zhao2023rick}, and use FID to evaluate the quality and diversity of domain adaptation.

    \item {\bf CLIP Distance.}
    The public data is scarce for other target domains, {\em e.g.,} \texttt{Pixar}. 
    For these domains, we follow IPL's idea \citep{guo2023ipl} to collect internet images as reference. However, since IPL did not make the collected samples publicly available, we had to repeat the same practice and collect the images.
    Then, we use the CLIP Distance \citep{gal2023an} which is defined as the cosine distance between the CLIP embeddings of the collected samples and the generated samples to measure the similarity of the generated samples to the target domain.
    Note that these collected target images are never exposed during adaptation, so the evaluation does not introduce target-image leakage into the optimization.
    Since no text is involved in this evaluation, the metric cannot be improved through trivial alignment with the target text.

    \item {\bf SigLIP/DINO Distance.}
    We further evaluate the quality of the domain adaptation by introducing two additional metrics SigLIP Distance \citep{zhai2023sigmoid} and DINOv2 Distance \citep{oquab2024dinov2}. Similar to the computation of CLIP Distance, SigLIP and DINOv2 Distance are defined as the cosine distance between the SigLIP/DINOv2 embeddings of collected and generated samples. SigLIP provides evaluation in a different multimodal representation space, while DINOv2 Distance is a fully image-based metric that does not involve text embeddings.
    
    \item {\bf Intra-LPIPS.}
    To measure the diversity of the domain adaptation, we use Intra-LPIPS metric \citep{ojha2021cdc} which first assigns generated samples to one of $K$ clusters, then averages pair-wise distance within the cluster members and reports the average value over $K$ clusters. 
    In zero-shot setup, since there are no training images, we follow \citep{gal2022stylegannada, jeon2023svl} to cluster around generated samples using $K$-Medoids \citep{kmedoids}, with $K=10$.

    \item {\bf User Study.} 
    We also conducted a user study to compare the quality and the diversity of the domain adaptation with different schemes based on human feedback. See more details in Sec. \ref{sec:User_study}.
\end{itemize}

We remark that similarly NADA reports Intra-LPIPS on AFHQ-Cat, and SVL reports both FID and Intra-LPIPS on AFHQ-Cat.

\textbf{Text-Guided Object-Centric Image Editing} Following Style-Editor \citep{park2025styleeditor}, we evaluate the edited foreground and the preserved background separately using the ground-truth object masks from MS COCO \citep{lin2014mscoco}. Let $I^{\mathrm{src}}$ and $I^{\mathrm{out}}$ denote the source and edited images, respectively, $T^{\mathrm{tgt}}$ the target text, $M_F$ the foreground mask, and $M_B=1-M_F$ the corresponding background mask. We define $I_F^x=\operatorname{crop}(I^x\odot M_F)$ and $I_B^x=I^x\odot M_B$, where $x\in\{\mathrm{src},\mathrm{out}\}$. The masks are used only for evaluation and are not provided to either editing method.

\begin{itemize}
    \item {\bf Foreground text similarity ($\mathrm{Sim}_F$).}
    This metric measures how well the edited foreground matches the target text using the cosine similarity between their CLIP embeddings: $\mathrm{Sim}_F=\operatorname{cos}\!\left(E_T(T^{\mathrm{tgt}}),E_I(I_F^{\mathrm{out}})\right).$
    A higher value indicates stronger alignment between the target description and the edited object.

    \item {\bf Foreground content distance ($\mathrm{Con}_F$).}
    We compute the VGG content loss between the source and edited foregrounds,
    $\mathrm{Con}_F=\mathcal{L}_{\mathrm{content}}^{\mathrm{VGG}}(I_F^{\mathrm{src}},I_F^{\mathrm{out}})$.
    A lower value indicates better preservation of the object's semantic content and structure during stylization.

    \item {\bf Background content distance ($\mathrm{Con}_B$).}
    We apply the same VGG content loss to the source and edited backgrounds,
    $\mathrm{Con}_B=\mathcal{L}_{\mathrm{content}}^{\mathrm{VGG}}(I_B^{\mathrm{src}},I_B^{\mathrm{out}})$.
    A lower value indicates less structural or semantic change outside the edited object.

    \item {\bf Background style distance ($\mathrm{Sty}_B$).}
    This metric is the mean--variance VGG style loss between the source and edited backgrounds,
    $\mathrm{Sty}_B=\mathcal{L}_{\mathrm{style}}^{\mathrm{VGG}}(I_B^{\mathrm{src}},I_B^{\mathrm{out}})$.
    A lower value indicates that the original background appearance is better preserved and that less target style leaks into irrelevant regions.
\end{itemize}

Other perceptual metrics such as LPIPS are not directly suitable for our tasks, because they generally assume that the two images depict the same underlying content and are approximately spatially aligned. They are therefore most meaningful for measuring perceptual distortion or reconstruction quality between paired images. In our settings, however, no paired ground-truth target image is available for either zero-shot generative model adaptation or text-guided object-centric image editing.
In addition, we believe the included visual results can help in transparency and reflecting the superior performance of our proposed method in terms of adaptation quality.

\subsection{More Experimental Results of Downstream Tasks}
\label{sec:more_experimental_results}

We include a total of \textbf{32} different configurations of zero-shot adaptation in this paper. The experimental setting and evaluation metric follow 
Sec. 5 
in the main paper.

\begin{figure*}[!htbp]
    \centering
    \includegraphics[width=0.9\textwidth]{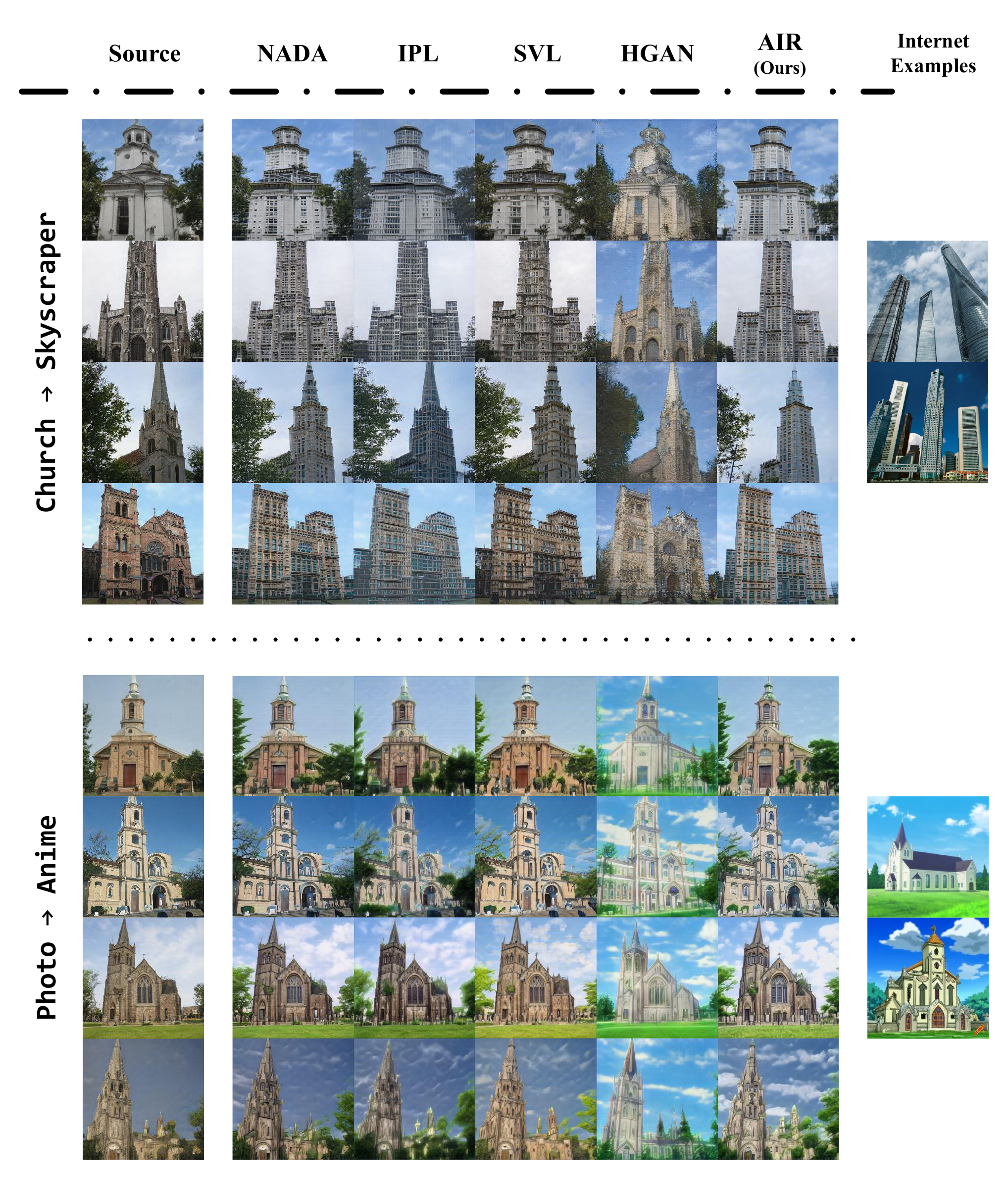}
    \caption{{\bf Additional zero-shot adaptation results from source domain \texttt{Church}.}
    Here we use a StyleGAN2 generator pre-trained on the LSUN-Church \cite{yu2015lsun} dataset as $G_{\mathcal{S}}$ and shift this to various target domains using different zero-shot approaches.
    We report the qualitative results for two setups: \texttt{Church}$\rightarrow$\texttt{Skyscraper} and \texttt{Photo}$\rightarrow$\texttt{Anime}. We also compute CLIP Distance on 5K generated samples as quantitative results, as shown in 
    Tab. \ref{Tab:Quant_GAN}, 
    our proposed AIR approach results in less CLIP distance meaning that the adapted generator is closer to the target domain. 
    Additionally, qualitative results indicate that \textbf{our method generally adapts better to the target domain and produces higher-quality outputs than NADA, IPL, and SVL, while HGAN suffers severely from mode collapse to one of the target examples.}
    For example, in \texttt{Church}$\rightarrow$\texttt{Skyscraper}, NADA and IPL samples contain artifacts around windows, and SVL still has some structures related to the church like the arch in the middle of the skyscraper.
    In \texttt{Photo}$\rightarrow$\texttt{Anime}, NADA, IPL, and SVL failed  to fully adapt to target style, and HGAN generates buildings that are partially transparent and exhibit undesirable green artifacts.
    }
    
    \label{fig:More_Quali_GAN_Church1}
\end{figure*}


\begin{figure*}[!htbp]
    \centering
    \includegraphics[width=0.92\textwidth]{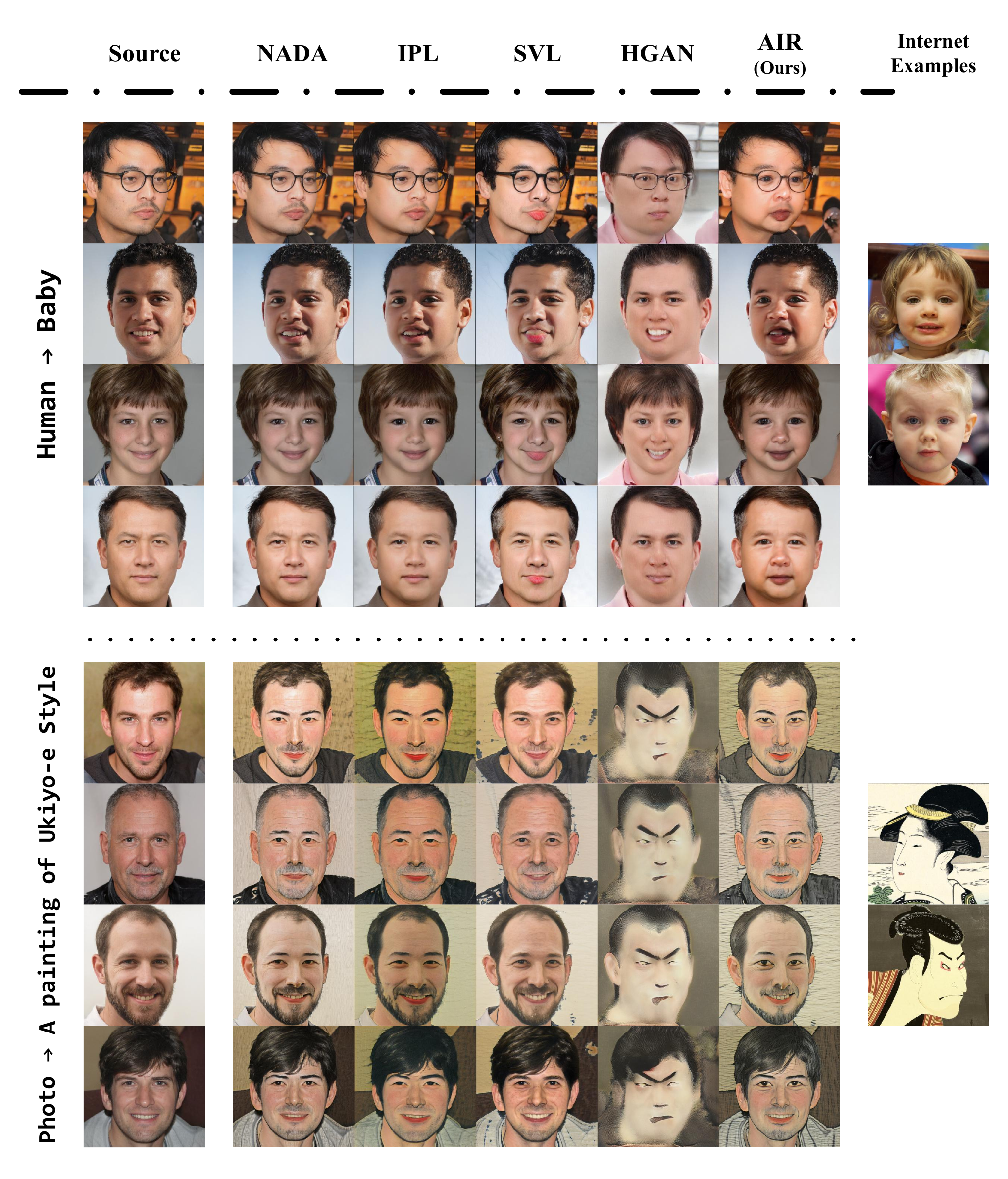}
    \caption{
    {\bf Additional zero-shot adaptation results from source domain \texttt{FFHQ}.}
    Here we use a StyleGAN2 generator pre-trained on the FFHQ \cite{karras2019style} (human faces) dataset as $G_{\mathcal{S}}$ and shift this to various target domains using different zero-shot approaches.
    We report the qualitative results for two setups: \texttt{Photo}$\rightarrow$\texttt{Baby} and \texttt{Photo}$\rightarrow$\texttt{A Painting of Ukiyo-e Style}. 
    We also compute CLIP Distance on 5K generated samples as quantitative results, as shown in 
    Tab. \ref{Tab:Quant_GAN}, 
    our proposed AIR approach results in less CLIP Distance meaning that the adapted generator are closer to the target domain. 
    Additionally, qualitative results indicate that \textbf{our method generally adapts better to the target domain and produces higher-quality outputs than NADA, IPL, and SVL, while HGAN suffers severely from mode collapse to one of the target examples.}
    }
    \label{fig:More_Quali_GAN_FFHQ1}
\end{figure*}

\begin{figure*}[!htbp]
    \centering
    \includegraphics[width=0.92\textwidth]{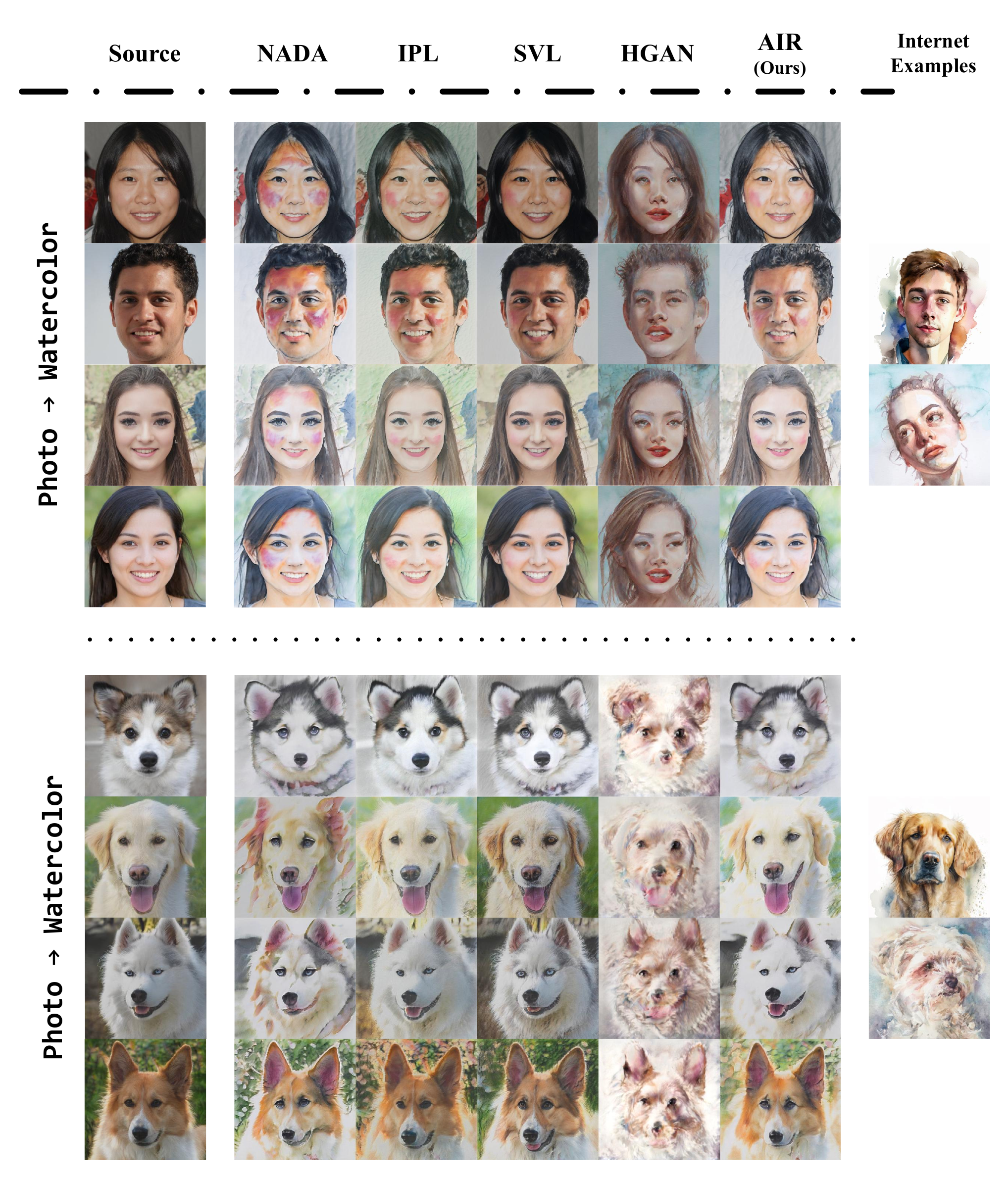}
    \caption{
    {\bf Additional zero-shot adaptation results from source domain \texttt{FFHQ}.}
    Here we use a StyleGAN2 generator pre-trained on the FFHQ \cite{karras2019style} (human faces) and AFHQ-Dog \cite{choi2020starganv2} (dog face) dataset as $G_{\mathcal{S}}$ and shift this to various target domains using different zero-shot approaches.
    We report the qualitative results for the setups: \texttt{Photo}$\rightarrow$\texttt{Watercolor}.
    We also compute CLIP Distance on 5K generated samples as quantitative results, as shown in 
    Tab. \ref{Tab:Quant_GAN}, 
    our proposed AIR approach results in less CLIP Distance meaning that the adapted generator are closer to the target domain. 
    Additionally, qualitative results indicate that \textbf{our method generally adapts better to the target domain and produces higher-quality outputs than NADA, IPL, and SVL, while HGAN suffers severely from mode collapse to one of the target examples.}
    }
    \label{fig:More_Quali_GAN_FFHQnDog}
\end{figure*}

\begin{figure*}[!htbp]
    \centering
    \includegraphics[width=0.92\textwidth]{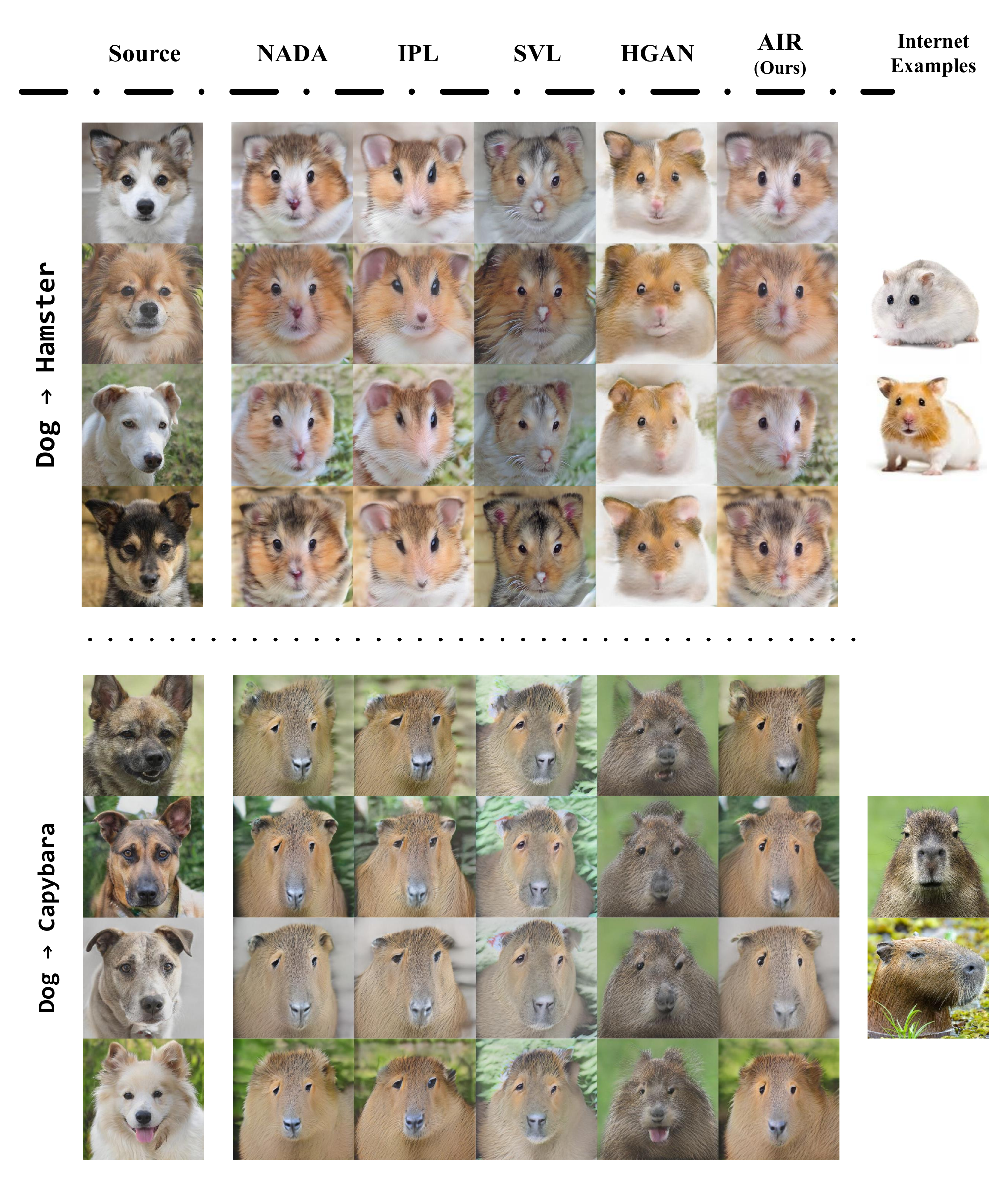}
    \caption{
    {\bf Additional zero-shot adaptation results from source domain \texttt{Dog}.}
    Here we use a StyleGAN2 generator pre-trained on the AFHQ-Dog \cite{choi2020starganv2} dataset as $G_{\mathcal{S}}$ and shift this to various target domains using different zero-shot approaches.
    We report the qualitative results for two setups: \texttt{Dog}$\rightarrow$\texttt{Hmaster} and \texttt{Dog}$\rightarrow$\texttt{Capybara}. We also compute CLIP Distance on 5K generated samples as quantitative results, as shown in 
    Tab. \ref{Tab:Quant_GAN}, 
    for both setups, our proposed AIR approach results in less CLIP Distance meaning that the adapted generator are closer to the target domain. 
    Additionally, qualitative results indicate that \textbf{our method generally adapts better to the target domain and produces higher-quality outputs than NADA, IPL, and SVL, while HGAN suffers severely from mode collapse to one of the target examples.}
    For example, for  \texttt{Dog}$\rightarrow$\texttt{Capybara} setup, generated samples with other approaches have degradations like unsymmetrical faces or eyes.
    }
    \label{fig:More_Quali_GAN_Dog1}
\end{figure*}

\begin{figure*}[!htbp]
    \centering
    \includegraphics[width=0.9\textwidth]{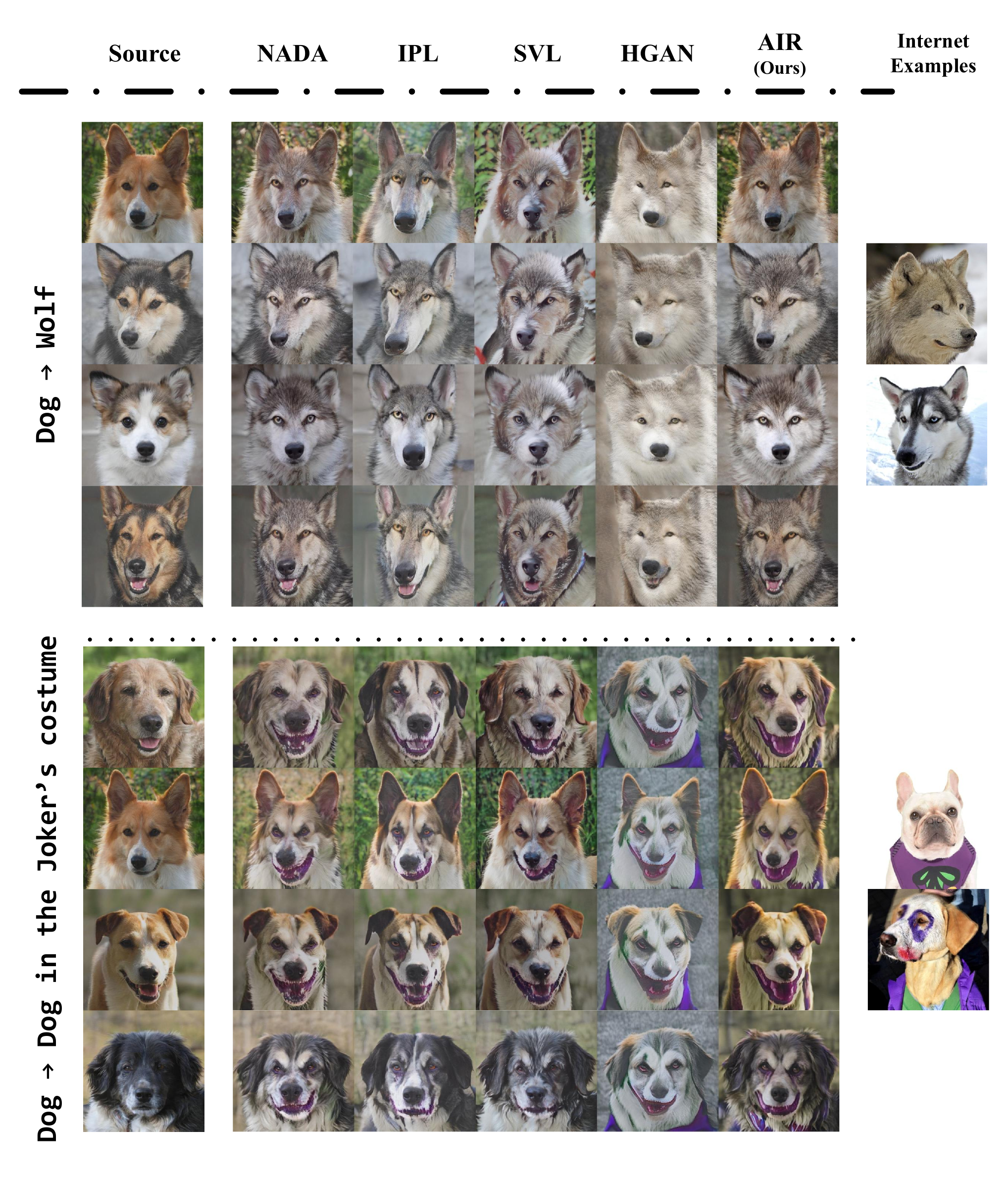}
    \caption{
    {\bf Additional zero-shot adaptation results from source domain \texttt{Dog}.}
    Here we use a StyleGAN2 generator pre-trained on the AFHQ-Dog \cite{choi2020starganv2} dataset as $G_{\mathcal{S}}$ and shift this to various target domains using different zero-shot approaches.
    We report the qualitative results for two setups: \texttt{Dog}$\rightarrow$\texttt{Wolf} and \texttt{Dog}$\rightarrow$\texttt{The Joker}. We also compute CLIP Distance on 5K generated samples as quantitative results, as shown in 
    Tab. \ref{Tab:Quant_GAN}, 
    for both setups, our proposed AIR approach results in less CLIP Distance meaning that the adapted generator are closer to the target domain. 
    Additionally, qualitative results indicate that \textbf{our method generally adapts better to the target domain and produces higher-quality outputs than NADA, IPL, and SVL, while HGAN suffers severely from mode collapse to one of the target examples.}
    For example, for \texttt{Dog}$\rightarrow$\texttt{Wolf} setup, IPL generates an unnaturally big snout and SVL has some artifacts in the generated sample. 
    For  \texttt{Dog}$\rightarrow$\texttt{The Joker} setup, our approach attains the mouth feature with a proper style and quality.
    }
    \label{fig:More_Quali_GAN_Dog2}
\end{figure*}

\begin{table*}[htbp]
\centering
\caption{Full per-setting guidance-misalignment results for zero-shot diffusion model adaptation. We compare the directional guidance used by each method with the corresponding reference visual offset. Results are reported as mean $\pm$ standard deviation.}
\label{Tab:OM_Alleviation_Diffusion_Full}
\resizebox{0.7\textwidth}{!}{
\begin{tabular}{@{}c|l|cc@{}}
\toprule
\multirow{2}{*}{\makecell{Pre-trained\\Dataset}} & \multirow{2}{*}{Source $\rightarrow$ Target Setting} & \multicolumn{2}{c}{Guidance Misalignment $\downarrow$} \\
\cmidrule(l){3-4}
& & NADA & AIR \\
\midrule
\multirow{7}{*}{FFHQ} & \texttt{Human} $\rightarrow$ \texttt{Baby} & 0.68{\scriptsize$\pm$ 0.02} & \textbf{0.52}{\scriptsize$\pm$ 0.08} \\
& \texttt{Human} $\rightarrow$ \texttt{Werewolf} & 0.71{\scriptsize$\pm$ 0.01} & \textbf{0.34}{\scriptsize$\pm$ 0.05} \\
& \texttt{Human} $\rightarrow$ \texttt{Pixar character} & 0.58{\scriptsize$\pm$ 0.01} & \textbf{0.38}{\scriptsize$\pm$ 0.09} \\
& \texttt{Photo} $\rightarrow$ \texttt{Sketch} & 0.42{\scriptsize$\pm$ 0.02} & \textbf{0.41}{\scriptsize$\pm$ 0.07} \\
& \texttt{Photo} $\rightarrow$ \texttt{Wall painting} & 0.71{\scriptsize$\pm$ 0.01} & \textbf{0.52}{\scriptsize$\pm$ 0.10} \\
& \texttt{Photo} $\rightarrow$ \texttt{Watercolor} & 0.56{\scriptsize$\pm$ 0.02} & \textbf{0.32}{\scriptsize$\pm$ 0.06} \\
& \texttt{Photo} $\rightarrow$ \texttt{Ukiyo-e} & 0.73{\scriptsize$\pm$ 0.01} & \textbf{0.52}{\scriptsize$\pm$ 0.08} \\
\midrule
\multirow{7}{*}{Dog} & \texttt{Dog} $\rightarrow$ \texttt{Cat} & 0.54{\scriptsize$\pm$ 0.02} & \textbf{0.24}{\scriptsize$\pm$ 0.09} \\
& \texttt{Dog} $\rightarrow$ \texttt{Wolf} & 0.33{\scriptsize$\pm$ 0.02} & \textbf{0.24}{\scriptsize$\pm$ 0.07} \\
& \texttt{Dog} $\rightarrow$ \texttt{Hamster} & 0.36{\scriptsize$\pm$ 0.01} & \textbf{0.26}{\scriptsize$\pm$ 0.11} \\
& \texttt{Dog} $\rightarrow$ \texttt{Capybara} & 0.29{\scriptsize$\pm$ 0.01} & \textbf{0.24}{\scriptsize$\pm$ 0.05} \\
& \texttt{Dog} $\rightarrow$ \texttt{the Joker} & 0.64{\scriptsize$\pm$ 0.02} & \textbf{0.42}{\scriptsize$\pm$ 0.10} \\
& \texttt{Photo} $\rightarrow$ \texttt{Cartoon} & 0.59{\scriptsize$\pm$ 0.01} & \textbf{0.38}{\scriptsize$\pm$ 0.08} \\
& \texttt{Photo} $\rightarrow$ \texttt{Watercolor} & 0.49{\scriptsize$\pm$ 0.02} & \textbf{0.41}{\scriptsize$\pm$ 0.06} \\
\bottomrule
\end{tabular}}
\end{table*}

\begin{table*}[htbp]
\centering
\caption{Full quantitative results of zero-shot diffusion model adaptation. Results are reported as mean $\pm$ standard deviation computed from 3 runs with different random seeds. All methods use the same CLIP guidance for fair comparison. FID is available only for \texttt{Human} $\rightarrow$ \texttt{Baby} and \texttt{Dog} $\rightarrow$ \texttt{Cat}, which have sufficient data for reliable evaluation}
\label{Tab:Quant_Diff_Full}
\resizebox{\textwidth}{!}{
\begin{tabular}{@{}cc|cc|cc|cc@{}}
\toprule
\multirow{2}{*}{\makecell[c]{Pre-trained \\ Dataset}} & \multirow{2}{*}{Adaptation} & \multicolumn{2}{c|}{CLIP Distance ($\downarrow$)} & \multicolumn{2}{c|}{Intra-LPIPS ($\uparrow$)} & \multicolumn{2}{c}{FID ($\downarrow$)} \\
\cline{3-8}
\noalign{\vskip 2pt} & & NADA & AIR & NADA & AIR & NADA & AIR \\
\midrule
\multirow{7}{*}{FFHQ}
& \texttt{Human} $\rightarrow$ \texttt{Baby} & 0.2612{\scriptsize$\pm$ 0.0018} & \textbf{0.2172}{\scriptsize$\pm$ 0.0022} & 0.5690{\scriptsize$\pm$ 0.0026} & \textbf{0.5763}{\scriptsize$\pm$ 0.0037} & 63.68{\scriptsize$\pm$ 0.75} & \textbf{58.74}{\scriptsize$\pm$ 1.56} \\
& \texttt{Human} $\rightarrow$ \texttt{Werewolf} & 0.2777{\scriptsize$\pm$ 0.0013} & \textbf{0.2323}{\scriptsize$\pm$ 0.0026} & \textbf{0.5194}{\scriptsize$\pm$ 0.0025} & 0.5185{\scriptsize$\pm$ 0.0048} & - & - \\
& \texttt{Human} $\rightarrow$ \texttt{Pixar character} & 0.4296{\scriptsize$\pm$ 0.0011} & \textbf{0.3865}{\scriptsize$\pm$ 0.0025} & \textbf{0.4565}{\scriptsize$\pm$ 0.0024} & 0.4558{\scriptsize$\pm$ 0.0039} & - & - \\
& \texttt{Photo} $\rightarrow$ \texttt{Sketch} & 0.4390{\scriptsize$\pm$ 0.0014} & \textbf{0.3601}{\scriptsize$\pm$ 0.0037} & \textbf{0.4873}{\scriptsize$\pm$ 0.0028} & 0.4815{\scriptsize$\pm$ 0.0035} & - & - \\
& \texttt{Photo} $\rightarrow$ \texttt{Wall painting} & 0.4781{\scriptsize$\pm$ 0.0011} & \textbf{0.4744}{\scriptsize$\pm$ 0.0022} & 0.5252{\scriptsize$\pm$ 0.0021} & \textbf{0.5270}{\scriptsize$\pm$ 0.0037} & - & - \\
& \texttt{Photo} $\rightarrow$ \texttt{Watercolor} & 0.3264{\scriptsize$\pm$ 0.0012} & \textbf{0.3229}{\scriptsize$\pm$ 0.0033} & 0.6224{\scriptsize$\pm$ 0.0019} & \textbf{0.6399}{\scriptsize$\pm$ 0.0037} & - & - \\
& \texttt{Photo} $\rightarrow$ \texttt{Ukiyo-e} & 0.2307{\scriptsize$\pm$ 0.0014} & \textbf{0.2152}{\scriptsize$\pm$ 0.0036} & 0.5431{\scriptsize$\pm$ 0.0018} & \textbf{0.5524}{\scriptsize$\pm$ 0.0035} & - & - \\
\midrule
\multirow{7}{*}{Dog}
& \texttt{Dog} $\rightarrow$ \texttt{Cat} & 0.1434{\scriptsize$\pm$ 0.0017} & \textbf{0.1428}{\scriptsize$\pm$ 0.0023} & 0.5411{\scriptsize$\pm$ 0.0018} & \textbf{0.5494}{\scriptsize$\pm$ 0.0034} & 84.75{\scriptsize$\pm$ 1.07} & \textbf{77.15}{\scriptsize$\pm$ 1.49} \\
& \texttt{Dog} $\rightarrow$ \texttt{Wolf} & 0.1460{\scriptsize$\pm$ 0.0015} & \textbf{0.1337}{\scriptsize$\pm$ 0.0036} & 0.4649{\scriptsize$\pm$ 0.0022} & \textbf{0.4735}{\scriptsize$\pm$ 0.0035} & - & - \\
& \texttt{Dog} $\rightarrow$ \texttt{Hamster} & 0.1851{\scriptsize$\pm$ 0.0017} & \textbf{0.1588}{\scriptsize$\pm$ 0.0021} & 0.4854{\scriptsize$\pm$ 0.0018} & \textbf{0.4952}{\scriptsize$\pm$ 0.0039} & - & - \\
& \texttt{Dog} $\rightarrow$ \texttt{Capybara} & 0.1576{\scriptsize$\pm$ 0.0011} & \textbf{0.1204}{\scriptsize$\pm$ 0.0026} & \textbf{0.4832}{\scriptsize$\pm$ 0.0022} & 0.4604{\scriptsize$\pm$ 0.0052} & - & - \\
& \texttt{Dog} $\rightarrow$ \texttt{the Joker} & 0.3955{\scriptsize$\pm$ 0.0012} & \textbf{0.3208}{\scriptsize$\pm$ 0.0022} & 0.4571{\scriptsize$\pm$ 0.0018} & \textbf{0.4663}{\scriptsize$\pm$ 0.0039} & - & - \\
& \texttt{Photo} $\rightarrow$ \texttt{Cartoon} & 0.2516{\scriptsize$\pm$ 0.0012} & \textbf{0.2481}{\scriptsize$\pm$ 0.0025} & 0.5572{\scriptsize$\pm$ 0.0027} & \textbf{0.5611}{\scriptsize$\pm$ 0.0040} & - & - \\
& \texttt{Photo} $\rightarrow$ \texttt{Watercolor} & 0.1932{\scriptsize$\pm$ 0.0016} & \textbf{0.1853}{\scriptsize$\pm$ 0.0027} & 0.5173{\scriptsize$\pm$ 0.0022} & \textbf{0.5322}{\scriptsize$\pm$ 0.0038} & - & - \\
\midrule
\multicolumn{2}{c|}{Avg.} & 0.2797{\scriptsize$\pm$ 0.0014} & \textbf{0.2513}{\scriptsize$\pm$ 0.0028} & 0.5164{\scriptsize$\pm$ 0.0022} & \textbf{0.5207}{\scriptsize$\pm$ 0.0039} & 74.22{\scriptsize$\pm$ 0.91} & \textbf{67.95}{\scriptsize$\pm$ 1.53} \\
\bottomrule
\end{tabular}}
\end{table*}

\subsubsection{Zero-shot GAN Adaptation}
\label{ssec:Zero-shot_GAN_Adaptation}

In this section, we provide additional experimental results including quantitative and qualitative results for different adaptation setups using GAN as the generator and introduce more evaluation metrics.

\textbf{Qualitatives Results.} 
In Fig. \ref{fig:More_Quali_GAN_Church1} 
, we perform zero-shot adaptation of a StyleGAN2 pre-trained on LSUN-Church \cite{yu2015lsun} to four different target domains including \texttt{Skyscraper}, and \texttt{Anime}.
In Fig. \ref{fig:More_Quali_GAN_FFHQ1} and Fig. \ref{fig:More_Quali_GAN_FFHQnDog} show the qualitative results of zero-shot adaptation of a StyleGAN2 pre-trained on FFHQ \cite{karras2019style} dataset to two different target domains including \texttt{Baby}, \texttt{Watercolor}, and \texttt{A Painting of Ukiyo-e Style}. Finally, Fig. \ref{fig:More_Quali_GAN_FFHQnDog}, Fig. \ref{fig:More_Quali_GAN_Dog1} and Fig. \ref{fig:More_Quali_GAN_Dog2} we report the zero-shot adaptation of a StyleGAN2 pre-trained on AFHQ-Dog \cite{choi2020starganv2} to four different target domains including \texttt{Watercolor}, \texttt{Hamster}, \texttt{Capybara}, \texttt{Wolf} and \texttt{The Joker}. The results show that our approach in general adapts better to the style of the target domain and has better sample quality (please check the caption of each image for more detailed discussion).

\textbf{Quantitative Results.} 
Quantitative results are reported by computing the CLIP Distance between the embeddings of 5K generated samples with each approach and the embedding of the reference images of the target domain in CLIP space. As the results show, generated samples by proposed AIR has smaller CLIP distance meaning that the adapted generators is closer to the target domain compared to the adapted generator by other zero-shot approaches.

\textbf{Additional Metrics.}
We further evaluate the quality of the domain adaptation by introducing two additional metrics SigLIP Distance \cite{zhai2023sigmoid} and DINOv2 Distance \cite{oquab2024dinov2}. Similar to the computation of CLIP Distance, SigLIP and DINOv2 Distance are defined as the cosine distance between the SigLIP/DINOv2 embeddings of collected and generated samples. As shown in Tab. \ref{Tab:Additional_Quant_GAN}, the results align with those in the main paper, further support the superiority of our proposed AIR.

\begin{table*}[ht]
\centering
\caption{Additional quantitative evaluation of zero-shot GAN adaptation using additional metrics of SigLIP/DINOv2, with the same setting of 
Tab. \ref{Tab:Quant_GAN}
in main paper.}
\resizebox{0.9\linewidth}{!}{
\label{Tab:Additional_Quant_GAN}
\begin{tabular}{@{}cc|ccccc|ccccc@{}}
\toprule
\multirow{2}{*}{\makecell[c]{Pre-trained \\ Dataset}} & \multirow{2}{*}{Target Domain} & \multicolumn{5}{c|}{SigLIP Distance  ($\downarrow$)} & \multicolumn{5}{c}{DINOv2 Distance ($\downarrow$)}  \\
\cline{3-7} \cline{8-12}
\noalign{\vskip 2pt} & & NADA & IPL & SVL & HGAN & AIR & NADA & IPL & SVL & HGAN & AIR \\
\midrule
\multirow{5}{*}{FFHQ} & \texttt{Baby} & 0.1925 & 0.1884 & 0.3474 & 0.3080 & \textbf{0.1833} 
& 0.5943 & 0.5993 & 0.8026 & 0.7703 & \textbf{0.5887} \\ 
& \texttt{Werewolf} & 0.3192 & 0.3930 & 0.4831 & 0.2425 & \textbf{0.2274} 
& 0.8500 & 0.8923 & 0.9365 & 0.6345 & \textbf{0.6097} \\ 
& \texttt{Pixar} & 0.2803 & 0.2762 & 0.4582 & 0.2857 & \textbf{0.2630} 
& 0.6935 & \textbf{0.6690} & 0.7848 & 0.7362 & 0.6785 \\ 
& \texttt{Sketch} & 0.2897 & 0.3173 & 0.3598 & 0.2939 & \textbf{0.2837} 
& 0.4682 & 0.5918 & 0.6291 & 0.4703 & \textbf{0.4420} \\ 
& \texttt{Wall painting} & 0.4205 & 0.4277 & 0.4489 & \textbf{0.4052} & 0.4103 
& 0.7256 & 0.7267 & 0.7836 & \textbf{0.6761} & 0.7004 \\ 
\midrule
\multirow{3}{*}{AFHQ-Dog} & \texttt{Cat} & 0.1395 & 0.2287 & 0.1819 & 0.1762 & \textbf{0.1297} 
& 0.8553 & 0.8737 & 0.8842 & 0.8623 & \textbf{0.8338} \\ 
 & \texttt{Cartoon} & 0.2580 & 0.2618 & 0.3140 & 0.3101 & \textbf{0.2518} 
 & 0.7899 & 0.8174 & 0.9078 & 0.8513 & \textbf{0.7644} \\ 
 & \texttt{Watercolor} & 0.1934 & 0.1980 & 0.2569 & 0.1916 & \textbf{0.1819} 
 & 0.8059 & 0.8328 & 0.8330 & \textbf{0.7721} & 0.7943 \\ 

\bottomrule
\end{tabular}
}
\end{table*}

\begin{figure*}[!htbp]
    \centering
    \includegraphics[width=0.85\textwidth]{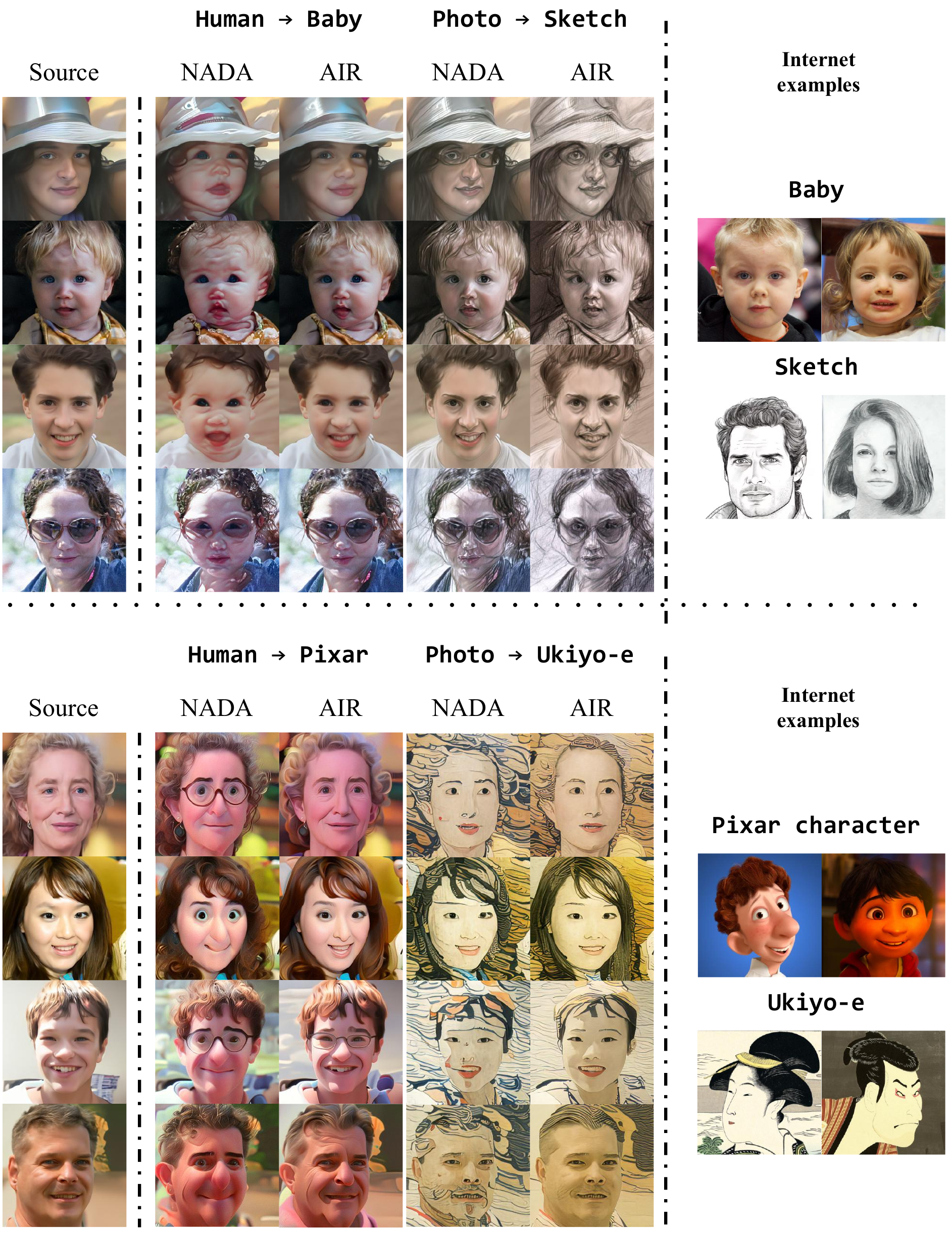}
    \caption{
    {\bf Additional zero-shot adaptation results.
    }
    We use a pre-trained Guided Diffusion model \cite{dhariwal2021diffusion} on FFHQ dataset \cite{karras2019style} as pre-trained generator $G_{\mathcal{S}}$ and perform zero-shot adaptation in four different setups: \texttt{Human}$\rightarrow$ \texttt{Baby}, \texttt{Photo}$\rightarrow$ \texttt{Sketch}, \texttt{Human}$\rightarrow$ \texttt{Pixar Character}, and \texttt{Photo}$\rightarrow$ \texttt{A painting in Ukiyo-e style} using both NADA and our proposed AIR approach. Quantitative results measured by CLIP distance in
    Tab. \ref{Tab:Quant_Diff} 
    shows that the domains adapted by our approach are closer to the target domain. In addition, qualitative results show that NADA suffers from degradation.
    }
    \label{fig:Diffusion_qualitative_FFHQ}
\end{figure*}

\begin{figure*}[!htbp]
    \centering
    \includegraphics[width=0.87\textwidth]{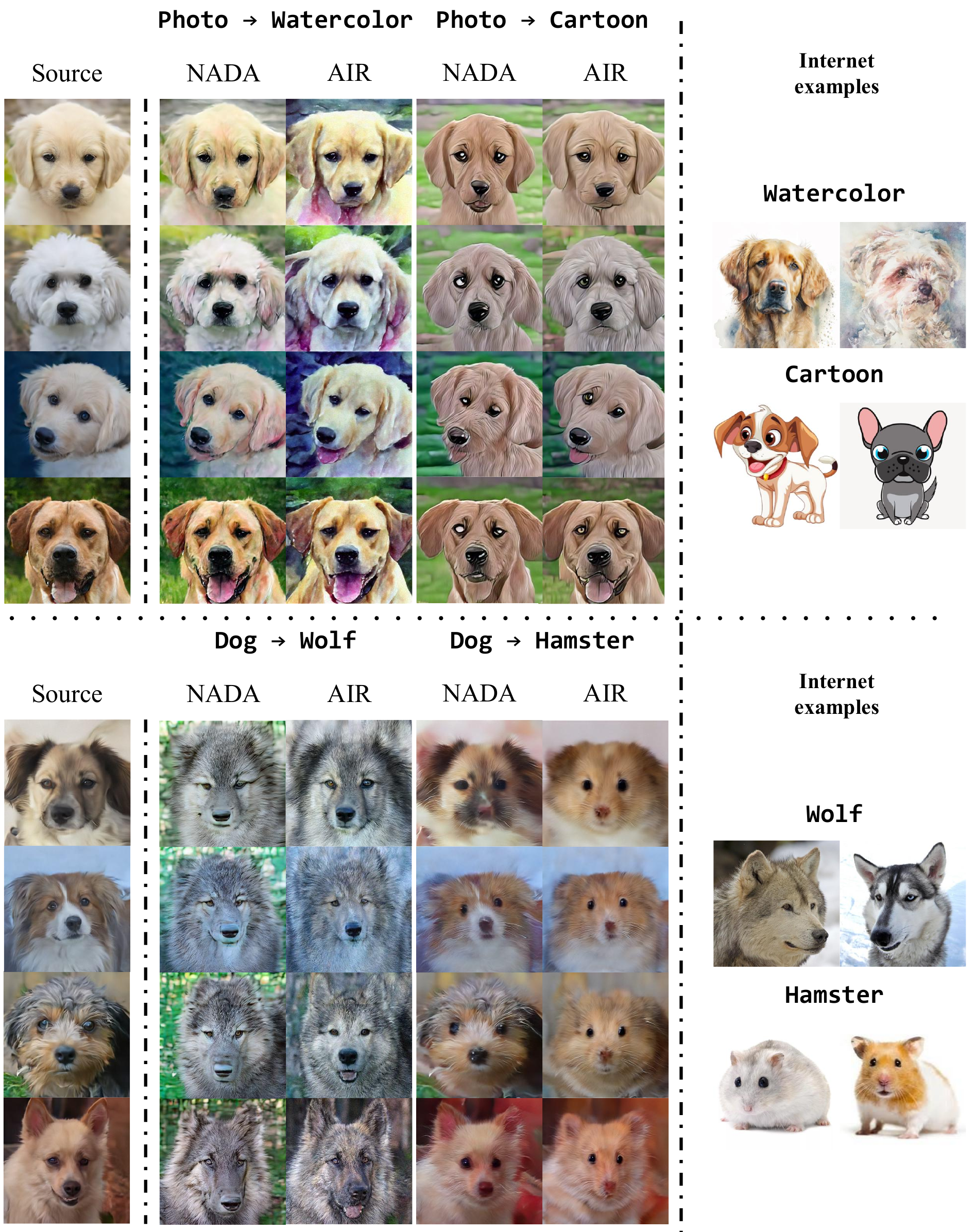}
    \caption{
    {\bf Additional zero-shot adaptation results.
    }
    We use a pre-trained Guided Diffusion model \cite{dhariwal2021diffusion} on AFHQ-Dog dataset \cite{choi2020starganv2} as pre-trained generator $G_{\mathcal{S}}$ and perform zero-shot adaptation in four different setups: \texttt{Photo}$\rightarrow$ \texttt{Watercolor}, \texttt{Photo}$\rightarrow$ \texttt{Cartoon}, \texttt{Dog}$\rightarrow$ \texttt{Wolf}, and \texttt{Dog}$\rightarrow$ \texttt{Hamster} using both NADA and our proposed AIR approach. Quantitative results measured by CLIP distance in 
    Tab. \ref{Tab:Quant_Diff} 
    shows that the adapted domain by our approach are closer to the target domain.
    Similarly, qualitative results show that our proposed AIR approach has better performance compared to NADA.
    }
    \label{fig:Diffusion_qualitative_Dog}
\end{figure*}

\subsubsection{Zero-shot Diffusion Model Adaptation}
\label{ssec:Zero-shot_Diffusion_Model_Adaptation}

In this section, we provide more qualitative and quantitative results of zero-shot diffusion model adaptation. 

\textbf{Qualitative Results.}
Here, we report the qualitative results of zero-shot diffusion model adaptation for the same configuration used in
Tab. \ref{Tab:Quant_Diff}
(main paper).
More specifically, we use the pre-trained Guided Diffusion model \cite{dhariwal2021diffusion} on two different source domains FFHQ \cite{karras2019style} (Fig. \ref{fig:Diffusion_qualitative_FFHQ}) and AFHQ-Dog \cite{choi2020starganv2} (Fig. \ref{fig:Diffusion_qualitative_Dog}) and shift these pre-trained models to different target domains using only text descriptions for both NADA and our proposed AIR approaches.
As illustrated in Fig. \ref{fig:Diffusion_qualitative_FFHQ} and Fig. \ref{fig:Diffusion_qualitative_Dog}, the domain adaptation with NADA suffer from degradation in the form of artifacts compared to our proposed AIR approach.

\subsection{Hyperparameters of AIR} \label{sec:Hyperparameters_of_AIR}

\subsubsection{Hyperparameters of AIR on ZSGM} \label{ssec:Hyperparameters_of_zero-shot_adaptation}

In Alg. \ref{alg:AIR}, for both GAN and diffusion model adaptation the batch size is set to 2.
Adaptation iteration $t_{adapt}$ is set to 
200 for in-domain changes like \texttt{Human}$\rightarrow$\texttt{Baby},
300 for texture-based changes such as \texttt{Photo}$\rightarrow$\texttt{Sketch}, 
and 2,000 for animal changes like \texttt{Dog}$\rightarrow$\texttt{Cat}. We set $t_{thresh}=50\%t_{adapt}$ to ensure there are some target domain concept encoded in $G_t$, and $t_{int}=10\%t_{adapt}$ to facilitate a stable and efficient training.

In Alg. \ref{alg:prompt_learning}, we generate 1,000 pairs of source and anchor images with the same batch of $w$ for each update. The number of learnable concept-description tokens $m$ is set to 4, and is initialized by "\texttt{A photo of a}". Each concept description learning session requires $k_{iter}=200$ iterations. 

For all experiments, we use an ADAM optimizer with a learning rate of 0.002 for both 
Algs.~\ref{alg:AIR} and~\ref{alg:prompt_learning}.
We conduct all the experiments on a single NVIDIA RTX 6000 Ada GPU. The training time is comparable to NADA as concept description learning in Alg. \ref{alg:prompt_learning} only requires $\sim$20 seconds.

\subsubsection{Hyperparameters of AIR on TOIE} \label{ssec:Hyperparameters_of_Text-Guided_Object-Centric_Image_Editing}

We follow Style-Editor \citep{park2025styleeditor} to set batch size as 1, and optimize for an iteration $t_{adapt}$ of 
200 for all settings. Similar to zero-shot generative model domain adaptation we set $t_{thresh}=50\%t_{adapt}$ to ensure there are some target domain concept encoded in $G_t$, and $t_{int}=10\%t_{adapt}$ to facilitate a stable and efficient training.

In Alg. \ref{alg:prompt_learning}, we take the transformed visual directly as the anchor visual. The number of learnable concept-description tokens $m$ is set to 4, and is initialized by "\texttt{A photo of a}". Each concept description learning session requires $k_{iter}=200$ iterations.

\subsection{Additional Ablation Studies}
\label{sec:Add_Ablation}

\subsubsection{Ablation on Hyperparameters Selection}
\label{ssec:Ablation_on_Hyperparameters_Selection}

We conduct an ablation study to determine the optimal hyperparameters. Specifically, Tab. \ref{Tab:Ablation_t_{int}} shows the ablation results for the adaptation interval $t_{int}$ to update anchor. Tab. \ref{Tab:Ablation_t_{thresh}} shows the ablation results for the starting iteration $t_{thresh}$ of applying AIR. A large $t_{int}$ which results in fewer updates of the anchor point, generally leads to a degradation in performance due to less precise adaptation. Conversely, a small $t_{int}$, while more computationally expensive, does not yield significant improvement. Thus, we set $t_{int}=10\%$ to balance the computation cost and adaptation precision. Similarly, neither excessively large nor small values of $t_{thresh}$ provide optimal adaptation performance. As shown in Fig. \ref{fig:Visual_Ablation} (b) and (c), the visual ablation results align with this conclusion. Hence, we empirically select $t_{thresh}=50\%$.

\begin{table}[ht]
    \centering
    \caption{Ablation study on adaptation interval $t_{int}$ to update anchor.}
    \label{Tab:Ablation_t_{int}}
    \scalebox{0.8}{
    \begin{tabular}{@{}c|cc|cc@{}}
    \toprule
    \multicolumn{5}{c}{$t_{int}$} \\
    \midrule
    \multirow{2}{*}{\% of $t_{adapt}$} & \multicolumn{2}{c|}{\texttt{Human} $\rightarrow$ \texttt{Baby}} & \multicolumn{2}{c}{\texttt{Dog} $\rightarrow$ \texttt{Cat}} \\
     & FID ($\downarrow$) & Intra-LPIPS ($\uparrow$) & FID ($\downarrow$) & Intra-LPIPS ($\uparrow$) \\
     \midrule
     5\% & 59.45 & 0.4512 & 59.97 & 0.4560 \\
     10\% & 62.13 & \textbf{0.4520} & \textbf{56.20} & \textbf{0.4628} \\
     15\% & 58.87 & 0.4515 & 61.92 & 0.4635 \\
     20\% & 64.54 & 0.4496 & 65.49 & 0.4537 \\
     25\% & \textbf{56.69} & 0.4511 & 67.49 & 0.4374 \\
     30\% & 76.39 & 0.4506 & 77.23 & 0.4513 \\
    \bottomrule
    \end{tabular}
    }
\end{table}

\begin{table}[ht]
    \centering
    \caption{Ablation study on starting iteration $t_{thresh}$ of applying AIR.}
    \label{Tab:Ablation_t_{thresh}}
    \scalebox{0.8}{
    \begin{tabular}{@{}c|cc|cc@{}}
    \toprule
    \multicolumn{5}{c}{$t_{thresh}$} \\
    \midrule
    \multirow{2}{*}{\% of $t_{adapt}$} & \multicolumn{2}{c|}{\texttt{Human} $\rightarrow$ \texttt{Baby}} & \multicolumn{2}{c}{\texttt{Dog} $\rightarrow$ \texttt{Cat}} \\
     & FID ($\downarrow$) & Intra-LPIPS ($\uparrow$) & FID ($\downarrow$) & Intra-LPIPS ($\uparrow$) \\
     \midrule
     0\% & 63.56 & 0.4324 & 83.33 & \textbf{0.4815} \\
     12.5\% & 75.92 & 0.4259 & 78.21 & 0.4642 \\
     25.0\% & 72.17 & 0.4386 & 73.65 & 0.4516 \\
     37.5\% & 64.14 & \textbf{0.4558} & 59.96 & 0.4496 \\
     50.0\% & \textbf{62.13} & 0.4520 & \textbf{56.20} & 0.4628 \\
     67.5\% & 68.25 & 0.4542 & 56.46 & 0.4344 \\
    \bottomrule
    \end{tabular}
    }
\end{table}

\subsubsection{Ablation on Anchor Label Initialization}
\label{ssec:Ablation_on_Anchor_Label_Initialization}

Our anchor concept description initialization in 
Sec. 4.2
is inspired by the standard 
concept description learning in VLM \cite{zhou2022coop, zhou2022cocoop}, which initializes label token $Y$ with class label of the image to serve as prior. 
However, in our setting, the anchor domain encodes both source and target concepts, so it cannot be described with natural language. Therefore, we leverage 
the continuous and semantically rich representation space of the text encoder to initialize anchor label $Y_{\mathcal{A}_i}$ via {\bf interpolation} between tokenized source and target descriptions.
We conduct an ablation study on the initialization of $Y_{\mathcal{A}_i}$. Results shown in Tab. \ref{Tab:Ablation_TokenInit} indicate the effectiveness of our idea by obtaining the best FID and Intra-LPIPS. 

\begin{table}[t]

    \centering
    \caption{Ablation study on initialization of $Y_{\mathcal{A}_i}$, using: a) Target domain label; b) Source domain label for the first half of adaptation, then target domain label; c) Interpolation as in our AIR.}
    \label{Tab:Ablation_TokenInit}
    \scalebox{0.9}{
    \begin{tabular}{@{}c|cc|cc@{}}
    \toprule
    \multirow{2}{*}{Init.} & \multicolumn{2}{c|}{\texttt{Human} $\rightarrow$ \texttt{Baby}} & \multicolumn{2}{c}{\texttt{Dog} $\rightarrow$ \texttt{Cat}} \\
    \cline{2-5}
    \noalign{\vskip 2pt} & FID ($\downarrow$) & Intra-LPIPS ($\uparrow$) & FID ($\downarrow$) & Intra-LPIPS ($\uparrow$) \\
    \midrule
    
    \noalign{\vskip 2pt} a) & 67.53 & 0.4513 & 65.83 & 0.4373 \\ 
    \noalign{\vskip 2pt} b) & 63.34 & 0.4512 & 56.44 & 0.4466  \\ 
    \noalign{\vskip 2pt} c) & \textbf{62.13} & \textbf{0.4520} & \textbf{56.20} & \textbf{0.4628} \\ 
    
    \bottomrule
    \end{tabular}
    }
\end{table}

\subsubsection{Visual Ablation Studies}
\label{ssec:Visual_Ablation_Studies}

\begin{figure*}[!htbp]
    \centering
    \includegraphics[width=0.6\linewidth]{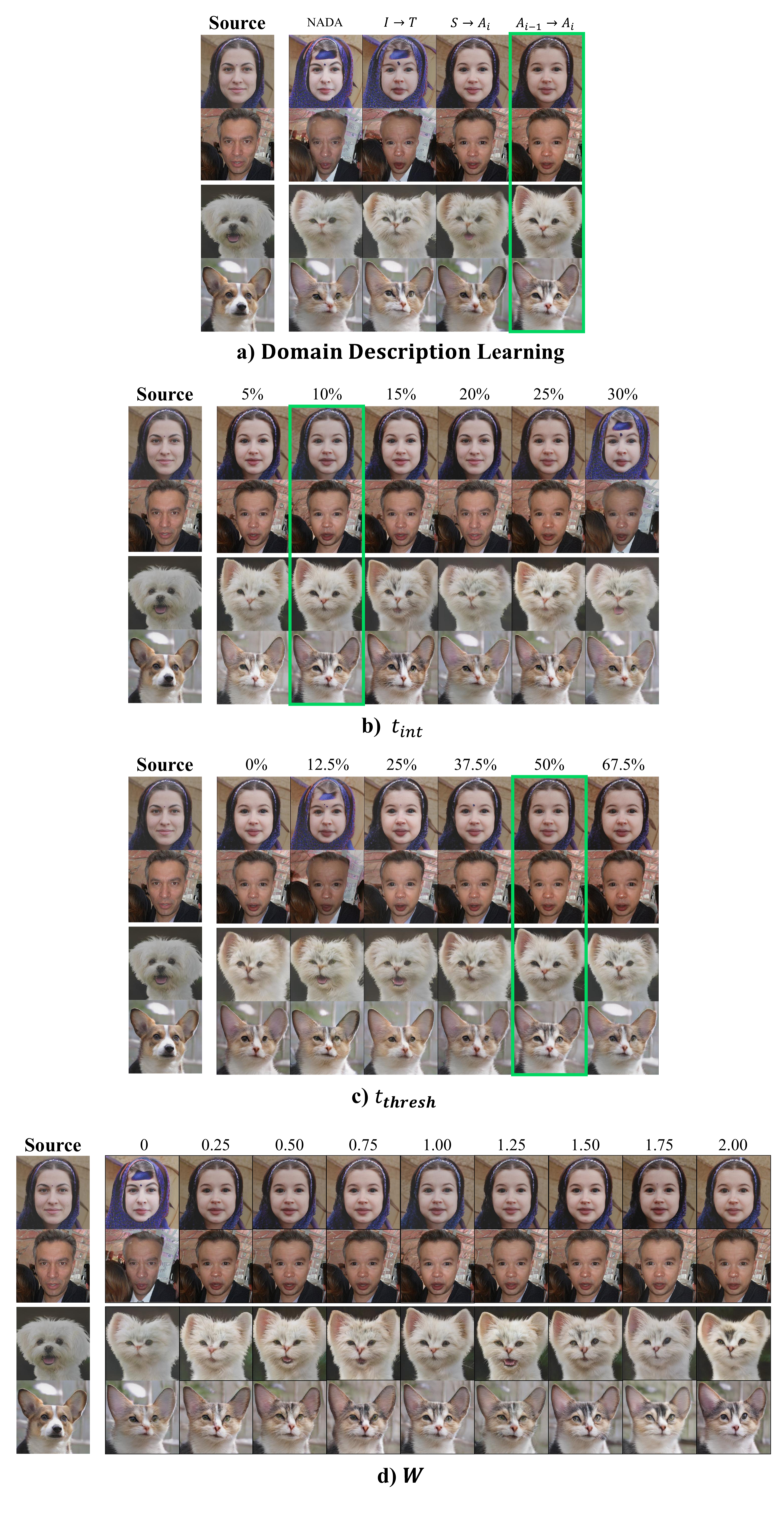}
    \caption{
    Visual ablation study of: a) Design choice of concept description learning; b) Adaptation interval $t_{int}$ to update anchor; c) Starting iteration $t_{thresh}$ of applying AIR. d) Loss weight $W$ for the proposed $\mathcal{L}_{adaptive}$}
    \label{fig:Visual_Ablation}
\end{figure*}

We perform visual ablation studies on concept description learning design and hyperparameters selection with the same experiments setting of 
Sec. 5.4, 
\ref{ssec:Ablation_on_Hyperparameters_Selection}, and Tab. \ref{Tab:Ablation_loss_weight}. The results in Fig. \ref{fig:Visual_Ablation} align consistently with the quantitative findings.

\subsubsection{Ablation on Loss Weighting}

We evaluate the sensitivity of our training objective $\mathcal{L} = \mathcal{L}_{directional} + W\mathcal{L}_{adaptive}$ by varying the weighting factor $W$. As shown in Tab. \ref{Tab:Ablation_loss_weight}, the effectiveness of our proposed $\mathcal{L}_{adaptive}$ is robust across different weight values, consistently improving over the baseline NADA setting ($W=0$). It is important to note that, we set $W=1$ in the main paper for simplicity and to avoid introducing an additional hyperparameter.

\begin{table}[ht]
    \centering
    \caption{Ablation study on loss weight for our proposed $\mathcal{L}_{adaptive}$. Results shows that the effectiveness of \textbf{our proposed $\mathcal{L}_{adaptive}$ is robust across different weight values, consistently improving over the baseline NADA setting ($W=0$).}}
    \label{Tab:Ablation_loss_weight}
    \scalebox{0.9}{
    \begin{tabular}{@{}c|cc|cc@{}}
    \toprule
    \multirow{2}{*}{$W$} & \multicolumn{2}{c|}{\texttt{Human} $\rightarrow$ \texttt{Baby}} & \multicolumn{2}{c}{\texttt{Dog} $\rightarrow$ \texttt{Cat}} \\
     & FID ($\downarrow$) & Intra-LPIPS ($\uparrow$) & FID ($\downarrow$) & Intra-LPIPS ($\uparrow$) \\
     \midrule
     0.00 & 68.35 & 0.4474 & 70.87 & 0.4439 \\
     0.25 & 66.59 & 0.4501 & 66.90 & 0.4659 \\
     0.50 & 62.54 & 0.4647 & 69.43 & 0.4580 \\
     0.75 & 58.87 & 0.4494 & 66.62 & 0.4515 \\
     1.00 & 62.13 & 0.4520 & 56.20 & 0.4628 \\
     1.25 & 59.35 & 0.4510 & 57.51 & 0.4458 \\
     1.50 & 63.02 & 0.4525 & 52.62 & 0.4459 \\
     1.75 & 66.38 & 0.4468 & 63.55 & 0.4513 \\
     2.00 & 65.17 & 0.4506 & 55.35 & 0.4596 \\
    \bottomrule
    \end{tabular}
    }
\end{table}

\subsection{Zero-shot Adaptation with Other Multimodal Contrastive Models}
\label{sec:Adaptation_with_Other_Multimodal_Contrastive_Models}

We present additional adaptation results with guidance from more multimodal constrastive representation spaces in Tab. \ref{Tab:Adaptation_with_more_Multimodal_Contrastive_Model}. 

\begin{table}[htbp]
\centering
\caption{Quantitative results of zero-shot generative model adaptation with guidance from more multimodal contrastive models. Results demonstrate that better-aligned multimodal contrastive representation spaces generally lead to better adaptation quality and diversity. AIR achieves SOTA across these multimodal contrastive models as guidance.
}
\resizebox{\linewidth}{!}{
\label{Tab:Adaptation_with_more_Multimodal_Contrastive_Model}
\begin{tabular}{@{}ccc|cc|cc|cc@{}}
  \toprule
  \multirow{2}{*}{Adaptation} & \multirow{2}{*}{\makecell[c]{Multimodal\\ Contrastive Model}} & \multirow{2}{*}{\makecell[c]{Avg. Offset\\ Misalignment}} & \multicolumn{2}{c|}{CLIP Distance ($\downarrow$)} & \multicolumn{2}{c|}{Intra-LPIPS ($\uparrow$)} & \multicolumn{2}{c}{FID ($\downarrow$)} \\
  \cline{4-9}
  \noalign{\vskip 2pt} & & & NADA & AIR & NADA & AIR & NADA & AIR \\
  \midrule
  \multirow{8}{*}{\texttt{Human} $\rightarrow$ \texttt{Baby}}
  & CLIP ViT-B/32 & 0.537 & 0.3327 & \textbf{0.3325} & 0.4474 & \textbf{0.4520} & 68.35 & \textbf{62.13} \\
  & CLIP ConvNext-L & 0.408 & 0.3204 & \textbf{0.3126} & 0.4551 & \textbf{0.4566} & 63.02 & \textbf{57.61} \\
  & CLIP RN50x64 & 0.613 & 0.3418 & \textbf{0.3406} & 0.4422 & \textbf{0.4471} & 72.96 & \textbf{66.04} \\
  & SigLIP ViT-L/16-256 & 0.387 & 0.3171 & \textbf{0.3014} & 0.4569 & \textbf{0.4588} & 61.72 & \textbf{56.08} \\
  & SigLIP SoViT-400m & 0.352 & 0.3142 & \textbf{0.2937} & 0.4599 & \textbf{0.4608} & 59.35 & \textbf{54.22} \\
  & EVA-CLIP 8B & 0.485 & 0.3269 & \textbf{0.3261} & 0.4507 & \textbf{0.4535} & 66.17 & \textbf{60.12} \\
  & DFN5B-CLIP ViT-H/14-378 & 0.331 & 0.3113 & \textbf{0.3106} & \textbf{0.4627} & 0.4620 & 58.84 & \textbf{53.67} \\
  & MobileCLIP & 0.652 & 0.3497 & \textbf{0.3446} & \textbf{0.4409} & 0.4411 & 74.82 & \textbf{68.31} \\
  \midrule
  \multirow{8}{*}{\texttt{Dog} $\rightarrow$ \texttt{Cat}}
  & CLIP ViT-B/32 & 0.537 & 0.1493 & \textbf{0.1320} & 0.4439 & \textbf{0.4628} & 70.87 & \textbf{56.20} \\
  & CLIP ConvNext-L & 0.408 & 0.1418 & \textbf{0.1297} & 0.4564 & \textbf{0.4638} & 62.28 & \textbf{49.31} \\
  & CLIP RN50x64 & 0.613 & 0.1549 & \textbf{0.1462} & 0.4387 & \textbf{0.4502} & 76.64 & \textbf{60.47} \\
  & SigLIP ViT-L/16-256 & 0.387 & 0.1409 & \textbf{0.1284} & 0.4582 & \textbf{0.4645} & 61.36 & \textbf{48.86} \\
  & SigLIP SoViT-400m & 0.352 & 0.1382 & \textbf{0.1196} & 0.4607 & \textbf{0.4653} & 58.56 & \textbf{46.57} \\
  & EVA-CLIP 8B & 0.485 & 0.1454 & \textbf{0.1402} & 0.4491 & \textbf{0.4632} & 67.44 & \textbf{53.29} \\
  & DFN5B-CLIP ViT-H/14-378 & 0.331 & 0.1368 & \textbf{0.1253} & \textbf{0.4669} & 0.4659 & 57.41 & \textbf{45.63} \\
  & MobileCLIP & 0.652 & 0.1592 & \textbf{0.1367} & 0.4332 & \textbf{0.4409} & 89.12 & \textbf{64.01} \\
  \bottomrule
\end{tabular}
}
\end{table}



    

\subsection{Concept Shifts during Transformation}
\label{sec:concept_shifts_during_adaptation}

\begin{figure*}[t]
    \centering
    \includegraphics[width=0.45\textwidth]{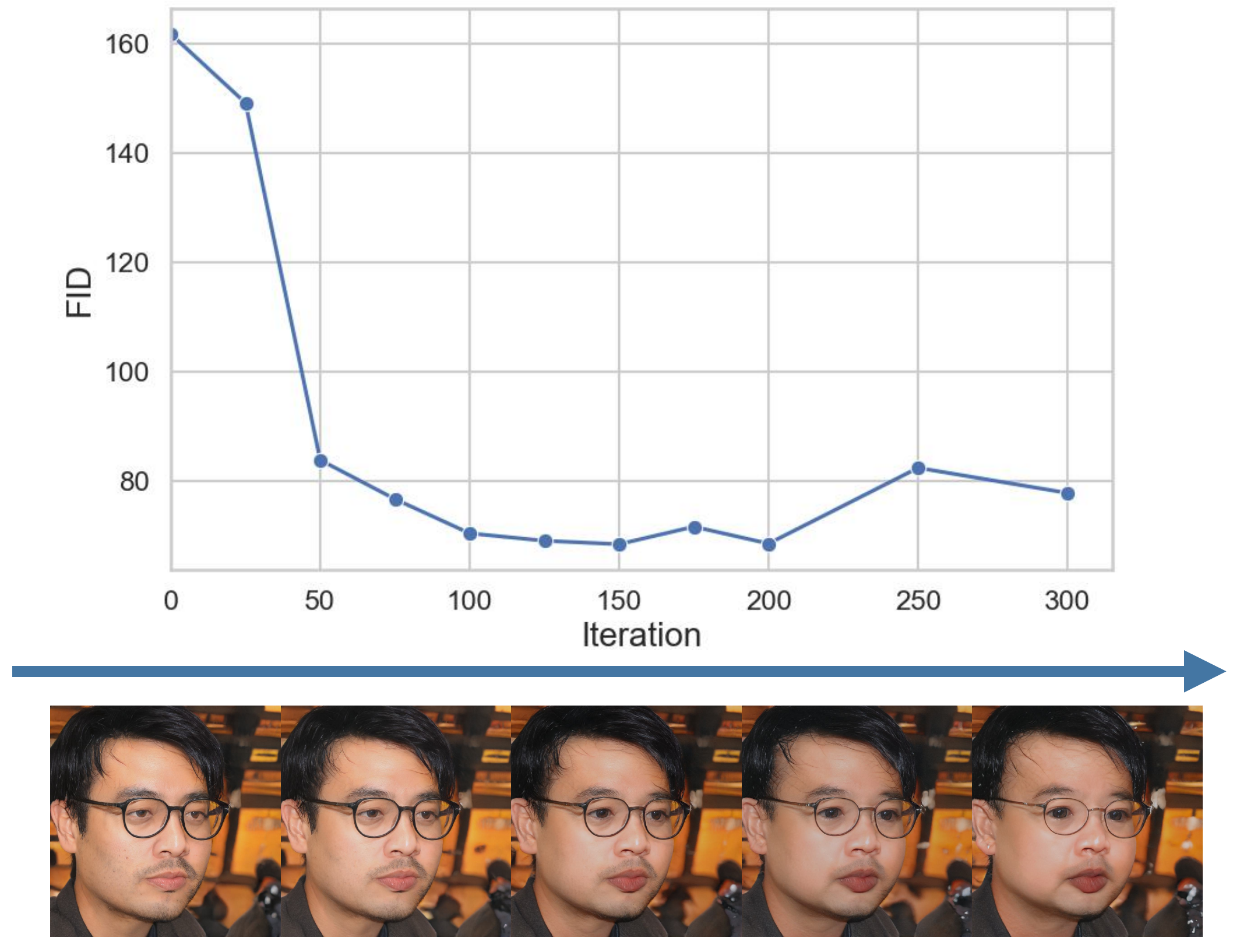}
    \caption{Concept shifts during adaptation. Our result shows that after limited iterations of adaptation using directional loss, \textbf{the encoded concept in the adapted generator is already closer to the target domain than the encoded concept in source generator.}}
    \label{fig:concept_shift}
\end{figure*}

Text-driven visual transformation is an iterative process: even a limited number of updates can move the transformed visual state closer to the target concept than the source. In this section, we design an experiment to demonstrate that the transformed visual already encodes some knowledge related to the target domain. Specifically, following zero-shot generative model domain adaptation setup \citep{gal2022stylegannada}, we perform adaptation on \texttt{Human}$\rightarrow$\texttt{Baby} with StyleGAN2-ADA pretrained on FFHQ \citep{karras2019style}. We report FID score throughout the adaptation process to measure the knowledge related to target domain encoded in the transformed visual. As shown in Fig.~\ref{fig:concept_shift}, FID decreases during the initial iterations, indicating that the generated distribution moves closer to the target domain. Additionally, we present qualitative results using the same latent code to further support our findings. 

This observation motivates the initial transformation stage of AIR described in Sec.~\ref{sec:Methodology}. Therefore, we consider NADA generally provides a useful early anchor, and we proposed AIR to subsequently refines the guidance to further improve adaptation quality, without requiring manual identification of a suitable intermediate generator before refinement. The consistency of this behavior is further supported by AIR’s improvements across all 40 adaptation settings under the same general configuration.

\subsection{User Study} \label{sec:User_study}
\begin{figure*}[!htbp]
    \centering
    \includegraphics[width=1\textwidth]{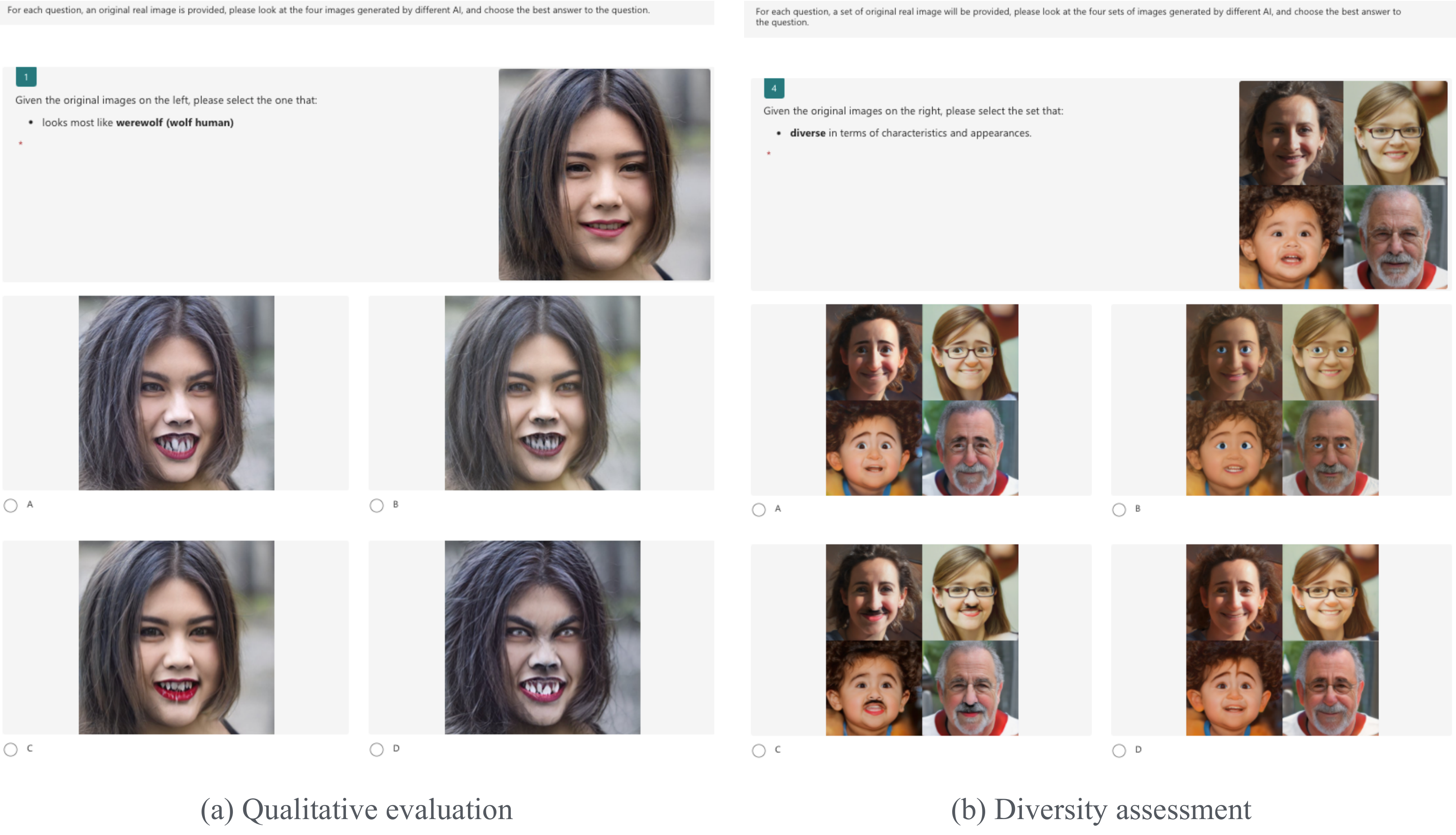}
    \caption{Examples of user study on (a) Quality assessment and (b) Diversity assessment. We omit one option in the screenshot for better presentation.}
    \label{fig:User_study}
\end{figure*}

We conduct a user study
to compare the quality and the diversity of domain adaptation with different schemes based on human feedback.
The questionnaire is performed using the generated samples by different schemes including NADA, IPL, SVL, HGAN and our proposed AIR. It includes 12 questions for quality evaluation and 4 questions for diversity assessment. We include examples for quality and diversity evaluation of our questionnaire in Fig. \ref{fig:User_study}. Finally, we report the percentage of the user preference from 220 responses for each method and for both quality and diversity metrics in 
Tab. \ref{Tab:User_study}. Our user study does not collect identifying information, and does not involve sensitive topics or interventions.

\begin{table}[ht]
    \centering
    \caption{Results of our user study (\%) experiments. }
    \label{Tab:User_study}
    \begin{tabular}{@{}c|ccccc@{}}
    \toprule
    Evaluation & NADA & IPL & SVL & HGAN & AIR \\
    \midrule
    Quality & 24.8 & 4.2 & 3.6 & 11.7 & \textbf{55.7} \\
    Diversity & 22.4 & \textbf{32.8} & 10.8 & 3.4 & 30.6 \\
    \bottomrule
    \end{tabular}
\end{table}

\subsection{Impact of Offset Misalignment on Downstream Tasks}
\label{ssec:Impact_of_Offset_Misalignment_on_Generative_Model_Adaptation}

In the previous section, we performed an empirical study that revealed image-text offset misalignment.
In this section, we take a step further and investigate the effect of this misalignment on the generative model adaptation from a source domain (concept) $\mathcal{S}$ to a target domain (concept) $\mathcal{T}$.
Specifically, following ZSGM setup \citep{gal2022stylegannada}, for source domain $\mathcal{S}$, we assume a pre-trained generator $G_\mathcal{S}$ and a textual description $T_\mathcal{S}$ is available. However, for the target domain, only textual description $T_\mathcal{T}$ is available.
To simulate different degrees of misalignment between source and target, we augment target text to get a set of textual descriptions $\{ T^i_{\mathcal{T}} \}$.
Then, we perform zero-shot adaptation using the directional loss (Eq. \ref{eq:Loss_directional}) from the source domain $\mathcal{S}$ to each of these target text $T^i_{\mathcal{T}}$ and measure the generation performance of adapted generator.

{\bf Experimental Setup.}
For this experiment, we perform adaptation on  \texttt{Human} $\rightarrow$ \texttt{Baby} and \texttt{Dog} $\rightarrow$ \texttt{Cat}.
Our contribution targets the adaptation guidance mechanism and is not tied to a specific generator architecture. To enable rapid validation of the guidance mechanism, 
we use StyleGAN2-ADA \citep{karras2020ada} pre-trained on FFHQ \citep{karras2019style} and AFHQ-Dog \citep{choi2020starganv2} as the pre-trained model for a computationally efficient adaptation testbed.
Following 
\citep{karras2020ada},
we use CLIP. 
We fix the source text $T_\mathcal{S}$ and augment the target text $T_\mathcal{T}$ by sampling handcrafted textual descriptions from CLIP ImageNet template (INt)\footnote{\scriptsize\url{https://github.com/openai/CLIP/blob/main/notebooks/Prompt\_Engineering\_for\_ImageNet.ipynb}} in order to simulate different degrees of misalignment.
Then, we follow exactly the same hyperparameters as NADA to adapt the source generator to different target text $T^i_{\mathcal{T}}$. 
We use FID to measure the performance of the adapted generator against offset misalignment.

\begin{figure*}[htbp]
    \centering
    \includegraphics[width=\textwidth]{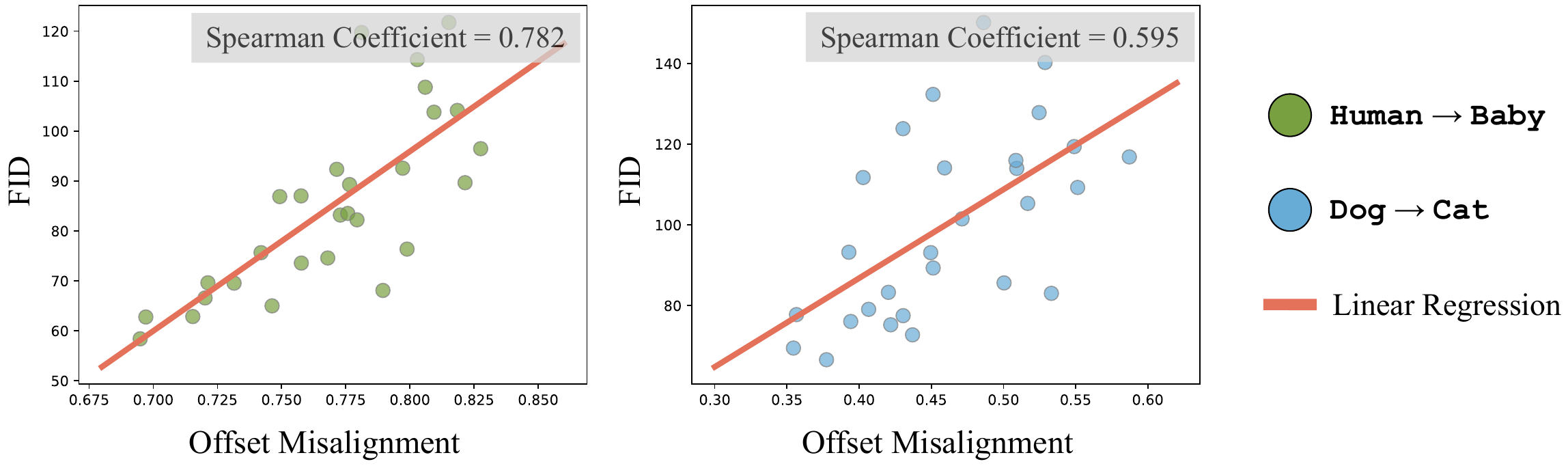}
    \caption{{\bf Impact of offset misalignment on zero-shot generative model adaptation with directional loss:} For each of the two setups, we fix the source domain and augment the textual description of the target domain to simulate various degrees of misalignment between image offset and text offset. Then, we perform the text-driven visual transformation using directional loss in Eq. \ref{eq:Loss_directional} for each setup. Results show that transformation performance degrades as offset misalignment increases.}
    \label{fig:offset vs fid}
\end{figure*}

Our results in Fig. \ref{fig:offset vs fid} demonstrate that, in general, {\bf performance of text-driven visual transformation with directional loss degrades as offset misalignment increases}.
This downstream validation supports our representation space finding and motivates AIR to iteratively refine text-offset guidance.

\subsection{Additional Experiments of ZSGM}
\label{sec:Additional_Experiments_of_ZSGM}

\subsubsection{Latent Space Interpolation} \label{sec:Latent_interpolation}

\begin{figure*}[t]
    \centering
    \includegraphics[width=1\textwidth]{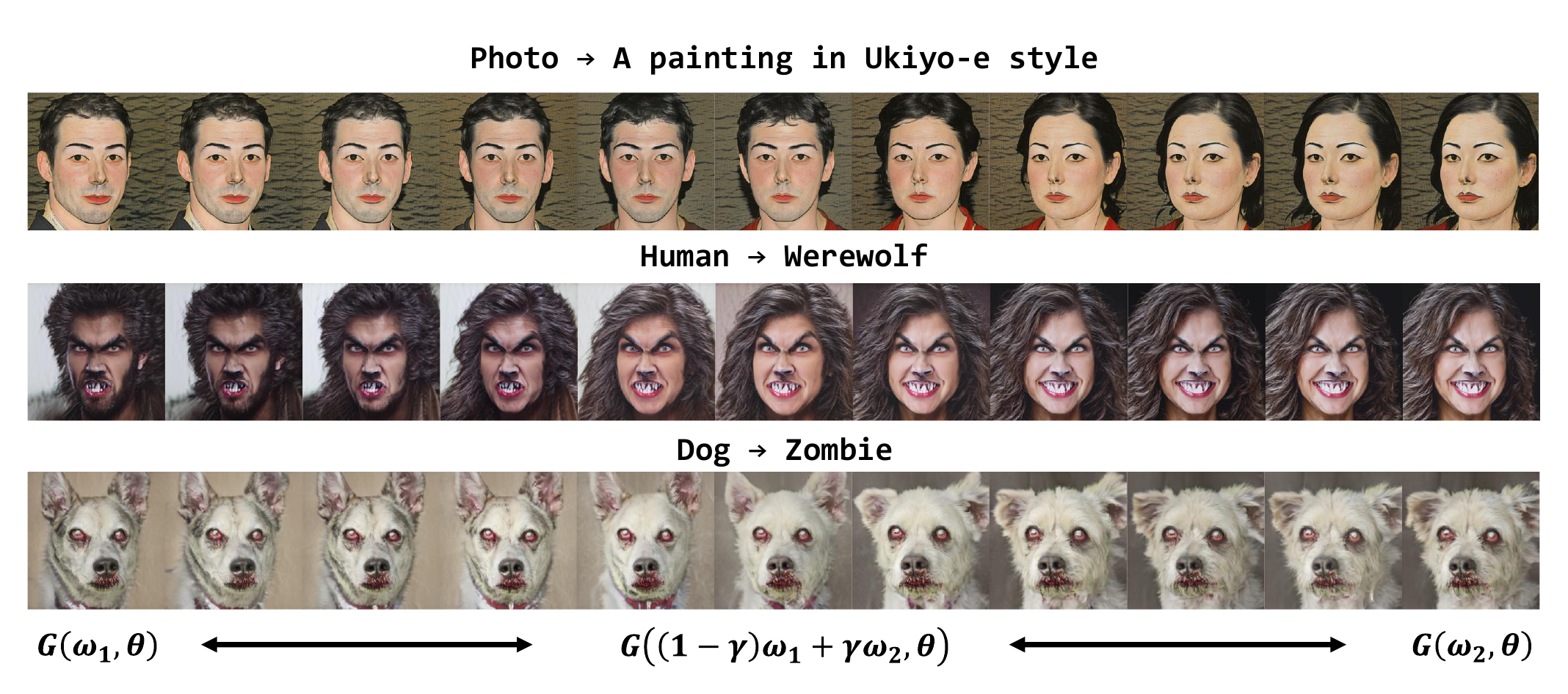}
    \caption{Latent space interpolation. For each row, the left-most column and right-most column are respectively two images synthesized with two different latent codes. The remaining columns refer to images synthesized with interpolated latent codes. Our results demonstrate that \textbf{the target domain generators refined through our method retain a smooth latent space property.}}
    \label{fig:Latent interpolation}
\end{figure*}

Building on prior research, we demonstrate that the target domain generators refined through our method retain a smooth latent space property. As illustrated in Fig. \ref{fig:Latent interpolation}, each row features a series of images from the same target domain. The left-most and right-most images in each row, labeled as $G_t(w_1)$ and $G_t(w_2)$ respectively, are generated using distinct latent codes $w_1$ and $w_2$. Latent space interpolation between these codes produces an image $G_t((1 - \gamma)w_1 + \gamma w_2)$, where $\alpha$ varies from 0 to 1.  The visual results show that our method has good robustness and generalization ability. The various target domain spaces obtained by our method are consistently smooth.

\subsubsection{Cross-model Interpolation} \label{sec:Model_interpolation}

\begin{figure*}[t]
    \centering
    \includegraphics[width=1\textwidth]{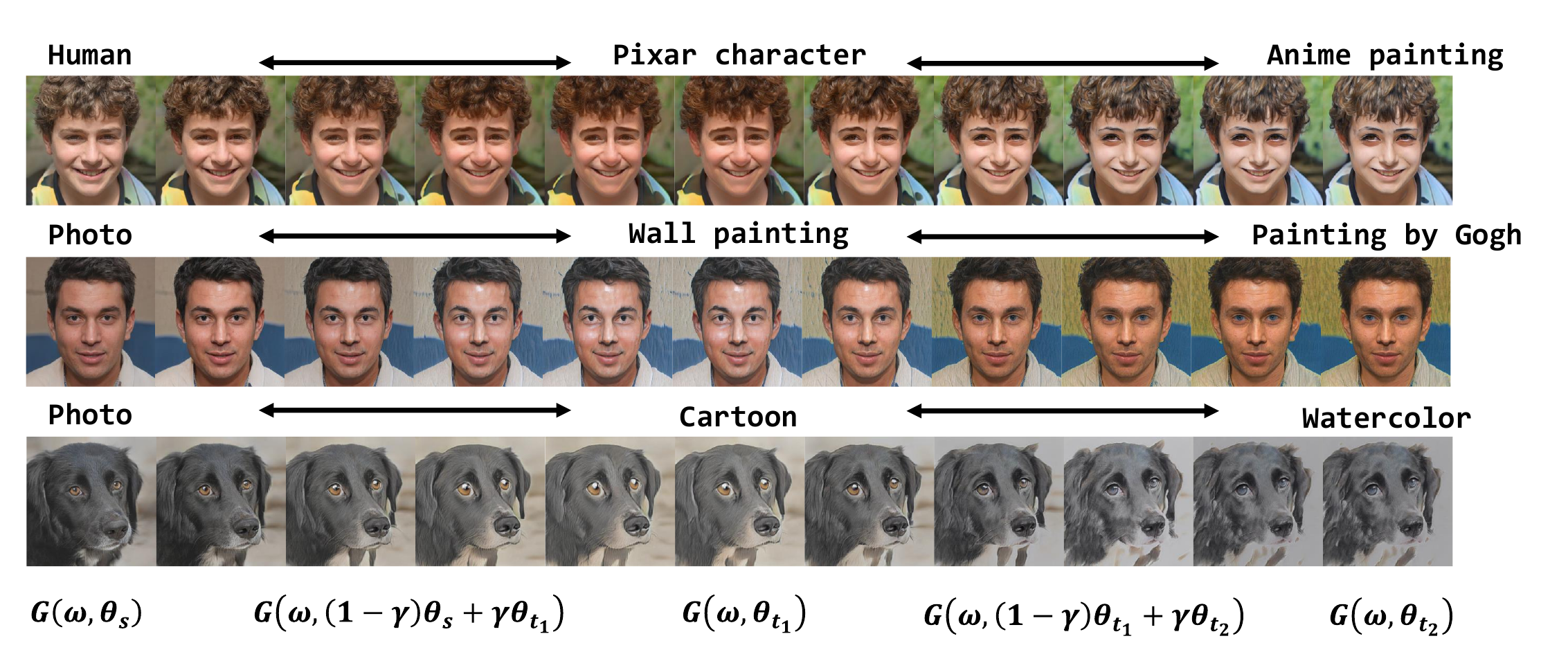}
    \caption{Cross-model interpolation. In each row, the left-most image is generated by the source generator. The middle and the right-most images are synthesized by two different target domain generators. The other images represent cross-model interpolations between two different domains. Our results shows that \textbf{the model's weight smoothness across various domains.}}
    \label{fig:Model interpolation}
\end{figure*}

In addition to demonstrating latent space interpolation, we also explore the model's weight smoothness across various domains. Specifically, we perform linear interpolation in the weight space between $G(\cdot, \theta_s)$ and $G(\cdot, \theta_{t_1})$, or between $G(\cdot, \theta_{t_1})$ and $G(\cdot, \theta_{t_2})$. Here, $G(\cdot, \theta_s)$ represents the source domain generator, while $G(\cdot, \theta_{t_1})$ and $G(\cdot, \theta_{t_2})$ are generators adapted to two different target domains. Given a latent code $w$, we produce images via an interpolated model, $G(w, (1 - \gamma)\theta_1 + \gamma\theta_2)$, where $\gamma$ ranges from 0 to 1. As illustarted in Fig. \ref{fig:Model interpolation}, our approach effectively supports smooth cross-model interpolation, whether transitioning from a source to a target domain or between different target domains.

\subsubsection{Image Manipulation} \label{Sec:Image_manipulation}

To further demonstrate the effectiveness of our proposed method, we also conduct experiments on text-to-image manipulation. It first inverts a image to the latent code by a pre-trained inversion model and then feeds it to the trained target domain generator to get the translated target domain image. 

We experiment on both GAN and diffusion model. 
We use Restyle \cite{alaluf2021restyle} with e4e encoder \cite{tov2021e4e} to invert a real image into the latent space $w$ for StyleGANs. 
For the diffusion model, we follow the setting of DiffusionCLIP \cite{diffusionclip} to diffuse a real image and fintune the model to generate an image with target domain features using the diffused image.

\begin{figure*}[t]
    \centering
    \includegraphics[width=1\textwidth]{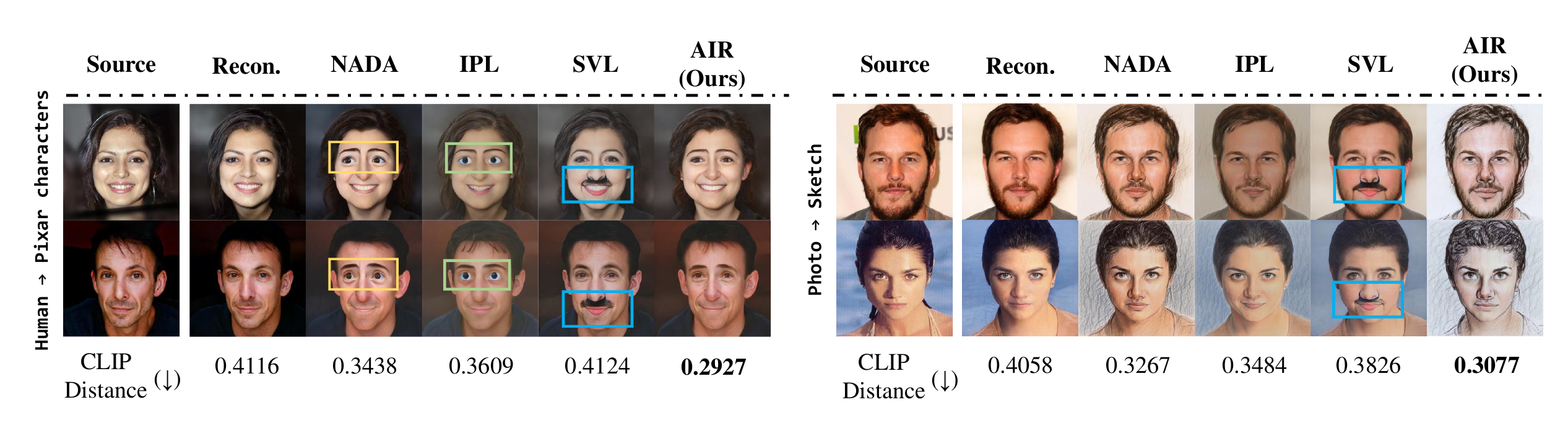}
    \caption{Image manipulation with GAN. The reference image are the same as in 
    Fig. 1
    }
    \label{fig:GAN-image translation}
\end{figure*}

\textbf{GAN-based Image Manipulation} \label{ssec:GAN image manipulation}
For GAN-based generators, we perform the experiment by utilizing the inversion model Restyle \cite{alaluf2021restyle} with e4e encoder \cite{tov2021e4e}. As illustrated in Fig. \ref{fig:GAN-image translation}, our method qualitatively exhibits a higher fidelity of target domain features compared to previous methods. Quantitatively, our approach more closely aligns with the reference target images, indicating a greater semantic similarity. 

\textbf{Diffusion-based Image Manipulation} \label{ssec:Diffusion image manipulation}

We implement based on DiffusionCLIP \cite{diffusionclip} which seamlessly integrates with the existing zero-shot adaptation methods.

\begin{figure*}[t]
    \centering
    \includegraphics[width=1\textwidth]{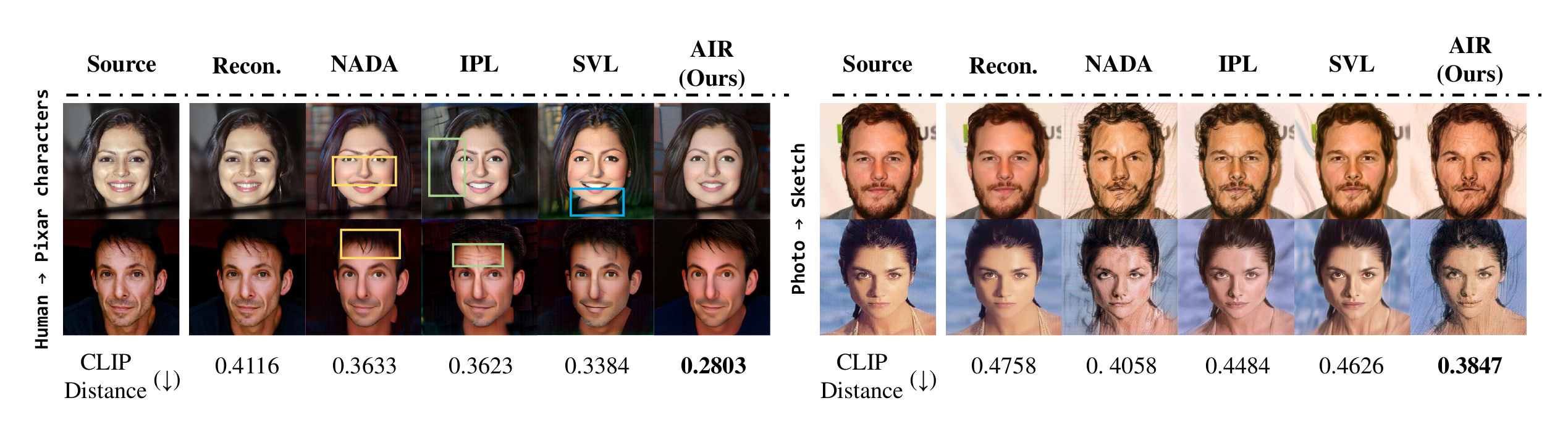}
    \caption{Diffusion model image manipulation.}
    \label{fig:Image translation}
\end{figure*}

As illustrated in Fig Fig. \ref{fig:Image translation}, our method qualitatively exhibits a higher fidelity of target domain feature compared to previous methods. Quantitatively, our approach more closely aligns with the reference target images, indicating a greater semantic similarity.

\begin{figure*}[t]
    \centering
    \includegraphics[width=1\textwidth]{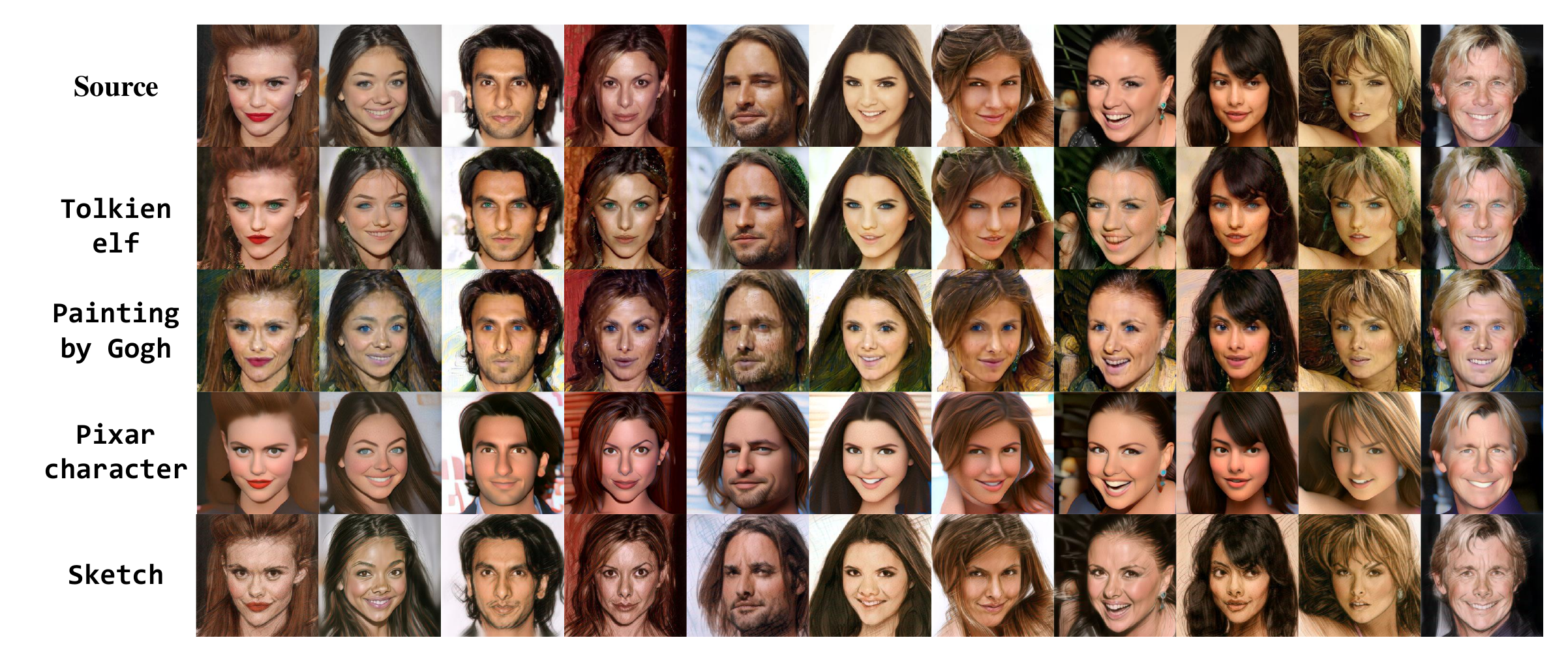}
    \caption{Additional results of image manipulation with diffusion model.}
    \label{fig:Supp-Diff}
\end{figure*}

Fig. \ref{fig:Supp-Diff} illustrates real-world image manipulation results for diffusion AIR.

\section{Related Work} \label{sec:Related_work}

\textbf{Analogical Reasoning}
Research in NLP has shown that word representations of language models are surprisingly good at capturing semantic regularities in language \cite{Collobert2008, turian2010word}.
Specifically, analogical reasoning \cite{mikolov2013linguistic, mikolov2013efficient, mikolov2013distributed, levygoldberg2014linguistic}, utilizing the semantic regularities of word representations, aims to solve analogy tasks by using one pair of word vectors to identify the unknown member of a different pair of words, commonly via alignment of offsets, 
This is commonly modeled as using the vector offset between two words $a'-a$, and applying it to a new word $b$ to predict the missing word $b'$ that pair with $b$,
as illustrated by the famous example of using \textit{v(“Man”) - v(“Woman”)} and \textit{v(“King”)} to identify \textit{v("Queen")}, where \textit{v($\cdot$)} denotes word representation.
This approach attracted a lot of attention for the vital role that analogical reasoning plays in human cognition for discovering new knowledge and understanding new concepts.
It is already used in many downstream NLP tasks, such as splitting compounds \cite{daiber-etal-2015-splitting}, semantic search \cite{Expansion-by-Analogy}, cross-language relational search \cite{Duc}, etc.

Importantly, previous works \cite{levy2015improving, koper2015multilingual, vylomova2015take} demonstrate that the effectiveness of analogical reasoning varies across different categories and semantic relations. 
More recent studies \cite{rogers2017too, fournier2020analogies}, present a series of experiments performed with BATS dataset \cite{BATS} on various pre-trained vector space, e.g., GloVe \cite{glove}, Word2Vec \cite{mikolov2013distributed}, and Skip-gram \cite{mikolov2013efficient}, 
indicate that it is more effective to use $a'-a$ and $b$ to determine $b'$
when $b$ and $b'$ are close in vector space; and less so when $b$ and $b'$ are more apart.

Inspired by these studies, in this work, we perform an empirical study of offset misalignment in multimodal contrastive representation spaces 
and observe that for distant concepts in multimodal contrastive representation space, image and text offsets suffer from more misalignment, while closely related concepts suffer less. 
Based on this analysis, AIR iteratively refines text offsets to reduce image-text offset misalignment and improve directional guidance in text-driven visual transformation tasks.

{\bf Generative model adaptation} studies how a pre-trained generator can be adapted to new domains under limited supervision while preserving sample quality and diversity
\cite{li2020ewc, ojha2021cdc, zhao2022dcl, zhou2024deformable, Anees2024HyperGANCLIP, zhu2025domainstudio, guo2025everything, cai2025diffusion, abdollahzadeh2025a}.
A particularly challenging setting within generative model adaptation is {\em Zero-Shot Generative Model Adaptation (ZSGM)}, where no target domain (concept) images are available and adaptation must rely solely on semantic descriptions of the target domain (concept). 
Recent methods address this problem by leveraging multimodal contrastive models such as CLIP or SigLIP to derive directional guidance from textual descriptions.
In this framework, textual descriptions of the target domain guide generator updates by aligning semantic directions in the representation space, enabling the generator to shift from the source domain toward the target domain \cite{gal2022stylegannada, guo2023ipl, jeon2023svl}.
In particular, approaches such as StyleGAN-NADA \cite{gal2022stylegannada} use text offsets in the joint image-text representation space to guide the visual domain transformation during adaptation.

{\bf Image editing} studies how an input image can be modified according to a desired semantic change while preserving content irrelevant to the edit \cite{shen2020interpreting, harkonen2020ganspace, wu2021stylespace, abdal2021styleflow, meng2022sdedit}. 
A particularly flexible setting is {\em text-guided image editing (TGIE)}, where the desired manipulation is specified by natural-language descriptions rather than predefined attribute labels, reference images, or manually discovered editing directions. 
Similarly, recent TGIE methods leverage multimodal contrastive models such as CLIP to derive directional guidance from text \cite{patashnik2021styleclip, diffusionclip, kwon2022clipstyler, park2025styleeditor}. 
In this framework, the edit is modeled as a visual transformation in representation space: the source-to-edited image offset is encouraged to follow the corresponding text offset that describes the intended semantic change. 
For example, Style-Editor uses text offsets between source object (concept) descriptions and target object (concept) descriptions to guide object-level visual transformations for text-driven image editing \cite{park2025styleeditor}.

However, the adaptation direction in these two branch of works only focuses on the source and target concepts and is computed once before the visual transformation.

{\bf More application of image-text offset alignment.}
Several works address related aspects of text-driven visual transformation. Embedding Arithmetic of Multimodal Queries \citep{couairon2022embedding} evaluates offset-vector arithmetic for transformation-based image retrieval, but does not independently quantify the alignment between corresponding visual and textual offsets. DeltaEdit \citep{lyu2023deltaedit} assumes image-text offset alignment and replaces image offsets with text offsets to drive manipulation. EditCLIP \citep{wang2025editclip} learns a task-specific visual encoder to better represent edits and unintended structural changes, rather than analyzing the alignment between native image and text offsets. AugCLIP \citep{kim2025augclip} constructs an ideal edited-image representation from the source image and target text for evaluation, thereby retaining the alignment assumption rather than testing it.

However, none of these works systematically investigates the degree of image-text offset alignment, and assume the image and text offsets in multimodal contrastive representation space are well aligned. 
In contrast, our work is the first to directly measures the degree of image-text offset alignment and systematically studies how it varies across concept pairs.
In this paper, we draw inspiration from a similar problem called analogical reasoning in NLP, and empirically discover that the alignment of image and text offsets in multimodal contrastive representation space is correlated with concept proximity.
Based on this finding, AIR iteratively updates the transformation direction, effectively reducing the downstream effect of offset misalignment and achieving SoTA performance on both text-driven visual transformation tasks.



\section{Limitation} \label{sec:limitation}

Our proposed iterative refinement method seeks to improve the quality of zero-shot adaptation. 
As noted by \cite{guo2023ipl}, achieving adaptation across large domain gaps, such as \texttt{Human} to \texttt{Cat}, is particularly challenging. Similar to previous work, our approach necessitates that the trained generator somewhat closely approximates the target domain before initiating iterative refinement.
Additionally, while our experiments on 40 different setups are comprehensive compared to previous work, more setups can be experimented to understand the limitations. 
We also note that our image-text offset alignment analysis focuses on CLIP/SigLIP-like multimodal contrastive representation spaces and downstream tasks that explicitly use image-text offsets or directional guidance.

\section{Social Impact} \label{sec:social_impact}
The AIR methodology holds potential for enhancing artistic image synthesis in social media contexts and could serve as a beneficial data augmentation tool in other computer vision tasks such as recognition and detection. However, its capability to generate realistic images from real-world data raises ethical considerations. It is crucial to address these issues thoughtfully to prevent misuse and ensure responsible application of this technology.



\section{Licenses} 
\label{sec:licenses}

In Table Tab. \ref{Tab:licenses}, we specify the source and licenses of the models and datasets used in our work. Note that the FFHQ dataset consists of facial images collected from Flickr, which are under permissive licenses for non-commercial purposes.

\begin{table}[t]
    \centering
    \scalebox{0.9}{
    \begin{tabular}{|l|l|}
    \hline
    \textbf{Models} & \textbf{License} \\
    \hline
    StyleGAN2 \cite{Karras2019stylegan2} & Nvidia Source Code License \\
    CLIP \cite{radford2021CLIP} & MIT License \\
    StyleGAN2-pytorch \cite{Karras2019stylegan2} & MIT License \\
    e4e \cite{tov2021e4e} & MIT License \\
    StyleGAN-NADA \cite{gal2022stylegannada} & MIT License \\
    IPL \cite{guo2023ipl} & MIT License \\
    \hline
    \textbf{Datasets} & \textbf{License} \\
    \hline
    FFHQ \cite{Karras2019stylegan2} & CC BY-NC-SA 4.0 \\
    AFHQ \cite{choi2020starganv2} & CC BY NC 4.0 \\
    \hline
    \end{tabular}
    }
    \caption{Sources and licenses of the utilized models and datasets}
    \label{Tab:licenses}
\end{table}



\end{document}